\pdfoutput=1
\documentclass[11pt]{article}

\usepackage[]{EMNLP2023}

\usepackage{times}
\usepackage{latexsym}

\usepackage[T1]{fontenc}
\usepackage[utf8]{inputenc}

\usepackage{microtype}

\usepackage{inconsolata}

\usepackage{booktabs}
\usepackage[nolist]{acronym}                % abbreviations

\usepackage{rotating}
\usepackage{adjustbox}
\usepackage[inline]{enumitem}

\usepackage{afterpage}
\usepackage{adjustbox}

\title{LEXTREME: A Multi-Lingual and Multi-Task Benchmark \\ for the Legal Domain}

\author{
Joel Niklaus $^{1,2,6}$\thanks{\hspace{2mm} Equal contribution.}
\quad
Veton Matoshi $^{2*}$
\quad
Pooja Rani $^{3}$\AND
\quad
Andrea Galassi $^{4}$
\quad
Matthias Stürmer $^{1,2}$
\quad
Ilias Chalkidis $^{5}$\\\\
$^1$University of Bern\quad
$^2$Bern University of Applied Sciences\quad
$^3$University of Zurich\\
$^4$University of Bologna\quad
$^5$University of Copenhagen\quad
$^6$Stanford University\\
}

\begin{document}
\maketitle

\begin{acronym}[UMLX]
    \acro{FSCS}{Federal Supreme Court of Switzerland}
    \acro{SCI}{Supreme Court of India}
    \acro{ECHR}{European Convention of Human Rights}
    \acro{ECtHR}{European Court of Human Rights}
    \acro{SCOTUS}{Supreme Court of the United States}
    \acro{SPC}{Supreme People's Court of China}
    \acro{SJP}{Swiss-Judgment-Prediction}
    \acro{ASO}{Almost Stochastic Order}
    \acro{ILDC}{Indian Legal Documents Corpus}
    
    \acro{US}{United States}
    \acro{EU}{European Union}

    \acro{NLP}{Natural Language Processing}
    \acro{ML}{Machine Learning}
    \acro{LJP}{Legal Judgment Prediction}
    \acro{SJP}{Swiss-Judgment-Prediction}
    \acro{PJP}{Plea Judgment Prediction}
    
    \acro{BERT}{Bidirectional Encoder Representations from Transformers}
    \acro{LSTM}{ Long Short-Term Memory }
    \acro{GRU}{Gated Recurrent Unit}
    \acro{BiLSTM}{Bidirectional Long Short-Term Memory}
    \acro{CNN}{Convolutional Neural Networks}

    \acro{PLM}{pre-trained Language Model}

    \acro{CLT}{Cross-Lingual Transfer}
    \acro{HRL}{high resource language}
    \acro{LRL}{low resource language}

    \acro{POS}{Part-of-Speech}
    
    \acro{SLTC}{Single Label Text Classification}
    \acro{MLTC}{Multi Label Text Classification}
    \acro{NER}{Named Entity Recognition}
    \acro{NLU}{Natural Language Understanding}

    \acro{GNB}{Gaussian Naive Bayes}
    \acro{DT}{Decision Tree}
    \acro{SVM}{Support-vector Machine}
    \acro{RF}{ Random Forest}
    \acro{XGBoost}{eXtreme Gradient Boosting}
\end{acronym}

\begin{abstract}
Lately, propelled by phenomenal advances around the transformer architecture, the legal NLP field has enjoyed spectacular growth. To measure progress, well-curated and challenging benchmarks are crucial. Previous efforts have produced numerous benchmarks for general NLP models, typically based on news or Wikipedia. However, these may not fit specific domains such as law, with its unique lexicons and intricate sentence structures. Even though there is a rising need to build NLP systems for languages other than English, many benchmarks are available only in English and no multilingual benchmark exists in the legal NLP field. We survey the legal NLP literature and select 11 datasets covering 24 languages, creating LEXTREME. To fairly compare models, we propose two aggregate scores, i.e., dataset aggregate score and language aggregate score. Our results show that even the best baseline only achieves modest results, and also ChatGPT struggles with many tasks. This indicates that LEXTREME remains a challenging task with ample room for improvement. To facilitate easy use for researchers and practitioners, we release LEXTREME on huggingface along with a public leaderboard and the necessary code to evaluate models. We also provide a public Weights and Biases project containing all runs for transparency.
\end{abstract}

\section{Introduction}
\label{sec:introduction}

\begin{figure}[ht]
    \centering
    \resizebox{\columnwidth}{!}{
    \includegraphics{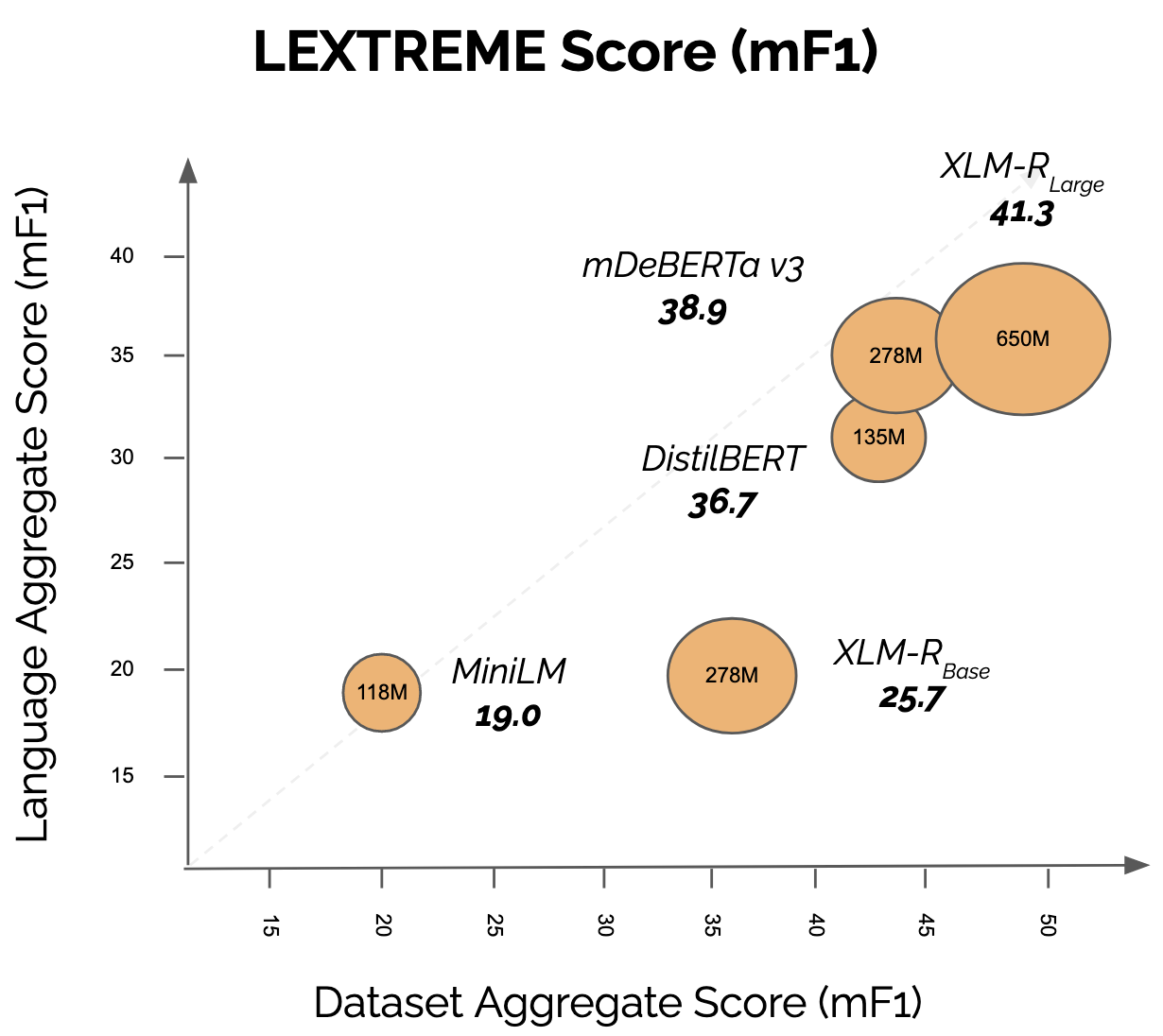}
    }
    %\vspace{-8mm}
    \caption{Overview of multilingual models on LEXTREME. The bubble size and text inside indicate the parameter count. The bold number below the model name indicates the LEXTREME score (harmonic mean of the language agg. score and the dataset agg. score).}
    \label{fig:lextreme}
    \vspace{-3mm}
%\vspace{-5mm}
\end{figure}

% NLP is relevant for Legal AI
In the last decade, Natural Language Processing (NLP) has gained relevance in Legal Artificial Intelligence, transitioning from symbolic to subsymbolic techniques \cite{Villata2022ThirtyYO}.
Such a shift is motivated partially by the nature of legal resources, which appear primarily in a textual format (legislation, legal proceedings, contracts, etc.).
Following the advancements in NLP technologies, the legal NLP literature \cite{zhong-etal-2020-nlp, nllp-2022-natural, katz_natural_2023} is flourishing with many new resources, such as large legal corpora \cite{henderson_pile_2022}, task-specific datasets \cite{shen2022multilexsum,christen_resolving_2023,brugger_multilegalsbd,niklaus_automatic_2023}, and pre-trained legal-oriented language models \cite{chalkidis-etal-2020-legal,zlucia/custom-legalbert,lawformer,niklaus-giofre-2023-pretrain,hua_legalrelectra_2022, chalkidis-etal-2023-lexfiles}.
\citet{DBLP:journals/corr/abs-2308-05502} offer a comprehensive survey on the topic.

% Foundation Models are very good on GLUE, SuperGLUE
Specifically, the emergence of \acp{PLM}
% AG: There is debate regarding the term "foundation models", some people consider it "Stanford propaganda", I would avoid using it, in case we get a reviewer that thinks this.
% Foundation Models (large neural networks trained on vast corpora) \cite{bommasani_opportunities_2022}
has led to significant performance boosts on popular benchmarks like GLUE \cite{GLUE} or SuperGLUE \cite{SUPERGLUE}, emphasizing the need for more challenging benchmarks to measure progress. 
Legal benchmark suites have also been developed to systematically evaluate the performance of \acp{PLM}, showcasing the superiority of legal-oriented models over generic ones on downstream tasks such as legal document classification or question answering \cite{chalkidis-etal-2022-lexglue,hwang2022a}.
%\pr{give one or two examples of such downstream tasks and cite the paper that shows the superiority}.
% \todo{Andrea}
%\textcolor{magenta}{briefly talk about language models, bert ecc}

%\todo{Write about recent rise of large language models, but argue, that they are not good yet on tasks with large label sets}

% general purpose models are likely insufficient for legal tasks
Even though these PLMs are shown to be effective for numerous downstream tasks, they are general-purpose models that are trained on broad-domain resources, such as Wikipedia or News, and therefore, can be insufficient to address tasks specific to the legal domain \cite{chalkidis-etal-2020-legal,hua_legalrelectra_2022,niklaus-giofre-2023-pretrain}.
%\pr{maybe we can cite more papers that shows the need of legal domain specific LLMs}
%However, general-purpose models, trained on resources such as Wikipedia, may be insufficient to address tasks in the legal domain.
Indeed, the legal domain is strongly characterized both by its lexicon and by specific knowledge typically not available outside of specialized domain resources. Laypeople even sometimes call the language used in legal documents ``legalese'' or ``legal jargon'', emphasizing its complexity.
Moreover, the length of a legal document usually exceeds the length of a Wikipedia or news article, and in some tasks the relationships between its entities may span across the entire document.
Therefore, it is necessary to develop specialized Legal \acp{PLM} trained on extensive collections of legal documents and evaluate them on standardized legal benchmarks.
While new PLMs capable of handling long documents have been developed in the last years, they are predominantly trained for the general domain and on English data only.

% GLUE contains many linguistic tasks with no direct applications
%Existing benchmarks, such as GLUE, often tackle linguistic tasks, such as semantic textual similarity or natural language inference, with no direct application in mind. There is a need for benchmarks that tackle use cases as close as possible to the real world to align model development with practical deployment needs. 
% In the field of legal NLP, there exist datasets, with clear and valuable use cases, such as the prediction of court cases from the allegations \cite{semo_classactionprediction_2022} or question answering for legal contract review \cite{hendrycks_cuad_2021}.

% Need for multilingual benchmarks
The rising need to build NLP systems for languages other than English, the lack of textual resources for such languages, and the widespread use of code-switching in many cultures~\cite{10.3389/fpsyg.2020.02130} is pushing researchers to %devise new multilingual learning approaches 
train models on massively multilingual data \cite{conneau-etal-2020-unsupervised}.
%\pr{may be an example of one such multi-lingual approach or model}.
Nonetheless, to the best of our knowledge, no multilingual legal language model has been proposed so far.
% Consequently, standardized multilingual benchmarks are needed to evaluate these models.
Consequently, there is a need for standardized multilingual benchmarks that can be used to evaluate existing models and assess whether more research efforts should be directed toward the development of domain-specific models.
This is particularly important for legal NLP where   inherently multinational (European Union, Council of Europe) or multilingual (Canada, Switzerland) legal systems are prevalent.
% \textcolor{magenta}{expand talking about the importance of multilingual benchmarks}

% we have the solution here
In this work, we propose a challenging multilingual benchmark for the legal domain, named LEXTREME. %,  which contains datasets with relevant use cases. 
We survey the literature from 2010 to 2022 and select 11 relevant NLU datasets
%out of 108 papers based on the formulated exclusion and inclusion criteria. 
covering 24 languages in 8 subdivisions (Germanic, Romance, Slavic, Baltic, Greek, Celtic, Finnic, and Hungarian) from two language families (Indo-European and Uralic).
We evaluate five widely used multilingual encoder-based language models as shown in \autoref{fig:lextreme} and observe a correlation between the model size and performance on LEXTREME. Surprisingly, at the low end, DistilBERT \cite{Sanh2019DistilBERTAD} strongly outperforms MiniLM \cite{Wang2020} (36.7 vs 19.0 LEXTREME score) while only having marginally more parameters (135M vs 118M).

For easy evaluation of future models, we release the aggregate dataset on the huggingface hub \footnote{\url{https://huggingface.co/datasets/joelniklaus/lextreme}}
along with a public leaderboard  and the necessary code to run experiments on GitHub.%\footnote{Anonymous repository for review purpose: \url{https://www.dropbox.com/scl/fo/xiy78g2gxlkpvmrt10e2l/h?dl=0&rlkey=13v1d7lqb3f6poboasmjekxas}}
\footnote{\url{https://github.com/JoelNiklaus/LEXTREME}}
Knowing that our work can not encompass ``Everything in the Whole Wide Legal World'' \cite{WWWB}, we design 
% To further our goal of continuing
LEXTREME as a living benchmark and provide detailed guidelines on our repository and encourage the community to contribute high-quality multilingual legal datasets.\footnote{Since the release of this call in February 2023, already eleven new tasks have been contributed and integrated.} Finally, we integrated LEXTREME together with the popular English legal benchmark LexGLUE \cite{chalkidis-etal-2022-lexglue} into HELM \cite{liang_holistic_2022} (an effort to evaluate language models holistically using a large number of datasets from diverse tasks) to ease the adoption of curated legal benchmarks also for the evaluation of large language models such as GPT-3 \cite{brown_language_2020}, PALM \cite{chowdhery_palm_2022} or LLaMA \cite{touvron_llama_2023}.

%\todo{add monolingual results}
%\todo{Discuss Author order}

\noindent\textbf{Contributions} of this paper are two-fold:
\setlist{nolistsep}
\begin{enumerate}[itemsep=0em]
    % build a multilingual legal benchmark
    \item We review the legal NLP literature to find relevant legal datasets and compile a multilingual legal benchmark of 11 datasets in 24 languages from 8 language groups.
    % evaluate various baselines on it
    \item We evaluate several baselines on LEXTREME to provide a reference point for researchers and practitioners to compare to.
\end{enumerate}

\section{Related Work}
\label{s_related}

\subsection{Benchmarks}

\begin{table*}[t]
    \centering
    \footnotesize
    \resizebox{\textwidth}{!}{%
    \begin{tabular}{llcrrccc}
         \toprule
         \textbf{Name} & \textbf{Source} & \textbf{Domain}   & \textbf{Tasks} & \textbf{Datasets} & \textbf{Languages} & \textbf{Agg. Score} \\
         \midrule
         GLUE       & \cite{GLUE}           & Misc. Texts       & 7  & 9  & English & Yes\\
         SUPERGLUE  & \cite{SUPERGLUE}      & Misc.  Texts      & 8  & 8  & English & Yes\\
        MMLU       & \cite{MMLU}           & Misc. Texts       & 1  & 57 & English & Yes \\
         CLUE       & \cite{CLUE}           & Misc.   Texts     & 9  & 9  & Chinese & Yes \\
         XTREME     & \cite{XTREME}         & Misc. Texts       & 6  & 9  & 40      & Yes \\
         BLUE       & \cite{BLUE-benchmark} & Biomedical Texts  & 5  & 10 & English & Yes \\
         CBLUE      & \cite{cblue}          & Biomedical Texts  & 9  & 9  & Chinese & Yes \\
         LegalBench & \cite{legalbench}     & Legal Texts       & 44 & 8. & English    & No \\
         LexGLUE    & \cite{chalkidis-etal-2022-lexglue}        & Legal Texts       & 7  & 6  & English & Yes \\
         FairLex    & \cite{chalkidis_fairlex_2022}   & Legal Texts  & 4  & 4  & 5 & No \\
         LBOX       & \cite{hwang2022a}     & Legal Texts       & 5  & 5  & Korean  & Yes \\
         LEXTREME   & (our work)            & Legal Texts       & 18 & 11 & 24      & Yes \\
         \midrule
         SUPERB     & \cite{SUPERB}         & Speech            & 10 & 10 & English & No \\
         SUPERB-SG  & \cite{SUPERB-SG}      & Speech            & 5  & 5  & English & No\\
         \midrule
         TAPE       & \cite{TAPE}           & Proteins          & 5  & 5  & n/a     & No \\
        \bottomrule
    \end{tabular}
    }
    \caption{Characteristics of popular existing NLP benchmarks.}
    %\pr{are these benchmarks sorted by some order? alphabetically, by domain? language? We can sort on some order and remove duplicates from some columns (it will look more spacious then)}
    \label{tab:benchmarks}
\end{table*}

Benchmarking is an established method to enable easy and systematic comparison of approaches.
\textsc{GLUE}~\cite{GLUE} is one of the first benchmarks to evaluate general-purpose neural language models. It is a set of supervised sentence understanding predictive tasks in English that were created through aggregation and curation of several existing datasets.
%
% The experimental results of \citeauthor{GLUE} showed the advantages of training language models in a multi-task learning and therefore the benefits of sharing knowledge across different tasks.
%
However, it became quickly obsolete due to advanced contextual language models, such as BERT \cite{devlin-etal-2019-bert}, which excelled on most tasks.
Subsequently, its updated version, named \textsc{SuperGLUE}~\citep{SUPERGLUE} was proposed, incorporating new predictive tasks that are solvable by humans but are difficult for machines. 
Both benchmarks proposed an evaluation score
computed as an aggregation of the scores obtained by the same model on each task.
They are also agnostic regarding the pre-training of the model, and do not provide a specific corpus for it.
%
% \agc{Include CLUE\cite{CLUE}}
%
Inspired by these works, numerous benchmarks have been proposed over the years. We describe some well-known ones in ~\autoref{tab:benchmarks}.
%many other benchmarks have been proposed, Table~\ref{tab:benchmarks} provides an overview of the most popular ones.

The \textsc{MMLU} benchmark is specifically designed to evaluate the knowledge acquired during the pre-training phase of the model by featuring only zero-shot and few-shot learning tasks~\cite{MMLU}. 
%It contains about 16K multiple-choice questions divided into 57 subtasks, spanning subjects in the humanities, social and hard sciences, etc.
Similarly, \textsc{SUPERB}~\cite{SUPERB} and \textsc{SUPERB-SG}~\cite{SUPERB-SG} were proposed for speech data, unifying well-known datasets.
However, they mainly vary in tasks, e.g., \textsc{SUPERB-SG} includes both predictive and generative tasks, which makes it different from the other benchmarks discussed in this section. Additionally, \textsc{SUPERB-SG} includes diverse tasks, such as speech translation and cross-lingual automatic speech recognition, which require knowledge of languages other than English. Neither of the two (\textsc{SUPERB} or \textsc{SUPERB-SG}) proposes an aggregated score.

% The idea of processing protein sequences in a similar fashion to sentences in documents has led to the development of % many machine learning solutions for bioinformatics domains.
% \textsc{TAPE} benchmark~\citep{TAPE}. It contains a dataset for the self-supervised task of modeling the ``languages of proteins'', and 5 supervised tasks that address 3 major subfields of protein sciences.
% The authors do not propose a single aggregated score for the whole benchmark. %, maybe because the scores obtained on the different tasks are not comparable.

\textsc{XTREME}~\cite{XTREME} is a benchmark specifically designed to evaluate the ability of cross-lingual generalization of models. It includes six cross-lingual predictive tasks over ten datasets of miscellaneous texts, covering a total of 40 languages. While some original datasets in it were already designed for cross-lingual tasks, others were extended by translating part of the data using human professionals and automatic methods.

% In certain domains of application, the language used in documents presents aspects so peculiar (e.g., frequent technical lexicon and uncommon expressions) that language models trained on ``general domain'' texts are insufficient. The need for domain-specific language models is leading to the creation of domain-specific benchmarks that can properly assess the performance of a model within its domain of application~\cite{XXX}.
%
% For example, the \textsc{BLUE} benchmark\footnote{Not be confused with the homonym evaluation metric.}~\cite{BLUE-benchmark} was designed for the biomedical domain and includes 5 tasks over 10 datasets covering both biomedical and clinical texts.
% \agc{Include CBLUE\cite{cblue}}

%\todo{Talk about HELM}

\subsection{Benchmarks for the Legal Domain}

\textsc{LexGLUE}~\cite{chalkidis-etal-2022-lexglue} is the first benchmark for the legal domain and covers six predictive tasks over five datasets made of textual documents in English from the US, EU, and Council of Europe.
%
%The tasks cover a wide range of NLP applications in the legal domain and specifically are:
%(i) given \textit{facts} about a case of the European Court of Human Rights, find which articles was violated;
%(ii) given an \textit{opinion} of the US Supreme Court, classify it in issue areas (e.g., Criminal Procedure, Civil Rights, etc.);
%(iii) given a document of the European Union legislation, classify its  concepts;
%(iv) given a \textit{contract provision} from the US Securities and Exchange Commission, classify its category; 
%(v) given a \textit{terms of service contract}, identify which sentences potentially violate user rights according to EU consumer law;
%(vi) given an excerpt from a US court decision that contains a reference to a particular case, choose the correct \textit{holding} out of 5 possibilities.
%
% \agc{Not sure about this:}
While some tasks may not require specific legal knowledge to be solved, others would probably need, or at least benefit from, information regarding the EU or US legislation on the specific topic.
%\pr{Why EU and US legislation? Which are such topics?}.
%
Among the main limitations of their benchmark, \citeauthor{chalkidis-etal-2022-lexglue} highlight its monolingual nature and remark that ``\textit{there is an increasing need for developing models for other languages}''.
% 
%Our work is strongly inspired by \textsc{LexGLUE} and our purpose is to propose a benchmark that, we hope, will help the development of multilingual models for the legal domain.
In parallel, \citet{chalkidis_fairlex_2022} released FairLex, a multilingual benchmark for the evaluation of fairness in legal NLP tasks.
With a similar aim, \citet{hwang2022a} released the \textsc{LBOX} benchmark, covering two classification tasks, two legal judgment prediction tasks, and one Korean summarization task.
%However, it focused only on one language. 
Motivated by \textsc{LexGLUE} and \textsc{LBOX}, we propose a benchmark to encourage multilingual models, diverse tasks, and datasets for the legal domain.
\citet{legalbench} proposed the \textsc{LegalBench} initiative that aims to establish an open and collaborative legal reasoning benchmark for few-shot evaluation of LLMs where legal practitioners and other domain experts can contribute by submitting tasks.
At its creation, the authors have already added 44 lightweight tasks. While most tasks require legal reasoning based on the common law system (mostly prevalent in the UK and former colonies), there is also a clause classification task.
%\pr{Justify why it is relevant to discuss here, do you want to do the same? does it inspire your work in some ways, in terms of tasks, languages, law systems}
For a more comprehensive overview of the many tasks related to automated legal text analysis, we recommend reading the works of \citet{chalkidis-etal-2022-lexglue} and \citet{zhong-etal-2020-nlp}.

\subsection{Legal Language Models}
Several works have proposed legal language models (models specifically trained for the legal domain) for several languages other than English.
%Concerning language models specifically trained for the legal domain, many have been proposed for specific languages but, to the best of our knowledge, no multilingual model has been proposed yet. 
For example, legal language models for English~\cite{chalkidis-etal-2020-legal,Ying.Habernal.2022.NLLP,chalkidis-etal-2023-lexfiles}, French~\cite{douka-etal-2021-juribert}, Romanian~\cite{masala-etal-2021-jurbert}, Italian~\cite{lamberta,itlegalbert}, Chinese~\cite{lawformer}, Arabic~\cite{arabert}, Korean~\cite{hwang2022a}, and Portuguese~\cite{BertBR}.
Recently, pre-trained multilingual legal language models~\cite{niklaus2023multilegalpile,rasiah2023scale} have been released. Unfortunately, these models were not available at the time of submission, so we do not present results as part of this work.

% More detailed information regarding specific languages and tasks can be found in the surveys of \citet{survey-greek-nllp}, \citet{survey-judgmentprediction}, ...

\section{LEXTREME Datasets and Tasks}

\subsection{Dataset and Task Selection}
%\pr{data selection}

% We formulate several criteria to find relevant datasets for the LEXTREME benchmark.
To find relevant datasets for the LEXTREME benchmark we explore the literature of NLP and the legal domain,
% via the ACL anthology.
% AG: ICAIL for example is not in ACL anthology
%\footnote{\url{https://aclanthology.org/}}  
identifying relevant venues such as ACL, EACL, NAACL, EMNLP, LREC, ICAIL, and the NLLP workshop.
We search the literature on these venues for the years 2010 to 2022. We search for some common keywords (case insensitive) that are related to the legal domain, e.g., \emph{criminal}, \emph{judicial}, \emph{judgment}, \emph{jurisdictions}, \emph{law},  \emph{legal}, \emph{legislation}, and dataset, e.g., \emph{dataset}, and \emph{corpus} vie their union.
These keywords help to select 108 potentially relevant papers.
Then, we formulate several criteria to select the datasets. Finally, three authors analyze the candidate papers and perform the selection. 
%analyze these papers based on the inclusion and exclusion criteria given below to ensure that they indeed propose a legal dataset.
We handled the disagreement between authors based on mutual discussion and the majority voting mechanism.

\begin{table}[ht]
\centering
\resizebox{\columnwidth}{!}{
\begin{tabular}{llll}
\toprule
\textbf{Task} &     \textbf{\# Examples} & \textbf{\# Labels} \\
\midrule
 BCD-J &         3234 / 404 / 405 &               3 / 3 / 3 \\
 BCD-U &         1715 / 211 / 204 &               2 / 2 / 2 \\
   GAM &      19271 / 2726 / 3078 &               4 / 4 / 4 \\
 GLC-V &      28536 / 9511 / 9516 &            47 / 47 / 47 \\
 GLC-C &      28536 / 9511 / 9516 &         386 / 377 / 374 \\
 GLC-S &      28536 / 9511 / 9516 &      2143 / 1679 / 1685 \\
   SJP &     59709 / 8208 / 17357 &               2 / 2 / 2 \\
OTS-UL &         2074 / 191 / 417 &               3 / 3 / 3 \\
OTS-CT &      19942 / 1690 / 4297 &               9 / 8 / 9 \\
   C19 &         3312 / 418 / 418 &               8 / 8 / 8 \\
 MEU-1 & 817239 / 112500 / 115000 &            21 / 21 / 21 \\
 MEU-2 & 817239 / 112500 / 115000 &         127 / 126 / 127 \\
 MEU-3 & 817239 / 112500 / 115000 &         500 / 454 / 465 \\
   GLN &      17699 / 4909 / 4017 &            17 / 17 / 17 \\
   LNR &         7552 / 966 / 907 &             11 / 9 / 11 \\
  LNB &       7828 / 1177 / 1390 &            13 / 13 / 13 \\
 MAP-C &     27823 / 3354 / 10590 &            13 / 11 / 11 \\
 MAP-F &     27823 / 3354 / 10590 &            44 / 26 / 34 \\

\bottomrule
\end{tabular}
}
\caption{Dataset and task overview. \textit{\# Examples} and \textit{\# Labels} show values for train, validation, and test splits. 
%For a detailed overview of for the language-specific subsets of each multilingual task, see the \vm{online repository\footnote{\protect URL Anonymized}}.
}
\label{tab:overview_dataset_split_labels}
\end{table}

%\vspace{3mm}
\paragraph{Inclusion criteria:}
\setlist{nolistsep}
\begin{enumerate}[label=I\arabic*:, start=1, itemsep=0em]
\item It is about legal text (e.g., patents are not considered part of the legal text)
%\item It performs legal tasks (e.g., judgment prediction) and not other linguistic tasks such as \ac{POS} tagging,
\item It performs legal tasks (e.g., judgment prediction) or NLU tasks on legal text in order to have datasets that understand or reason about the legal text, similar to LEXGLUE \cite{chalkidis-etal-2022-lexglue}
%It performs NLU tasks (e.g., information retrieval tasks are not considered due to their evaluation complexity),
\item The current tasks are set in a European language, as per the scope of our present work, but we aim to incorporate a broader range of languages in future iterations of LEXTREME
%(e.g., China has its own large legal NLP community and likely would not benefit much from multilingual models), 
\item The dataset is annotated by humans directly or indirectly (e.g., judgement labels are extracted with regexes)
\end{enumerate}

\paragraph{Exclusion criteria:}
\begin{enumerate}[label=E\arabic*:, start=1, itemsep=0em]
  \item The dataset is not publicly available
  \item The dataset does not contain a public license or does not allow data redistribution
  \item The dataset contains labels generated with ML systems
  \item It is not a peer-reviewed paper
  %\pr{we have some pre-print datasets}.
\end{enumerate}

After applying the above criteria, we select 11 datasets from 108 papers. We provide the list of all these datasets in our repository.
%\footnote{\url{https://github.com/JoelNiklaus/LEXTREME}}

\begin{table}[t]
  \footnotesize
\resizebox{\columnwidth}{!}{%
\begin{tabular}{lp{3cm}p{2cm}}
\toprule
\textbf{Dataset} &                 \textbf{Jurisdiction} &                                                                         \textbf{Languages} \\
\midrule
             BCD & BR & pt \\
             GAM & DE & de \\             
             GLC & GR & el \\
             SJP & CH &     de, fr, it \\
             OTS & EU & de, en, it, pl \\
             C19 & BE, FR, HU, IT, NL, PL, UK & en, fr, hu, it, nb, nl, pl \\
             MEU & EU & 24 EU langs \\
             GLN & GR & el \\
             LNR & RO & ro \\
             LNB & BR & pt \\
             MAP & EU & 24 EU langs \\
\bottomrule
\end{tabular}
}
\caption{Overview of datasets, the jurisdiction, and the languages. The 24 EU languages are: bg, cs, da, de, el, en, es, et, fi, fr, ga, hu, it, lt, lv, mt, nl, pt, ro, sk, sv.}
\label{tab:datasets_general_overview_juris_languages}
\end{table}

\subsection{LEXTREME Datasets}
In the following, we briefly describe the selected datasets. \autoref{tab:overview_dataset_split_labels} provides more information about the number of examples and label classes per split for each task. For a detailed overview of the jurisdictions as well as the number of languages covered by each dataset, see \autoref{tab:datasets_general_overview_juris_languages}.
Each dataset %can be either monolingual or multilingual and 
can have several configurations (tasks), which are the basis of our analyses, i.e., the pre-trained models have always been fine-tuned on a single task. LEXTREME consists of three task types: \ac{SLTC}, \ac{MLTC}, and \ac{NER}. We use the existing train, validation, and test splits if present. Otherwise, we split the data randomly ourselves (80\% train, 10\% validation, and 10\% test). 

\paragraph{Brazilian Court Decisions (BCD).}  
Legal systems are often huge and complex, and the information is scattered across various sources. Thus, predicting case outcomes from multiple vast volumes of litigation is a difficult task. \citet{lage2022predicting} propose an approach to predict Brazilian legal decisions to support legal practitioners.  
We use their dataset from the State Supreme Court of Alagoas (Brazil). 
% The cases are split 80\% training (3234), 10\% validation (404), 10\% testing (405).
%They used ML algorithms, including classifiers, such as GNB, DT, SVM, RF, and XGBoost, and deep learning models, such as BERT, LSTM, GRU, BiLSTM, and CNN. They performed the 5-fold cross-validation approach and used 80\% of the total dataset for training and 20\% for the test dataset. They provided a prototype with 80.2\% F1-score performance on 4\,043 cases from a Brazilian court. 
The input to the models is always the case description. We perform two \ac{SLTC} tasks: In the BCD-J subset models predict the approval or dismissal of the case or appeal with the three labels \textit{no, partial, yes}, and in the BCD-U models predict the judges' unanimity on the decision alongside two labels, namely \textit{unanimity, not-unanimity}.  

\paragraph{German Argument Mining (GAM).}  Identifying arguments in court decisions is vital and challenging for legal practitioners. \citet{Urchs_et_all2021} assembled a dataset of 200 German court decisions for sentence classification based on argumentative function. We utilize this dataset for a \ac{SLTC} task. Model input is a sentence, and output is categorized as \textit{conclusion}, \textit{definition}, \textit{subsumption}, or \textit{other}.
%\emph{conclusion}, \emph{definition}, \emph{subsumption}, \emph{other}
%We split the dataset into 80\% train, 10\% validation, and 10\% test.. 

\paragraph{Greek Legal Code (GLC).} 
Legal documents can cover a wide variety of topics, which makes accurate topic classification all the more important. \citet{Papa2021m} compiled a dataset for topic classification of Greek legislation documents. The documents cover 47 main thematic topics which are called \textit{volumes}. Each of them is divided into thematic sub categories which are called \textit{chapters} and subsequently, each chapter breaks down to \textit{subjects}. Therefore, the dataset is used to perform three different \ac{SLTC} tasks along volume level (GLC-V), chapter level (GLC-C), and subject level (GLC-S). The input to the models is the entire document, and the output is one of the several topic categories. 
%The dataset is split into 60\% train, 20\% validation, and 20\% test.

\paragraph{Swiss Judgment Prediction (SJP).} 
\citet{Nikl2021s,niklaus-etal-2022-empirical}, focus on predicting the judgment outcome of 85K cases from the Swiss Federal Supreme Court (FSCS). The input to the models is the appeal description, and the output is whether the appeal is approved or dismissed (SLTC task). %It is also a \ac{SLTC} task.

\paragraph{Online Terms of Service (OTS).} 
While multilingualism's benefits (e.g., cultural diversity) in the EU legal world are well-known \cite{european2005new}, creating an official version of every legal act in 24 languages raises interpretative challenges. \citet{draw2021c} attempt to automatically detect unfair clauses in terms of service documents. We use their dataset of 100 contracts to perform a \ac{SLTC} and \ac{MLTC} task. For the \ac{SLTC} task (OTS-UL), model inputs are sentences, and outputs are classifications into three unfairness levels: \emph{clearly fair}, \emph{potentially unfair} and \emph{clearly unfair}. The \ac{MLTC} task (OTS-CT) involves identifying sentences based on nine clause topics.

\paragraph{COVID19 Emergency Event (C19).} 
The COVID-19 pandemic showed various exceptional measures governments worldwide have taken to contain the virus.
\citet{tzia2021}, presented a dataset, also known as EXCEPTIUS, that contains legal documents with sentence-level annotation from several European countries to automatically identify the measures. We use their dataset to perform the \ac{MLTC} task of identifying the type of measure described in a sentence. The input to the models are the sentences, and the output is neither or at least one of the measurement types.
%The examples for italian are split into 72\% training, 9\% validation, 19\% testing.

\paragraph{MultiEURLEX (MEU).} 
Multilingual transfer learning has gained significant attention recently due to its increasing applications in NLP tasks. \citet{chalkidis-etal-2021-multieurlex} explored the cross-lingual transfer for legal NLP and presented a corpus of 65K EU laws annotated with multiple labels from the EUROVOC taxonomy. We perform a \ac{MLTC} task to identify labels (given in the taxonomy) for each document. Since the taxonomy exists on multiple levels, we prepare configurations according to three levels (MEU-1, MEU-2, MEU-3).

\paragraph{Greek Legal NER (GLN).} 
Identifying various named entities from natural language text plays an important role for \ac{NLU}. \citet{Angelidis2018NamedER} compiled an annotated dataset for \ac{NER} in Greek legal documents. The source material are 254 daily issues of the Greek Government Gazette over the period 2000-2017. In \textit{all} \ac{NER} tasks of LEXTREME the input to the models is the list of tokens, and the output is an entity label for each token.

\begin{table*}
  \centering
  \footnotesize
  \resizebox{\textwidth}{!}{
  \begin{tabular}{llrrrrrr}
    \toprule
    \textbf{Model} & \textbf{Source} & \textbf{\# Parameters} & \textbf{Vocab} & \textbf{\# Steps} & \textbf{Batch Size}  & \textbf{Corpus} & \textbf{\# Langs} \\
    \midrule
    MiniLM              & \citet{Wang2020}                                   & 118M & 250K & 1M & 256       & 2.5TB CC100 & 100\\
    DistilBERT          & \citet{Sanh2019DistilBERTAD}                               & 135M & 120K & n/a & < 4000           & Wikipedia   & 104\\
    mDeBERTa-v3         & \citet{He2021a, He2021b}                 & 278M & 128K & 500K & 8192    & 2.5TB CC100 & 100\\
    XLM-R\textsubscript{Base/Large}        & \citet{conneau-etal-2020-unsupervised}       & 278M/560M & 250K & 1.5M & 8192    & 2.5TB CC100 & 100\\
    %mT5\textsubscript{Small}   & \citet{xue_mt5_2021}                     & 300M  &   250K    &   1M & 1024 & mC4 (CC)   & 101\\
    %mT5\textsubscript{Small/Base/Large}   & \citet{xue_mt5_2021}                     & 300M/580M/1.2B  &   250K    &   1M & 1024 & mC4 (CC)   & 101\\
    \bottomrule
  \end{tabular}
  }
  \caption{Multilingual models. All models support a maximum sequence length of 512 tokens. The third column shows the total number of parameters, including the embedding layer.
  }
  \label{tab:multilingual_models}
  %\vspace{-3mm}
\end{table*}

\paragraph{LegalNERo (LNR).} 
Similar to GLN, \citet{pais_vasile_2021_4922385} manually annotated Romanian legal documents for various named entities. The dataset is derived from 370 documents from the larger MARCELL Romanian legislative subcorpus.\footnote{\href{https://marcell-project.eu/deliverables.html}{https://marcell-project.eu/deliverables.html}}

\paragraph{LeNER BR (LNB).} 
\citet{luz2018a} compiled a dataset for \ac{NER} for Brazilian legal documents. 66 legal documents from several Brazilian Courts and four legislation documents were collected, resulting in a total of 70 documents annotated for named entities.

\paragraph{MAPA (MAP).} 
\citet{de2022s} built a multilingual corpus based on EUR-Lex \cite{baisa-etal-2016-european} for NER annotated at a coarse-grained (MAP-C) and fine-grained (MAP-F) level.
%The structure of the dataset is the same as the other datasets for \ac{NER}.

\section{Models Considered}
Since our benchmark only contains \ac{NLU} tasks, we consider encoder-only models for simplicity. Due to resource constraints, we did not fine-tune models larger than 1B parameters.

\subsection{Multilingual}
We considered the five multilingual models listed in \autoref{tab:multilingual_models},
% , MiniLM \cite{wang_minilm_2020}, DistilBERT \cite{sanh_distilbert_2020}, mDeBERTa-v3 \cite{he_debertav3_2021}, XLM-R \cite{conneau_unsupervised_2020}, and mT5 \cite{xue_mt5_2021}, 
trained on at least 100 languages each (more details are in Appendix~\ref{sec:model_descriptions}). For XLM-R we considered both the base and large version.
%For the time being we only considered encoder-based models.
%For mT5 we evaluated the small version, but in future work we will also evaluate the base and large sizes. We also considered BLOOM \cite{scao_bloom_2022}, but since many languages of LEXTREME are not in the training data, we did not evaluate it.
Furthermore, we used ChatGPT (gpt-3.5-turbo) for zero-shot evaluation of the text classification tasks with less than 50 labels.\footnote{We excluded MultiEurlex because it only contains numeric labels and not textual ones, and because the inputs are very long in 24 languages rendering a valid comparison with reasonable costs impossible.} To be fair across tasks we did not consider few-shot evaluation or more sophisticated prompting techniques because of prohibitively long inputs in many tasks.

\subsection{Monolingual}
In addition to the multilingual models, we also fine-tuned available monolingual models on the language specific subsets.
We chose monolingual models only if a certain language was represented in at least three datasets.\footnote{Which is why we did not include Norwegian pre-trained models, even though Norwegian is covered in C19.}
We made a distinction between general purpose models, i.e., models that have been pre-trained on generic data aka \emph{NativeBERTs}, and legal models, i.e., models that have been trained (primarily) on legal data aka \emph{NativeLegalBERTs}. A list of the monolingual models can be found in the appendix in \autoref{tab:overview_monolingual_models}.

\subsection{Hierarchical Variants}
\label{sec:hierarchical_variants}
A significant part of the datasets consists of very long documents, the best examples being all variants of MultiEURLEX; we provide detailed using different tokenizers on all datasets in our online repository.
% AG: for now, no need to put all these footnotes
%\footnote{\url{https://github.com/JoelNiklaus/LEXTREME/blob/main/visualizations/histograms}}
However, the models we evaluated were all pre-trained with a maximum sequence length of only 512 tokens. 
Directly applying pre-trained language models on lengthy legal documents may necessitate substantial truncation, severely restricting the models. To overcome this limitation, we use hierarchical versions of pre-training models for datasets containing long documents.
%It is possible to use the models without further ado for documents that exceed this length by far. However, this can only be achieved by a massive truncation of the original document. This procedure has the consequence that only the first section of a document is available for classification tasks. This is the reason why we used hierarchical variants of pre-training models for fine-tuning on data sets with particularly long documents.
%(cf. \href{https://github.com/JoelNiklaus/LEXTREME/blob/main/visualizations/histograms/Histograms_for_datasets_with_hierarchical_models.jpg}{histograms}). 

The hierarchical variants used here are broadly equivalent to those in \cite{chalkidis-etal-2021-paragraph, niklaus-etal-2022-empirical}. First, we divide each document into chunks of 128 tokens each. Second, we use the model to be evaluated to encode each of these paragraphs in parallel and to obtain the [CLS] embedding of each chunk which can be used as a context-unaware chunk representation. In order to make them context-aware, i.e. aware of the surrounding chunks, the chunk representations are fed into a 2-layered transformer encoder. Finally, max-pooling over the context-aware paragraph representations is deployed, which results in a document representation that is fed to a classification layer. 

Unfortunately, to the best of our knowledge models capable of handling longer context out of the box, such as Longformer \cite{DBLP:journals/corr/abs-2004-05150} and SLED \cite{ivgi-etal-2023-efficient} are not available multilingually and predominantly trained on English data only.

\section{Experimental Setup}

%In our experimental setup, we aimed to compare the performance of multilingual and monolingual pre-trained models on monolingual and multilingual datasets. To achieve this, we selected a set of multilingual and monolingual datasets and fine-tuned the multilingual models on all of them. In contrast, the monolingual models were only fine-tuned on those datasets that contained examples in the language the monolingual model had been pre-trained on (e.g., no Greek model fine-tuned on Swiss Judgment Prediction, not containing any Greek examples). To ensure a match between the language of the dataset and the language of the monolingual model, before fine-tuning, we applied a filter on the multilingual dataset, keeping only those examples that matched the target language of the pre-trained model. For example, when fine-tuning Italian models on Swiss Judgment Prediction, we considered only Italian examples for training and evaluation. 

%This experimental setup had the consequence that monolingual models had less data for fine-tuning on multilingual datasets than multilingual models. This becomes especially evident for smaller datasets, such as C19, which, as \citet{tzia2021} states, has a "\textit{very limited size and number of positive class examples}" for Hungarian, cf. Table \ref{subset_for_language_specific_overview_num_ex}. This limits the performance of the Hungarian monolingual model on this dataset, cf. Table \ref{dataset_language_aggregate_for_monolingual_models}. 

Multilingual models were fine-tuned on all languages of specific datasets. Monolingual models used only the given model's language subset.

Some datasets are highly imbalanced, one of the best examples being BCD-U with a proportion of the minority class of about 2\%. Therefore, we applied random oversampling on all SLTC datasets, except for GLC, since all its subsets have too many labels, which would have led to a drastic increase in the data size and thus in the computational costs for fine-tuning. For each run, we used the same hyperparameters, as described in Section \hyperref[sec:hyperparameters]{A.3}. 

As described in Section \ref{sec:hierarchical_variants}, some tasks contain very long documents, requiring the usage of hierarchical variants to process sequence lenghts of 1024 to 4096 tokens. Based on the distribution of the sequence length per example for each task (cf. \autoref{sec:histograms}), we decided on suitable sequence lengths for each task before fine-tuning. A list of suitable sequence lengths are in \hyperref[sec:max_seq_length]{A.1}. 
%Tasks with a maximum sequence length of over 512 required the usage of hierarchical variants.

%\vm{Mention for NER: IOB2 scheme for all tasks and seqeval strict mode for evaluation}

\subsection{Evaluation Metrics.}

We use the macro-F1 score for all datasets to ensure comparability across the entire benchmark, since it can be computed for both text classification and NER tasks. Mathew's Correlation Coefficient (MCC) \citep{MATTHEWS1975442} is a suitable score for evaluating text classification tasks but its applicability to NER tasks is unclear. For brevity, we do not display additional scores, but more detailed (such as precision and recall, and scores per seed) and additional scores (such as MCC) can be found online on our Weights and Biases project.\footnote{\url{https://wandb.ai/lextreme/paper_results}}

% TODO: Maybe expand MCC

\begin{table*}[ht]
  \centering
  \footnotesize
  \resizebox{\textwidth}{!}{
    \begin{tabular}{lccccccccccccc}
        \toprule
                    \textbf{Model} & \textbf{BCD} & \textbf{GAM} & \textbf{GLC} & \textbf{SJP} & \textbf{OTS} & \textbf{C19} & \textbf{MEU} & \textbf{GLN} & \textbf{LNR} & \textbf{LNB} & \textbf{MAP} & \textbf{Agg.} \\
        \midrule
                            MiniLM &         52.0 &     \bf 73.3 &         12.3 &         67.7 &         21.8 &          4.5 &         12.2 &         43.5 &         46.4 &         86.1 &         52.9 &          19.9 \\
                        DistilBERT &         53.7 &         69.5 &         53.4 &         66.8 &         52.4 &         21.2 &         23.2 &         38.1 &         48.0 &         78.7 &         53.0 &          43.2 \\
                       mDeBERTa v3 &         59.1 &         71.3 &         26.5 &         69.1 &         63.7 &     \bf 26.4 &         24.7 &         44.8 &         46.7 &         87.3 &     \bf 58.6 &          44.1 \\
         XLM-R\textsubscript{Base} &     \bf 62.6 &         71.9 &         42.1 &     \bf 69.3 &         64.6 &         18.4 &         11.4 &     \bf 46.4 &         45.6 &         87.3 &         53.2 &          36.8 \\
        XLM-R\textsubscript{Large} &         58.0 &         73.1 &     \bf 71.7 &         68.9 &     \bf 73.8 &         20.6 &     \bf 27.9 &         45.1 &     \bf 55.4 &     \bf 88.4 &         55.1 &      \bf 48.5 \\
        \bottomrule
        \end{tabular}
      }
    \caption{Dataset aggregate scores for multilingual models. The best scores are in bold.}
    \label{dataset_aggregate_multilingual_models}
% \vspace{-3mm}
\end{table*}

\begin{table*}[ht]
\setlength{\tabcolsep}{3pt}
  \centering
  \footnotesize
  \resizebox{\textwidth}{!}{
    \begin{tabular}{lccccccccccccccccccccccccc}
\toprule
            \textbf{Model} & \textbf{bg} & \textbf{cs} & \textbf{da} & \textbf{de} & \textbf{el} & \textbf{en} & \textbf{es} & \textbf{et} & \textbf{fi} & \textbf{fr} & \textbf{ga} & \textbf{hr} & \textbf{hu} & \textbf{it} & \textbf{lt} & \textbf{lv} & \textbf{mt} & \textbf{nl} & \textbf{pl} & \textbf{pt} & \textbf{ro} & \textbf{sk} & \textbf{sl} & \textbf{sv} & \textbf{Agg.} \\
\midrule
                    MiniLM &        20.9 &        20.4 &        19.8 &        27.6 &        19.0 &         8.2 &        21.2 &        19.9 &        19.7 &        15.9 &        40.2 &        11.9 &        20.3 &        15.1 &        20.3 &        20.3 &        14.9 &        20.5 &        15.2 &        30.5 &        25.9 &        20.2 &        12.4 &        20.5 &          18.1 \\
                DistilBERT &        33.7 &        32.7 &        32.1 &        47.4 &        35.1 &        38.0 &        36.1 &        31.0 &        30.8 &        38.8 &        43.8 &        22.5 &        31.5 &        41.0 &        31.7 &        31.6 &        29.9 &        19.0 &        25.0 &        44.5 &        37.6 &        32.0 &        22.9 &        33.3 &          31.9 \\
               mDeBERTa v3 &        34.6 &        34.4 &        33.6 &        49.9 &        33.5 &        41.0 &        36.6 &        34.6 &        33.9 &        39.8 &    \bf 49.4 &        24.7 &        35.6 &        44.5 &        34.9 &        35.0 &        33.4 &        24.5 &        29.2 &        46.3 &        39.5 &        35.6 &        24.8 &        35.9 &          34.8 \\
 XLM-R\textsubscript{Base} &        19.9 &        19.4 &        18.8 &        33.3 &        25.6 &        27.6 &        19.4 &        18.8 &        18.6 &        27.3 &        44.9 &        11.6 &        13.4 &        31.0 &        18.7 &        18.9 &        16.0 &        15.2 &        21.4 &        30.3 &        24.1 &        18.9 &        11.7 &        19.4 &          19.7 \\
XLM-R\textsubscript{Large} &    \bf 38.5 &        37.9 &        38.0 &        51.5 &        44.1 &        44.7 &        39.7 &        36.9 &        35.3 &        42.1 &        48.6 &    \bf 28.1 &        22.7 &    \bf 48.0 &    \bf 37.4 &    \bf 37.9 &    \bf 34.1 &        19.3 &    \bf 32.7 &        48.9 &        42.3 &    \bf 37.1 &    \bf 28.0 &        37.0 &      36.0 \\
           NativeLegalBERT &      -       &       -      &      -       &       -      &        -     &        43.8 &    \bf 40.3 &      -       &       -      &       -      &       -      &      -       &      -       &        34.0 &      -       &        -     &      -       &      -       &      -       &     -        &        38.8 &       -      &     -        &      -       &        38.9       \\
                NativeBERT &        24.3 &    \bf 47.4 &    \bf 42.8 &    \bf 56.0 &    \bf 47.9 &    \bf 49.4 &        33.3 &    \bf 38.3 &    \bf 43.2 &    \bf 43.5 &        44.0 &    -         &    \bf 45.4 &        42.5 &       -     &       -      &      -       &    \bf 36.2 &        21.6 &    \bf 54.9 &    \bf 44.4 &        29.1 &    -         &    \bf 46.1 &        \bf 39,1       \\
\bottomrule
\end{tabular}
    }
    \caption{Language aggregate scores for multilingual models. The best scores are in bold. For each language, we also list the best-performing monolingual legal model under \textit{NativeLegalBERT} and the best-performing monolingual non-legal model under \textit{NativeBERT}. Missing values indicate that no suitable models were found.
    %NativeBERT and NativeLegalBERT denote the best monolingual models trained on general and legal corpora respectively.
    }
    \label{language_aggregate_multilingual_models}
  %\vspace{-3mm}
\end{table*}

\subsection{Aggregate Score}

%Thoughs on Metrics: 
%2 Options: 
%1. Macro-F1 everywhere: 
%+ simplicity
%+ comparability
%- not best metric for multiclass tc
%
%2. individual metrics
%Multiclass: MCC
%Multilabel: Macro-F1
%NER: Macro-F1
%+ better metric for multiclass
%- worse comparability
%- complexity
%
%==> The work is complex enough, we choose Macro-F1 %everywhere

We acknowledge that the datasets included in LEXTREME are diverse and hard to compare due to variations in the number of samples and task complexity \cite{WWWB}. This is why we always report the scores for each dataset subset, enabling a fine-grained analysis. However, we believe that by taking the following three measures, an aggregate score can provide more benefits than drawbacks, encouraging the community to evaluate multilingual legal models on a curated benchmark, thus easing comparisons. 

We (a) evaluate all datasets with the same score (macro-F1) making aggregation more intuitive and easier to interpret, (b) aggregating the F1 scores again using the harmonic mean, since F1 scores are already rates and obtained using the harmonic mean over precision and recall, following \citet{shavrina2021not}, 
and (c) basing our final aggregate score on two intermediate aggregate scores –– the dataset aggregate and language aggregate score -- thus weighing datasets and languages equally promoting model fairness, following \citet{shavrina2021not} and \citet{chalkidis-etal-2022-lexglue}.

The final LEXTREME score is computed using the harmonic mean of the dataset and the language aggregate score. We calculate the dataset aggregate by successively taking the harmonic mean of (i) the languages in the configurations (e.g., de,fr,it in SJP), (ii) configurations within datasets (e.g., OTS-UL, OTS-CT in OTS), and (iii) datasets in LEXTREME (BCD, GAM).
The language aggregate score is computed similarly: by taking the harmonic mean of (i) configurations within datasets, (ii) datasets for each language (e.g., MAP, MEU for lv), and (iii) languages in LEXTREME (bg,cs).

We do not address the dimension of the jurisdiction, which we consider beyond the scope of this work.

% Dataset Aggregate Score
% 1. Average over languages inside configurations, 2. Average over configurations (e.g. judgment and unanimity) inside datasets, 3. Average over Datasets

% Language Aggregate Score
% 1. Average over configurations (e.g. judgment and unanimity) inside datasets, 2. Average over Datasets, 3. Average over languages

%\todo{Maybe write formulas here for the aggregate scores?}
% Harmonic Mean: 
% $\frac{1}{n} \sum\limits_{i=1}^n \frac{1}{x_i}$

%\subsection{Considered Efficiency Score}
%\todo{Encourage people to use method by Show Your Work: Improved Reporting of Experimental Results paper for efficiency metric}

%\todo{Mentions Survey efficient methods for NLP (section 8): https://arxiv.org/abs/2209.00099.pdf. "Properly measuring efficiency still remains a big challenge."}

\section{Results}

% \jn{Write more about bad performance of MiniLM}

In this section, we discuss baseline evaluations. Scores and standard deviations for validation and test datasets across seeds are on our Weights and Biases project or can be found in Table
\ref{tab:arithmetic_mean_multilingual_models_eval_set}, \ref{tab:arithmetic_mean_multilingual_models_test_set}, \ref{tab:arithmetic_mean_monolingual_models_eval_set}, \ref{tab:arithmetic_mean_monolingual_models_test_set}. %\footnote{URL Anonymized}
Comparisons with prior results on each dataset can be drawn from the tables provdided in section \ref{sec:paper_results} in the appendix. Aggregated results by dataset and language are in Tables~\ref{dataset_aggregate_multilingual_models}~and~\ref{language_aggregate_multilingual_models}.

\paragraph{Larger models are better} For both, we see a clear trend that larger models perform better. However, when looking at the individual datasets and languages, the scores are more erratic. Note that XLM-R\textsubscript{Base} underperforms on MEU (especially on MEU-3; see \autoref{tab:arithmetic_mean_multilingual_models_eval_set} and \autoref{tab:arithmetic_mean_multilingual_models_test_set}) leading to a low dataset aggregate score due to the harmonic mean. Additionally, low performance on MEU-3 has a large impact on its language aggregate score, since it affects all 24 languages. 

% on some datasets models vary a lot, on others less
\paragraph{Differing model variance across datasets}
We observe significant variations across datasets such as GLC, OTS or C19, with differences as large as 52 (in OTS) between the worst-performing MiniLM and the best-performing XLM-R large. MiniLM seems to struggle greatly with these three datasets, while even achieving the best performance on GAM. On other datasets, such as GAM, SJP, and MAP the models are very close together (less than 6 points between best and worst model). Even though XLM-R\textsubscript{Large} takes the top spot on aggregate, it only has the best performance in six out of eleven datasets. 
%SJP, MEU and MAP are the largest datasets in LEXTREME, thus probably decreasing the influence of the pre-training on downstream performance and leveling the playing field. LNR, however, is the smallest \ac{NER} task, opposing this hypothesis.

\paragraph{Less variability across languages}
In contrast to inconsistent results on the datasets, XLM-R\textsubscript{Large} outperforms the other multilingual models on most languages (21 out of 24). Additionally, we note that model variability within a language is similar to the variability within a dataset, however, we don't see extreme cases such as GLC, OTS, or C19.

\paragraph{Monolingual models are strong}
Monolingual general-purpose models (NativeBERT in Table~\ref{language_aggregate_multilingual_models}) show strong performance with only a few exceptions (on Bulgarian, Spanish, Polish, and Slovak). In 13 out of 19 available languages they reach the top performance, leading to the top language aggregate score.
%Monolingual general-purpose models (NativeBERT in Table~\ref{language_aggregate_multilingual_models}) perform mostly on par with similarly sized multilingual models such as XLM-R and mDeBERTa with some exceptions. The monolingual Spanish and Italian models underperform whereas the Czech and Greek models outperform and even achieve the best language-specific scores overall. We hypothesize that the tokenizer and training data of multilingual models still tend to represent lower resource languages such as Czech and Greek worse than higher resource languages such as Spanish and Italian. This could be an advantage for monolingual models, that can allocate the entire vocabulary to one language only. 
The few available models pre-trained on legal data (NativeLegalBERT) slightly outperform multilingual models of the same size.

\begin{table}[t]
\centering
\resizebox{0.8\columnwidth}{!}{%
\begin{tabular}{lrr}
\toprule
\bf{Task} &  \bf{XLM-R\textsubscript{Large}} &  \bf{ChatGPT} \\
\midrule
BCD-J  &  \bf 58.1 &  52.1 \\
BCD-U  &  \bf 70.4 &  48.2 \\
GAM    &  \bf 73.0 &  35.5 \\
GLC-V  &  \bf 58.2 &  32.9 \\
SJP    &  \bf 60.9 &  51.2 \\
OTS-UL &  \bf 79.8 &  15.1 \\
OTS-CT &  \bf 64.5 &  12.7 \\
C19    &  \bf 27.7 &  23.6 \\
\bottomrule
\end{tabular}
}
\caption{Results with ChatGPT on the validation sets performed on June 15, 2023.
%To save costs, we limited the evaluation size to 1000 samples for ChatGPT.
Best results are in \textbf{bold}.}
\label{tab:chatgpt_results}
\end{table}

% legal models are worse than monolingual models
\paragraph{ChatGPT underperforms}
We show a comparison of ChatGPT with the best performing multilingual model XLM-R\textsubscript{Large} in \autoref{tab:chatgpt_results}.
To save costs, we limited the evaluation size to 1000 samples for ChatGPT.
We use the validation set instead of the test set to be careful not to leak test data into ChatGPT, possibly affecting future evaluation. \citet{chalkidis2023chatgpt} showed that ChatGPT is still outperformed by supervised approaches on LexGLUE. Similarly, we find that much smaller supervised models clearly outperform ChatGPT in all of tested tasks, with very large gaps in GAM and OTS.

\section{Conclusions and Future Work}

%\subsection{Answers to Research Questions}

\paragraph{Conclusions}
%In our work, we tried to capture a wide range of different tasks and languages, but it is important to remark that it is impossible for any single benchmark to capture the complexity of an entire research field~\cite{WWWB}.
We survey the literature and select 11 datasets out of 108 papers with rigorous criteria to compile the first multilingual benchmark for legal NLP. By open-sourcing both the dataset and the code, we invite researchers and practitioners to evaluate any future multilingual models on our benchmark. We provide baselines for five popular multilingual encoder-based language models of different sizes.
We hope that this benchmark will foster the creation of novel multilingual legal models and therefore contribute to the progress of natural legal language processing. We imagine this work as a living benchmark and invite the community to extend it with new suitable datasets.

\paragraph{Future Work}
% extend to other tasks
In future work, we plan to extend this benchmark with other \ac{NLU} tasks and also generation tasks such as summarization, simplification, or translation.
% extend with other languages/jurisdictions
% Additionally, extending datasets in more languages or jurisdictions, and evaluating other models such as mT5 \cite{xue_mt5_2021} can be other promising directions.
Additionally, a deeper analysis of the differences in the behavior of monolingual general-purpose models versus models trained on legal data could provide useful insights for the development of new models.
Another relevant aspect that deserves further studies is the impact of the jurisdiction and whether the jurisdiction information is predominantly learned as part of the LLM or is instead learned during fine-tuning. 
% Possibly, this could be analyzed by comparing the embedded representation of some keywords inside the LLMs in the different training settings.
%We believe that this research direction would require an entire paper of its own focused on these experiments, and therefore should be considered as a future research direction.

Finally, extending datasets in more languages and evaluating other models such as mT5 \cite{xue_mt5_2021} can be other promising directions.
% use other models
%Finally, we leave the evaluation of other models such as mT5 \cite{xue_mt5_2021} to future work. 

\section*{Acknowledgements}
% We thank the anonymous reviewers for their insightful comments. Blablabla
% Thank the TPU Research Cloud for providing us with TPU resources for pre-training the multilingual models
Joel Niklaus is funded by the Swiss National Research Programme “Digital Transformation” (NRP-77) grant number 187477.
Pooja Rani is funded by the Swiss National Science Foundation with Projects No. 200021\_197227.
Andrea Galassi is funded by the European Commission’s NextGeneration EU Programme, PNRR - M4C2 - Investimento 1.3, Partenariato Esteso PE00000013 - FAIR - Future Artificial Intelligence Research - Spoke 8 Pervasive AI. 
Ilias Chalkidis is funded by the Novo Nordisk Foundation (grant NNF 20SA0066568).

\section*{Limitations}
\label{sec:limits}

% \todo{Finish first paragraph}

% The creation of language models specific for a domain, but general enough to perform different tasks is .... Therefore, the creation of a novel benchmark designed to evaluate them on multiple tasks is ...
% Nonetheless, 
It is important to not exceed the enthusiasm for language models and the ambitions of benchmarks: many recent works have addressed the limits of these tools and analyzed the consequences of their misuse. 
For example, \citet{climbingNLU} argue that language models do not really learn ``meaning''. \citet{DBLP:conf/fat/BenderGMS21} further expand the discussion by addressing the risks related to these technologies and proposing mitigation methods.
\citet{life-dataset} evaluate the use of datasets inside scientific communities and highlight that many machine learning sub-communities focus on very few datasets and that often these dataset are ``borrowed'' from other communities.
\citet{WWWB} offer a detailed exploration of the limits of popular ``general'' benchmarks, such as GLUE \cite{GLUE} and ImageNET \cite{DBLP:conf/cvpr/DengDSLL009}. Their analysis covers 3 aspects: limited task design, de-contextualized data and performance reporting, inappropriate community use.

The first problem concerns the fact that typically tasks are not chosen considering proper theories and selecting what would be needed to prove generality. Instead, they are limited to what is considered interesting by the community, what is available, or other similar criteria.
These considerations hold also for our work. Therefore, we cannot claim that our benchmark can be used to assess the “generality” of a model or proving that it “understands natural legal language”.

The second point addresses the fact that any task, data, or metric are limited to their context, therefore ``data benchmarks are closed and inherently subjective, localized constructions''.
In particular, the content of the data can be too different from real data and the format of the tasks can be too homogeneous compared to human activities.
Moreover, any dataset inherently contains biases.
We tackle this limitation by deciding to include only tasks and data that are based on real world scenarios, in an effort to minimize the difference between the performance of a model on our benchmark and its performance on a real world problem.

%\agc{the I/O is not very diverse, but most of the tasks and datasets come from real world problems.} \\
% \agc{for the biases, maybe we can address this by pointing out for each task and dataset which are the limits of that specific dataset?}

The last aspect regards the negative consequences that benchmarks can have. The competitive testing may encourage misbehavior and the aggregated performance evaluation does create a mirage of cross-domain comparability. The presence of popular benchmarks can influence a scientific community up to the point of steering towards techniques that perform well on that specific benchmark, in disfavor of those that do not. Finally, benchmarks can be misused in marketing to promote commercial products while hiding their flaws.
These behaviours obviously cannot be forecasted in advance, but we hope that this analysis of the shortcomings of our work will be sufficient to prevent misuses of our benchmark and will also inspire research directions for complementary future works.
For what specifically concerns aggregated evaluations, they provide an intuitive but imprecise understanding of the performance of a model.  While we do not deny their potential downsides, we believe that their responsible use is beneficial, especially when compared to the evaluation of a model on only an arbitrarily selected set of datasets.
Therefore, we opted to provide an aggregated evaluation and to weigh languages and tasks equally to make it as robust and fair as possible.
%

% \agc{The point about ``the aggregated performance evaluation'' discourages us from proposing an aggregated metric. I don't think we should not use it, but if we do we MUST justify it here somehow.}

% We analyze the shortcomings of our work and suggest complementary aspects that we do not cover.

% We analyze potential shortcomings of our work following \citet{WWWB}, with the intent of preventing misuse of this work and of suggesting research directions for future benchmarks and datasets.

While \citeauthor{WWWB} and \citeauthor{life-dataset} argument against the misrepresentations and the misuses of benchmarks and datasets, they do not argue against their usefulness.
On the contrary, they consider the creation and adoption of novel benchmarks a sign of a healthy scientific community.

Finally, we want to remark that for many datasets the task of outcome prediction is based not on the document provided by the parties, but on the document provided by the judge along with its decision. For example, \citet{semo-etal-2022-classactionprediction} provide a more realistic setup of judgment prediction than other datasets, using actual complaints as inputs.
However, due to very limited access to the complaint documents, especially multilingually, creating such datasets is extremely challenging. Thus, most recent works used text from court decisions as proxies. However, predicting the judgment outcome based on text written by the court itself can still be a hard task (as evidenced by results on these datasets). Moreover, it may still require legal reasoning capabilities from models because of the need to pick out the correct information. Additionally, we believe that these tasks can also be interesting to conduct post hoc analyses of decisions.

\section*{Ethics Statement}

The scope of this work is to release a unified multi-lingual legal NLP benchmark to accelerate the development and evaluation of multilingual legal language models. A transparent multilingual and multinational benchmark for NLP in the legal domain might serve as an orientation for scholars and industry researchers by broadening the discussion and helping practitioners to build assisting technology for legal professionals and laypersons. We believe that this is an important application field, where research should be conducted \cite{tsarapatsanis-aletras-2021-ethical} to improve legal services and democratize law, while also highlight (inform the audience on) the various multi-aspect shortcomings seeking a responsible and ethical (fair) deployment of legal-oriented technologies.

Nonetheless, irresponsible use (deployment) of such technology is a plausible risk, as in any other application (e.g., online content moderation) and domain (e.g., medical). We believe that similar technologies should only be deployed to assist human experts (e.g., legal scholars in research, or legal professionals in forecasting or assessing legal case complexity) with notices on their limitations.

All datasets included in LEXTREME, are publicly available and have been previously published. We referenced the original work and encourage LEXTREME users to do so as well. In fact, we believe this work should only be referenced, in addition to citing the original work, when experimenting with multiple LEXTREME datasets and using the LEXTREME evaluation infrastructure. Otherwise, only the original work should be cited.

\bibliography{anthology,custom,references}

\begin{thebibliography}{98}
\expandafter\ifx\csname natexlab\endcsname\relax\def\natexlab#1{#1}\fi

\bibitem[{Al-Qurishi et~al.(2022)Al-Qurishi, AlQaseemi, and Souissi}]{arabert}
Muhammad Al-Qurishi, Sarah AlQaseemi, and Riad Souissi. 2022.
\newblock Aralegal-bert: A pretrained language model for arabic legal text.
\newblock In \emph{NLLP}.

\bibitem[{Aletras et~al.(2022)Aletras, Chalkidis, Barrett,
  Goan{\textcommabelow{t}}{\u{a}}, and
  Preo{\textcommabelow{t}}iuc-Pietro}]{nllp-2022-natural}
Nikolaos Aletras, Ilias Chalkidis, Leslie Barrett, C{\u{a}}t{\u{a}}lina
  Goan{\textcommabelow{t}}{\u{a}}, and Daniel
  Preo{\textcommabelow{t}}iuc-Pietro, editors. 2022.
\newblock \href {https://aclanthology.org/2022.nllp-1.0} {\emph{Proceedings of
  the Natural Legal Language Processing Workshop 2022}}. Association for
  Computational Linguistics, Abu Dhabi, United Arab Emirates (Hybrid).

\bibitem[{Angelidis et~al.(2018)Angelidis, Chalkidis, and
  Koubarakis}]{Angelidis2018NamedER}
Iosif Angelidis, Ilias Chalkidis, and Manolis Koubarakis. 2018.
\newblock Named entity recognition, linking and generation for greek
  legislation.
\newblock In \emph{{JURIX}}, volume 313 of \emph{Frontiers in Artificial
  Intelligence and Applications}, pages 1--10. {IOS} Press.

\bibitem[{Baisa et~al.(2016)Baisa, Michelfeit, Medve{\v{d}}, and
  Jakub{\'\i}{\v{c}}ek}]{baisa-etal-2016-european}
V{\'\i}t Baisa, Jan Michelfeit, Marek Medve{\v{d}}, and Milo{\v{s}}
  Jakub{\'\i}{\v{c}}ek. 2016.
\newblock \href {https://aclanthology.org/L16-1445} {{E}uropean {U}nion
  language resources in {S}ketch {E}ngine}.
\newblock In \emph{Proceedings of the Tenth International Conference on
  Language Resources and Evaluation ({LREC}'16)}, pages 2799--2803,
  Portoro{\v{z}}, Slovenia. European Language Resources Association (ELRA).

\bibitem[{Barry et~al.(2021)Barry, Wagner, Cassidy, Cowap, Lynn, Walsh,
  Meachair, and Foster}]{DCU-NLP/bert-base-irish-cased-v1}
James Barry, Joachim Wagner, Lauren Cassidy, Alan Cowap, Teresa Lynn, Abigail
  Walsh, M'iche'al J.~'O Meachair, and Jennifer Foster. 2021.
\newblock gabert — an irish language model.
\newblock In \emph{International Conference on Language Resources and
  Evaluation}.

\bibitem[{Beltagy et~al.(2020)Beltagy, Peters, and
  Cohan}]{DBLP:journals/corr/abs-2004-05150}
Iz~Beltagy, Matthew~E. Peters, and Arman Cohan. 2020.
\newblock Longformer: The long-document transformer.
\newblock \emph{CoRR}, abs/2004.05150.

\bibitem[{Bender et~al.(2021)Bender, Gebru, McMillan{-}Major, and
  Shmitchell}]{DBLP:conf/fat/BenderGMS21}
Emily~M. Bender, Timnit Gebru, Angelina McMillan{-}Major, and Shmargaret
  Shmitchell. 2021.
\newblock \href {https://doi.org/10.1145/3442188.3445922} {On the dangers of
  stochastic parrots: Can language models be too big?}
\newblock In \emph{FAccT '21: 2021 {ACM} Conference on Fairness,
  Accountability, and Transparency, Virtual Event / Toronto, Canada, March
  3-10, 2021}, pages 610--623. {ACM}.

\bibitem[{Bender and Koller(2020)}]{climbingNLU}
Emily~M. Bender and Alexander Koller. 2020.
\newblock \href {https://doi.org/10.18653/v1/2020.acl-main.463} {Climbing
  towards {NLU:} on meaning, form, and understanding in the age of data}.
\newblock In \emph{Proceedings of the 58th Annual Meeting of the Association
  for Computational Linguistics, {ACL} 2020, Online, July 5-10, 2020}, pages
  5185--5198. Association for Computational Linguistics.

\bibitem[{Brown et~al.(2020)Brown, Mann, Ryder, Subbiah, Kaplan, Dhariwal,
  Neelakantan, Shyam, Sastry, Askell, Agarwal, Herbert-Voss, Krueger, Henighan,
  Child, Ramesh, Ziegler, Wu, Winter, Hesse, Chen, Sigler, Litwin, Gray, Chess,
  Clark, Berner, McCandlish, Radford, Sutskever, and
  Amodei}]{brown_language_2020}
Tom~B. Brown, Benjamin Mann, Nick Ryder, Melanie Subbiah, Jared Kaplan,
  Prafulla Dhariwal, Arvind Neelakantan, Pranav Shyam, Girish Sastry, Amanda
  Askell, Sandhini Agarwal, Ariel Herbert-Voss, Gretchen Krueger, Tom Henighan,
  Rewon Child, Aditya Ramesh, Daniel~M. Ziegler, Jeffrey Wu, Clemens Winter,
  Christopher Hesse, Mark Chen, Eric Sigler, Mateusz Litwin, Scott Gray,
  Benjamin Chess, Jack Clark, Christopher Berner, Sam McCandlish, Alec Radford,
  Ilya Sutskever, and Dario Amodei. 2020.
\newblock \href {http://arxiv.org/abs/2005.14165} {Language {Models} are
  {Few}-{Shot} {Learners}}.
\newblock \emph{arXiv:2005.14165 [cs]}.
\newblock ArXiv: 2005.14165.

\bibitem[{Brugger et~al.(2023)Brugger, St\"{u}rmer, and
  Niklaus}]{brugger_multilegalsbd}
Tobias Brugger, Matthias St\"{u}rmer, and Joel Niklaus. 2023.
\newblock \href {https://doi.org/10.1145/3594536.3595132} {Multilegalsbd: A
  multilingual legal sentence boundary detection dataset}.
\newblock In \emph{Proceedings of the Nineteenth International Conference on
  Artificial Intelligence and Law}, ICAIL '23, page 42–51, New York, NY, USA.
  Association for Computing Machinery.

\bibitem[{Chalkidis(2023)}]{chalkidis2023chatgpt}
Ilias Chalkidis. 2023.
\newblock \href {https://doi.org/10.2139/ssrn.4385460} {Chatgpt may pass the
  bar exam soon, but has a long way to go for the lexglue benchmark}.
\newblock \emph{CoRR}, abs/2304.12202.

\bibitem[{Chalkidis et~al.(2021{\natexlab{a}})Chalkidis, Fergadiotis, and
  Androutsopoulos}]{chalkidis-etal-2021-multieurlex}
Ilias Chalkidis, Manos Fergadiotis, and Ion Androutsopoulos.
  2021{\natexlab{a}}.
\newblock \href {https://doi.org/10.18653/v1/2021.emnlp-main.559}
  {{M}ulti{EURLEX} - a multi-lingual and multi-label legal document
  classification dataset for zero-shot cross-lingual transfer}.
\newblock In \emph{Proceedings of the 2021 Conference on Empirical Methods in
  Natural Language Processing}, pages 6974--6996, Online and Punta Cana,
  Dominican Republic. Association for Computational Linguistics.

\bibitem[{Chalkidis et~al.(2020)Chalkidis, Fergadiotis, Malakasiotis, Aletras,
  and Androutsopoulos}]{chalkidis-etal-2020-legal}
Ilias Chalkidis, Manos Fergadiotis, Prodromos Malakasiotis, Nikolaos Aletras,
  and Ion Androutsopoulos. 2020.
\newblock \href {https://doi.org/10.18653/v1/2020.findings-emnlp.261}
  {{LEGAL}-{BERT}: The muppets straight out of law school}.
\newblock In \emph{Findings of the Association for Computational Linguistics:
  EMNLP 2020}, pages 2898--2904, Online. Association for Computational
  Linguistics.

\bibitem[{Chalkidis et~al.(2021{\natexlab{b}})Chalkidis, Fergadiotis,
  Tsarapatsanis, Aletras, Androutsopoulos, and
  Malakasiotis}]{chalkidis-etal-2021-paragraph}
Ilias Chalkidis, Manos Fergadiotis, Dimitrios Tsarapatsanis, Nikolaos Aletras,
  Ion Androutsopoulos, and Prodromos Malakasiotis. 2021{\natexlab{b}}.
\newblock \href {https://doi.org/10.18653/v1/2021.naacl-main.22}
  {Paragraph-level rationale extraction through regularization: A case study on
  {E}uropean court of human rights cases}.
\newblock In \emph{Proceedings of the 2021 Conference of the North American
  Chapter of the Association for Computational Linguistics: Human Language
  Technologies}, pages 226--241, Online. Association for Computational
  Linguistics.

\bibitem[{Chalkidis et~al.(2023)Chalkidis, Garneau, Goanta, Katz, and
  S{\o}gaard}]{chalkidis-etal-2023-lexfiles}
Ilias Chalkidis, Nicolas Garneau, Catalina Goanta, Daniel Katz, and Anders
  S{\o}gaard. 2023.
\newblock \href {https://doi.org/10.18653/v1/2023.acl-long.865} {{L}e{XF}iles
  and {L}egal{LAMA}: Facilitating {E}nglish multinational legal language model
  development}.
\newblock In \emph{Proceedings of the 61st Annual Meeting of the Association
  for Computational Linguistics (Volume 1: Long Papers)}, pages 15513--15535,
  Toronto, Canada. Association for Computational Linguistics.

\bibitem[{Chalkidis et~al.(2022{\natexlab{a}})Chalkidis, Jana, Hartung,
  Bommarito, Androutsopoulos, Katz, and Aletras}]{chalkidis-etal-2022-lexglue}
Ilias Chalkidis, Abhik Jana, Dirk Hartung, Michael Bommarito, Ion
  Androutsopoulos, Daniel Katz, and Nikolaos Aletras. 2022{\natexlab{a}}.
\newblock \href {https://doi.org/10.18653/v1/2022.acl-long.297} {{L}ex{GLUE}: A
  benchmark dataset for legal language understanding in {E}nglish}.
\newblock In \emph{Proceedings of the 60th Annual Meeting of the Association
  for Computational Linguistics (Volume 1: Long Papers)}, pages 4310--4330,
  Dublin, Ireland. Association for Computational Linguistics.

\bibitem[{Chalkidis et~al.(2022{\natexlab{b}})Chalkidis, Pasini, Zhang, Tomada,
  Schwemer, and Søgaard}]{chalkidis_fairlex_2022}
Ilias Chalkidis, Tommaso Pasini, Sheng Zhang, Letizia Tomada, Sebastian~Felix
  Schwemer, and Anders Søgaard. 2022{\natexlab{b}}.
\newblock \href {http://arxiv.org/abs/2203.07228} {{FairLex}: {A}
  {Multilingual} {Benchmark} for {Evaluating} {Fairness} in {Legal} {Text}
  {Processing}}.
\newblock \emph{arXiv:2203.07228 [cs]}.
\newblock ArXiv: 2203.07228.

\bibitem[{Chan et~al.(2020)Chan, Schweter, and M{\"o}ller}]{deepset/gbert-base}
Branden Chan, Stefan Schweter, and Timo M{\"o}ller. 2020.
\newblock \href {https://doi.org/10.18653/v1/2020.coling-main.598}
  {{G}erman{'}s next language model}.
\newblock In \emph{Proceedings of the 28th International Conference on
  Computational Linguistics}, pages 6788--6796, Barcelona, Spain (Online).
  International Committee on Computational Linguistics.

\bibitem[{Chowdhery et~al.(2022)Chowdhery, Narang, Devlin, Bosma, Mishra,
  Roberts, Barham, Chung, Sutton, Gehrmann, Schuh, Shi, Tsvyashchenko, Maynez,
  Rao, Barnes, Tay, Shazeer, Prabhakaran, Reif, Du, Hutchinson, Pope, Bradbury,
  Austin, Isard, Gur-Ari, Yin, Duke, Levskaya, Ghemawat, Dev, Michalewski,
  Garcia, Misra, Robinson, Fedus, Zhou, Ippolito, Luan, Lim, Zoph, Spiridonov,
  Sepassi, Dohan, Agrawal, Omernick, Dai, Pillai, Pellat, Lewkowycz, Moreira,
  Child, Polozov, Lee, Zhou, Wang, Saeta, Diaz, Firat, Catasta, Wei,
  Meier-Hellstern, Eck, Dean, Petrov, and Fiedel}]{chowdhery_palm_2022}
Aakanksha Chowdhery, Sharan Narang, Jacob Devlin, Maarten Bosma, Gaurav Mishra,
  Adam Roberts, Paul Barham, Hyung~Won Chung, Charles Sutton, Sebastian
  Gehrmann, Parker Schuh, Kensen Shi, Sasha Tsvyashchenko, Joshua Maynez,
  Abhishek Rao, Parker Barnes, Yi~Tay, Noam Shazeer, Vinodkumar Prabhakaran,
  Emily Reif, Nan Du, Ben Hutchinson, Reiner Pope, James Bradbury, Jacob
  Austin, Michael Isard, Guy Gur-Ari, Pengcheng Yin, Toju Duke, Anselm
  Levskaya, Sanjay Ghemawat, Sunipa Dev, Henryk Michalewski, Xavier Garcia,
  Vedant Misra, Kevin Robinson, Liam Fedus, Denny Zhou, Daphne Ippolito, David
  Luan, Hyeontaek Lim, Barret Zoph, Alexander Spiridonov, Ryan Sepassi, David
  Dohan, Shivani Agrawal, Mark Omernick, Andrew~M. Dai,
  Thanumalayan~Sankaranarayana Pillai, Marie Pellat, Aitor Lewkowycz, Erica
  Moreira, Rewon Child, Oleksandr Polozov, Katherine Lee, Zongwei Zhou, Xuezhi
  Wang, Brennan Saeta, Mark Diaz, Orhan Firat, Michele Catasta, Jason Wei,
  Kathy Meier-Hellstern, Douglas Eck, Jeff Dean, Slav Petrov, and Noah Fiedel.
  2022.
\newblock \href {http://arxiv.org/abs/2204.02311} {{PaLM}: {Scaling} {Language}
  {Modeling} with {Pathways}}.
\newblock \emph{arXiv:2204.02311 [cs]}.
\newblock ArXiv: 2204.02311.

\bibitem[{Christen et~al.(2023)Christen, Shaitarova, Stürmer, and
  Niklaus}]{christen_resolving_2023}
Ramona Christen, Anastassia Shaitarova, Matthias Stürmer, and Joel Niklaus.
  2023.
\newblock \href {https://arxiv.org/abs/2309.08695v1} {Resolving {Legalese}: {A}
  {Multilingual} {Exploration} of {Negation} {Scope} {Resolution} in {Legal}
  {Documents}}.

\bibitem[{Ciurlino(2021)}]{BertBR}
Victor~Hugo Ciurlino. 2021.
\newblock \href {https://bdm.unb.br/handle/10483/27824} {Bertbr: a pretrained
  language model for law texts}.
\newblock Master's thesis, Universidade de Brasília.

\bibitem[{Commission(2005)}]{european2005new}
European Commission. 2005.
\newblock A new framework strategy for multilingualism.

\bibitem[{Conneau et~al.(2020)Conneau, Khandelwal, Goyal, Chaudhary, Wenzek,
  Guzm{\'a}n, Grave, Ott, Zettlemoyer, and
  Stoyanov}]{conneau-etal-2020-unsupervised}
Alexis Conneau, Kartikay Khandelwal, Naman Goyal, Vishrav Chaudhary, Guillaume
  Wenzek, Francisco Guzm{\'a}n, Edouard Grave, Myle Ott, Luke Zettlemoyer, and
  Veselin Stoyanov. 2020.
\newblock \href {https://doi.org/10.18653/v1/2020.acl-main.747} {Unsupervised
  cross-lingual representation learning at scale}.
\newblock In \emph{Proceedings of the 58th Annual Meeting of the Association
  for Computational Linguistics}, pages 8440--8451, Online. Association for
  Computational Linguistics.

\bibitem[{de~Gibert et~al.(2022)de~Gibert, Garc{\'\i}a-Pablos, Cuadros, and
  Melero}]{de2022s}
Ona de~Gibert, A~Garc{\'\i}a-Pablos, Montse Cuadros, and Maite Melero. 2022.
\newblock Spanish datasets for sensitive entity detection in the legal domain.
\newblock In \emph{Proceedings of the Thirteenth International Conference on
  Language Resources and Evaluation (LREC’22), Marseille, France, june.
  European Language Resource Association (ELRA)}.
\newblock Dataset URL: https://tinyurl.com/mv65cp66.

\bibitem[{de~la Rosa et~al.(2022)de~la Rosa, Ponferrada, Romero, Villegas,
  de~Prado~Salas, and Grandury}]{bertin-project/bertin-roberta-base-spanish}
Javier de~la Rosa, Eduardo~G. Ponferrada, Manu Romero, Paulo Villegas,
  Pablo~Gonz{\'{a}}lez de~Prado~Salas, and Mar{\'{\i}}a Grandury. 2022.
\newblock \href
  {http://journal.sepln.org/sepln/ojs/ojs/index.php/pln/article/view/6403}
  {{BERTIN:} efficient pre-training of a spanish language model using
  perplexity sampling}.
\newblock \emph{Proces. del Leng. Natural}, 68:13--23.

\bibitem[{de~Vries et~al.(2019)de~Vries, van Cranenburgh, Bisazza, Caselli, van
  Noord, and Nissim}]{GroNLP/bert-base-dutch-cased}
Wietse de~Vries, Andreas van Cranenburgh, Arianna Bisazza, Tommaso Caselli,
  Gertjan van Noord, and Malvina Nissim. 2019.
\newblock Bertje: A dutch bert model.
\newblock \emph{ArXiv}, abs/1912.09582.

\bibitem[{Delobelle et~al.(2020)Delobelle, Winters, and
  Berendt}]{pdelobelle/robbert-v2-dutch-base}
Pieter Delobelle, Thomas Winters, and Bettina Berendt. 2020.
\newblock \href {https://doi.org/10.18653/v1/2020.findings-emnlp.292}
  {{R}ob{BERT}: a {D}utch {R}o{BERT}a-based {L}anguage {M}odel}.
\newblock In \emph{Findings of the Association for Computational Linguistics:
  EMNLP 2020}, pages 3255--3265, Online. Association for Computational
  Linguistics.

\bibitem[{Deng et~al.(2009)Deng, Dong, Socher, Li, Li, and
  Fei{-}Fei}]{DBLP:conf/cvpr/DengDSLL009}
Jia Deng, Wei Dong, Richard Socher, Li{-}Jia Li, Kai Li, and Li~Fei{-}Fei.
  2009.
\newblock \href {https://doi.org/10.1109/CVPR.2009.5206848} {Imagenet: {A}
  large-scale hierarchical image database}.
\newblock In \emph{2009 {IEEE} Computer Society Conference on Computer Vision
  and Pattern Recognition {(CVPR} 2009), 20-25 June 2009, Miami, Florida,
  {USA}}, pages 248--255. {IEEE} Computer Society.

\bibitem[{Devlin et~al.(2019)Devlin, Chang, Lee, and
  Toutanova}]{devlin-etal-2019-bert}
Jacob Devlin, Ming-Wei Chang, Kenton Lee, and Kristina Toutanova. 2019.
\newblock \href {https://doi.org/10.18653/v1/N19-1423} {{BERT}: Pre-training of
  deep bidirectional transformers for language understanding}.
\newblock In \emph{Proceedings of the 2019 Conference of the North {A}merican
  Chapter of the Association for Computational Linguistics: Human Language
  Technologies, Volume 1 (Long and Short Papers)}, pages 4171--4186,
  Minneapolis, Minnesota. Association for Computational Linguistics.

\bibitem[{Douka et~al.(2021)Douka, Abdine, Vazirgiannis, El~Hamdani, and
  Restrepo~Amariles}]{douka-etal-2021-juribert}
Stella Douka, Hadi Abdine, Michalis Vazirgiannis, Rajaa El~Hamdani, and David
  Restrepo~Amariles. 2021.
\newblock \href {https://doi.org/10.18653/v1/2021.nllp-1.9} {{J}uri{BERT}: A
  masked-language model adaptation for {F}rench legal text}.
\newblock In \emph{Proceedings of the Natural Legal Language Processing
  Workshop 2021}, pages 95--101, Punta Cana, Dominican Republic. Association
  for Computational Linguistics.

\bibitem[{Drawzeski et~al.(2021)Drawzeski, Galassi, Jablonowska, Lagioia,
  Lippi, Micklitz, Sartor, Tagiuri, and Torroni}]{draw2021c}
Kasper Drawzeski, Andrea Galassi, Agnieszka Jablonowska, Francesca Lagioia,
  Marco Lippi, Hans~Wolfgang Micklitz, Giovanni Sartor, Giacomo Tagiuri, and
  Paolo Torroni. 2021.
\newblock \href {https://doi.org/10.18653/v1/2021.nllp-1.1} {A corpus for
  multilingual analysis of online terms of service}.
\newblock In \emph{Proceedings of the Natural Legal Language Processing
  Workshop 2021}, pages 1--8.
\newblock Dataset URL: http://claudette.eui.eu/corpora/.

\bibitem[{Dumitrescu et~al.(2020)Dumitrescu, Avram, and
  Pyysalo}]{dumitrescustefan/bert-base-romanian-uncased-v1}
Stefan Dumitrescu, Andrei-Marius Avram, and Sampo Pyysalo. 2020.
\newblock \href {https://doi.org/10.18653/v1/2020.findings-emnlp.387} {The
  birth of {R}omanian {BERT}}.
\newblock In \emph{Findings of the Association for Computational Linguistics:
  EMNLP 2020}, pages 4324--4328, Online. Association for Computational
  Linguistics.

\bibitem[{Greco and Tagarelli(2023)}]{DBLP:journals/corr/abs-2308-05502}
Candida~Maria Greco and Andrea Tagarelli. 2023.
\newblock \href {https://doi.org/10.48550/arXiv.2308.05502} {Bringing order
  into the realm of transformer-based language models for artificial
  intelligence and law}.
\newblock \emph{Artificial Intelligence and Law}, To be published.

\bibitem[{Guha et~al.(2022)Guha, Ho, Nyarko, and R{\'{e}}}]{legalbench}
Neel Guha, Daniel~E. Ho, Julian Nyarko, and Christopher R{\'{e}}. 2022.
\newblock \href {https://doi.org/10.48550/arXiv.2209.06120} {Legalbench:
  Prototyping a collaborative benchmark for legal reasoning}.
\newblock \emph{CoRR}, abs/2209.06120.

\bibitem[{Guti{\'e}rrez-Fandi{\~n}o
  et~al.(2021{\natexlab{a}})Guti{\'e}rrez-Fandi{\~n}o, Armengol-Estap'e,
  Gonzalez-Agirre, and Villegas}]{PlanTL-GOB-ES/RoBERTalex}
Asier Guti{\'e}rrez-Fandi{\~n}o, Jordi Armengol-Estap'e, Aitor Gonzalez-Agirre,
  and Marta Villegas. 2021{\natexlab{a}}.
\newblock Spanish legalese language model and corpora.
\newblock \emph{ArXiv}, abs/2110.12201.

\bibitem[{Guti{\'e}rrez-Fandi{\~n}o
  et~al.(2021{\natexlab{b}})Guti{\'e}rrez-Fandi{\~n}o, Armengol-Estap'e,
  P{\`a}mies, Llop-Palao, Silveira-Ocampo, Carrino, Gonzalez-Agirre,
  Armentano-Oller, Rodr{\'i}guez-Penagos, and
  Villegas}]{PlanTL-GOB-ES/roberta-base-bne}
Asier Guti{\'e}rrez-Fandi{\~n}o, Jordi Armengol-Estap'e, Marc P{\`a}mies, Joan
  Llop-Palao, Joaqu{\'i}n Silveira-Ocampo, Casimiro~Pio Carrino, Aitor
  Gonzalez-Agirre, Carme Armentano-Oller, Carlos Rodr{\'i}guez-Penagos, and
  Marta Villegas. 2021{\natexlab{b}}.
\newblock Mar{IA}: Spanish language models.
\newblock \emph{ArXiv}, abs/2107.07253.

\bibitem[{He et~al.(2021)He, Gao, and Chen}]{He2021b}
Pengcheng He, Jianfeng Gao, and Weizhu Chen. 2021.
\newblock \href {http://arxiv.org/abs/2111.09543} {{DeBERTaV3: Improving
  DeBERTa using ELECTRA-Style Pre-Training with Gradient-Disentangled Embedding
  Sharing}}.
\newblock \emph{CoRR}, pages 1--17.

\bibitem[{He et~al.(2020)He, Liu, Gao, and Chen}]{He2021a}
Pengcheng He, Xiaodong Liu, Jianfeng Gao, and Weizhu Chen. 2020.
\newblock Deberta: Decoding-enhanced bert with disentangled attention.
\newblock \emph{ArXiv}, abs/2006.03654.

\bibitem[{Henderson et~al.(2022)Henderson, Krass, Zheng, Guha, Manning,
  Jurafsky, and Ho}]{henderson_pile_2022}
Peter Henderson, Mark~S. Krass, Lucia Zheng, Neel Guha, Christopher~D. Manning,
  Dan Jurafsky, and Daniel~E. Ho. 2022.
\newblock \href {http://arxiv.org/abs/2207.00220} {Pile of {Law}: {Learning}
  {Responsible} {Data} {Filtering} from the {Law} and a {256GB} {Open}-{Source}
  {Legal} {Dataset}}.
\newblock ArXiv:2207.00220 [cs].

\bibitem[{Hendrycks et~al.(2021)Hendrycks, Burns, Basart, Zou, Mazeika, Song,
  and Steinhardt}]{MMLU}
Dan Hendrycks, Collin Burns, Steven Basart, Andy Zou, Mantas Mazeika, Dawn
  Song, and Jacob Steinhardt. 2021.
\newblock \href {https://openreview.net/forum?id=d7KBjmI3GmQ} {Measuring
  massive multitask language understanding}.
\newblock In \emph{9th International Conference on Learning Representations,
  {ICLR} 2021, Virtual Event, Austria, May 3-7, 2021}. OpenReview.net.

\bibitem[{Hu et~al.(2020)Hu, Ruder, Siddhant, Neubig, Firat, and
  Johnson}]{XTREME}
Junjie Hu, Sebastian Ruder, Aditya Siddhant, Graham Neubig, Orhan Firat, and
  Melvin Johnson. 2020.
\newblock \href {http://proceedings.mlr.press/v119/hu20b.html} {{XTREME:} {A}
  massively multilingual multi-task benchmark for evaluating cross-lingual
  generalisation}.
\newblock In \emph{{ICML}}, volume 119 of \emph{Proceedings of Machine Learning
  Research}, pages 4411--4421. {PMLR}.

\bibitem[{Hua et~al.(2022)Hua, Zhang, Chen, Li, and
  Weber}]{hua_legalrelectra_2022}
Wenyue Hua, Yuchen Zhang, Zhe Chen, Josie Li, and Melanie Weber. 2022.
\newblock \href {https://doi.org/10.48550/arXiv.2212.08204} {{LegalRelectra}:
  {Mixed}-domain {Language} {Modeling} for {Long}-range {Legal} {Text}
  {Comprehension}}.
\newblock ArXiv:2212.08204 [cs].

\bibitem[{Hvingelby et~al.(2020)Hvingelby, Pauli, Barrett, Rosted, Lidegaard,
  and S{\o}gaard}]{Maltehb/danish-bert-botxo}
Rasmus Hvingelby, Amalie~Brogaard Pauli, Maria Barrett, Christina Rosted,
  Lasse~Malm Lidegaard, and Anders S{\o}gaard. 2020.
\newblock Dane: A named entity resource for danish.
\newblock In \emph{International Conference on Language Resources and
  Evaluation}.

\bibitem[{Hwang et~al.(2022)Hwang, Lee, Cho, Lee, and Seo}]{hwang2022a}
Wonseok Hwang, Dongjun Lee, Kyoungyeon Cho, Hanuhl Lee, and Minjoon Seo. 2022.
\newblock \href {https://openreview.net/forum?id=TaARsI_Iio} {A multi-task
  benchmark for korean legal language understanding and judgement prediction}.
\newblock In \emph{Thirty-sixth Conference on Neural Information Processing
  Systems Datasets and Benchmarks Track}.

\bibitem[{Ivgi et~al.(2023)Ivgi, Shaham, and Berant}]{ivgi-etal-2023-efficient}
Maor Ivgi, Uri Shaham, and Jonathan Berant. 2023.
\newblock \href {https://doi.org/10.1162/tacl_a_00547} {Efficient long-text
  understanding with short-text models}.
\newblock \emph{Transactions of the Association for Computational Linguistics},
  11:284--299.

\bibitem[{Jawahar et~al.(2019)Jawahar, Sagot, and
  Seddah}]{jawahar-etal-2019-bert}
Ganesh Jawahar, Beno{\^\i}t Sagot, and Djam{\'e} Seddah. 2019.
\newblock \href {https://doi.org/10.18653/v1/P19-1356} {What does {BERT} learn
  about the structure of language?}
\newblock In \emph{Proceedings of the 57th Annual Meeting of the Association
  for Computational Linguistics}, pages 3651--3657, Florence, Italy.
  Association for Computational Linguistics.

\bibitem[{Katz et~al.(2023)Katz, Hartung, Gerlach, Jana, and
  Bommarito}]{katz_natural_2023}
Daniel~Martin Katz, Dirk Hartung, Lauritz Gerlach, Abhik Jana, and
  Michael~James Bommarito. 2023.
\newblock \href {https://papers.ssrn.com/abstract=4336224} {Natural {Language}
  {Processing} in the {Legal} {Domain}}.

\bibitem[{Koch et~al.(2021)Koch, Denton, Hanna, and Foster}]{life-dataset}
Bernard Koch, Emily Denton, Alex Hanna, and Jacob~G. Foster. 2021.
\newblock \href
  {https://datasets-benchmarks-proceedings.neurips.cc/paper/2021/hash/3b8a614226a953a8cd9526fca6fe9ba5-Abstract-round2.html}
  {Reduced, reused and recycled: The life of a dataset in machine learning
  research}.
\newblock In \emph{NeurIPS Datasets and Benchmarks}.

\bibitem[{Koutsikakis et~al.(2020)Koutsikakis, Chalkidis, Malakasiotis, and
  Androutsopoulos}]{nlpaueb/bert-base-greek-uncased-v1}
John Koutsikakis, Ilias Chalkidis, Prodromos Malakasiotis, and Ion
  Androutsopoulos. 2020.
\newblock \href {https://doi.org/10.1145/3411408.3411440} {Greek-bert: The
  greeks visiting sesame street}.
\newblock In \emph{11th Hellenic Conference on Artificial Intelligence}, SETN
  2020, page 110–117, New York, NY, USA. Association for Computing Machinery.

\bibitem[{Lage-Freitas et~al.(2022)Lage-Freitas, Allende-Cid, Santana, and
  Oliveira-Lage}]{lage2022predicting}
Andr{\'e} Lage-Freitas, H{\'e}ctor Allende-Cid, Orivaldo Santana, and
  L{\'\i}via Oliveira-Lage. 2022.
\newblock Predicting brazilian court decisions.
\newblock \emph{PeerJ Computer Science}, 8:e904.
\newblock Dataset URL:
  https://github.com/proflage/predicting-brazilian-court-decisions.

\bibitem[{Liang et~al.(2022)Liang, Bommasani, Lee, Tsipras, Soylu, Yasunaga,
  Zhang, Narayanan, Wu, Kumar, Newman, Yuan, Yan, Zhang, Cosgrove, Manning,
  Ré, Acosta-Navas, Hudson, Zelikman, Durmus, Ladhak, Rong, Ren, Yao, Wang,
  Santhanam, Orr, Zheng, Yuksekgonul, Suzgun, Kim, Guha, Chatterji, Khattab,
  Henderson, Huang, Chi, Xie, Santurkar, Ganguli, Hashimoto, Icard, Zhang,
  Chaudhary, Wang, Li, Mai, Zhang, and Koreeda}]{liang_holistic_2022}
Percy Liang, Rishi Bommasani, Tony Lee, Dimitris Tsipras, Dilara Soylu,
  Michihiro Yasunaga, Yian Zhang, Deepak Narayanan, Yuhuai Wu, Ananya Kumar,
  Benjamin Newman, Binhang Yuan, Bobby Yan, Ce~Zhang, Christian Cosgrove,
  Christopher~D. Manning, Christopher Ré, Diana Acosta-Navas, Drew~A. Hudson,
  Eric Zelikman, Esin Durmus, Faisal Ladhak, Frieda Rong, Hongyu Ren, Huaxiu
  Yao, Jue Wang, Keshav Santhanam, Laurel Orr, Lucia Zheng, Mert Yuksekgonul,
  Mirac Suzgun, Nathan Kim, Neel Guha, Niladri Chatterji, Omar Khattab, Peter
  Henderson, Qian Huang, Ryan Chi, Sang~Michael Xie, Shibani Santurkar, Surya
  Ganguli, Tatsunori Hashimoto, Thomas Icard, Tianyi Zhang, Vishrav Chaudhary,
  William Wang, Xuechen Li, Yifan Mai, Yuhui Zhang, and Yuta Koreeda. 2022.
\newblock \href {https://doi.org/10.48550/arXiv.2211.09110} {Holistic
  {Evaluation} of {Language} {Models}}.
\newblock ArXiv:2211.09110 [cs].

\bibitem[{Licari and Comand{\'{e}}(2022)}]{itlegalbert}
Daniele Licari and Giovanni Comand{\'{e}}. 2022.
\newblock {ITALIAN-LEGAL-BERT:} {A} pre-trained transformer language model for
  italian law.
\newblock In \emph{{EKAW} (Companion)}, volume 3256 of \emph{{CEUR} Workshop
  Proceedings}. CEUR-WS.org.

\bibitem[{Liu et~al.(2019)Liu, Ott, Goyal, Du, Joshi, Chen, Levy, Lewis,
  Zettlemoyer, and Stoyanov}]{roberta-base}
Yinhan Liu, Myle Ott, Naman Goyal, Jingfei Du, Mandar Joshi, Danqi Chen, Omer
  Levy, Mike Lewis, Luke Zettlemoyer, and Veselin Stoyanov. 2019.
\newblock \href {http://arxiv.org/abs/1907.11692} {{RoBERTa: A Robustly
  Optimized BERT Pretraining Approach}}.
\newblock \emph{CoRR}, abs/2111.09543(1).

\bibitem[{Luz~de Araujo et~al.(2018)Luz~de Araujo, Campos, de~Oliveira,
  Stauffer, Couto, and Bermejo}]{luz2018a}
Pedro~Henrique Luz~de Araujo, Te{\'o}filo E~de Campos, Renato~RR de~Oliveira,
  Matheus Stauffer, Samuel Couto, and Paulo Bermejo. 2018.
\newblock Lener-br: a dataset for named entity recognition in brazilian legal
  text.
\newblock In \emph{International Conference on Computational Processing of the
  Portuguese Language}, pages 313--323. Springer.
\newblock Dataset URL: https://huggingface.co/datasets/lener\_br.

\bibitem[{Malmsten et~al.(2020)Malmsten, B{\"o}rjeson, and
  Haffenden}]{KB/bert-base-swedish-cased}
Martin Malmsten, Love B{\"o}rjeson, and Chris Haffenden. 2020.
\newblock Playing with words at the national library of sweden - making a
  swedish bert.
\newblock \emph{ArXiv}, abs/2007.01658.

\bibitem[{Martin et~al.(2020)Martin, Muller, Ortiz~Su{\'a}rez, Dupont, Romary,
  de~la Clergerie, Seddah, and Sagot}]{camembert-base}
Louis Martin, Benjamin Muller, Pedro~Javier Ortiz~Su{\'a}rez, Yoann Dupont,
  Laurent Romary, {\'E}ric de~la Clergerie, Djam{\'e} Seddah, and Beno{\^\i}t
  Sagot. 2020.
\newblock \href {https://doi.org/10.18653/v1/2020.acl-main.645} {{C}amem{BERT}:
  a tasty {F}rench language model}.
\newblock In \emph{Proceedings of the 58th Annual Meeting of the Association
  for Computational Linguistics}, pages 7203--7219, Online. Association for
  Computational Linguistics.

\bibitem[{Masala et~al.(2021)Masala, Iacob, Uban, Cidota, Velicu, Rebedea, and
  Popescu}]{masala-etal-2021-jurbert}
Mihai Masala, Radu Cristian~Alexandru Iacob, Ana~Sabina Uban, Marina Cidota,
  Horia Velicu, Traian Rebedea, and Marius Popescu. 2021.
\newblock \href {https://doi.org/10.18653/v1/2021.nllp-1.8} {jur{BERT}: A
  {R}omanian {BERT} model for legal judgement prediction}.
\newblock In \emph{Proceedings of the Natural Legal Language Processing
  Workshop 2021}, pages 86--94, Punta Cana, Dominican Republic. Association for
  Computational Linguistics.

\bibitem[{Matthews(1975)}]{MATTHEWS1975442}
B.W. Matthews. 1975.
\newblock \href {https://doi.org/https://doi.org/10.1016/0005-2795(75)90109-9}
  {Comparison of the predicted and observed secondary structure of t4 phage
  lysozyme}.
\newblock \emph{Biochimica et Biophysica Acta (BBA) - Protein Structure},
  405(2):442--451.

\bibitem[{Nemeskey(2020)}]{SZTAKI-HLT/hubert-base-cc}
Dávid~Márk Nemeskey. 2020.
\newblock \emph{Natural Language Processing Methods for Language Modeling}.
\newblock Ph.D. thesis, E\"otv\"os Lor\'and University.

\bibitem[{Niklaus et~al.(2021)Niklaus, Chalkidis, and St{\"u}rmer}]{Nikl2021s}
Joel Niklaus, Ilias Chalkidis, and Matthias St{\"u}rmer. 2021.
\newblock \href {https://doi.org/10.18653/v1/2021.nllp-1.3}
  {{S}wiss-judgment-prediction: A multilingual legal judgment prediction
  benchmark}.
\newblock In \emph{Proceedings of the Natural Legal Language Processing
  Workshop 2021}, pages 19--35, Punta Cana, Dominican Republic. Association for
  Computational Linguistics.

\bibitem[{Niklaus and Giofre(2023)}]{niklaus-giofre-2023-pretrain}
Joel Niklaus and Daniele Giofre. 2023.
\newblock \href {https://doi.org/10.18653/v1/2023.sustainlp-1.11} {Can we
  pretrain a {S}ot{A} legal language model on a budget from scratch?}
\newblock In \emph{Proceedings of The Fourth Workshop on Simple and Efficient
  Natural Language Processing (SustaiNLP)}, pages 158--182, Toronto, Canada
  (Hybrid). Association for Computational Linguistics.

\bibitem[{Niklaus et~al.(2023{\natexlab{a}})Niklaus, Mamié, Stürmer, Brunner,
  and Gygli}]{niklaus_automatic_2023}
Joel Niklaus, Robin Mamié, Matthias Stürmer, Daniel Brunner, and Marcel
  Gygli. 2023{\natexlab{a}}.
\newblock \href {https://doi.org/10.48550/arXiv.2310.04632} {Automatic
  {Anonymization} of {Swiss} {Federal} {Supreme} {Court} {Rulings}}.
\newblock ArXiv:2310.04632 [cs].

\bibitem[{Niklaus et~al.(2023{\natexlab{b}})Niklaus, Matoshi, Stürmer,
  Chalkidis, and Ho}]{niklaus2023multilegalpile}
Joel Niklaus, Veton Matoshi, Matthias Stürmer, Ilias Chalkidis, and Daniel~E.
  Ho. 2023{\natexlab{b}}.
\newblock \href {http://arxiv.org/abs/2306.02069} {Multilegalpile: A 689gb
  multilingual legal corpus}.

\bibitem[{Niklaus et~al.(2022)Niklaus, St{\"u}rmer, and
  Chalkidis}]{niklaus-etal-2022-empirical}
Joel Niklaus, Matthias St{\"u}rmer, and Ilias Chalkidis. 2022.
\newblock \href {https://aclanthology.org/2022.aacl-main.3} {An empirical study
  on cross-{X} transfer for legal judgment prediction}.
\newblock In \emph{Proceedings of the 2nd Conference of the Asia-Pacific
  Chapter of the Association for Computational Linguistics and the 12th
  International Joint Conference on Natural Language Processing (Volume 1: Long
  Papers)}, pages 32--46, Online only. Association for Computational
  Linguistics.

\bibitem[{Pais et~al.(2021)Pais, Mitrofan, Gasan, Coneschi, and
  Ianov}]{pais_vasile_2021_4922385}
Vasile Pais, Maria Mitrofan, Carol~Luca Gasan, Vlad Coneschi, and Alexandru
  Ianov. 2021.
\newblock \href {https://doi.org/10.18653/v1/2021.nllp-1.2} {Named entity
  recognition in the {R}omanian legal domain}.
\newblock In \emph{Proceedings of the Natural Legal Language Processing
  Workshop 2021}, pages 9--18, Punta Cana, Dominican Republic. Association for
  Computational Linguistics.

\bibitem[{Papaloukas et~al.(2021)Papaloukas, Chalkidis, Athinaios, Pantazi, and
  Koubarakis}]{Papa2021m}
Christos Papaloukas, Ilias Chalkidis, Konstantinos Athinaios, Despina Pantazi,
  and Manolis Koubarakis. 2021.
\newblock \href {https://doi.org/10.18653/v1/2021.nllp-1.6} {Multi-granular
  legal topic classification on {G}reek legislation}.
\newblock In \emph{Proceedings of the Natural Legal Language Processing
  Workshop 2021}, pages 63--75, Punta Cana, Dominican Republic. Association for
  Computational Linguistics.

\bibitem[{Peng et~al.(2019)Peng, Yan, and Lu}]{BLUE-benchmark}
Yifan Peng, Shankai Yan, and Zhiyong Lu. 2019.
\newblock \href {https://doi.org/10.18653/v1/W19-5006} {Transfer learning in
  biomedical natural language processing: An evaluation of {BERT} and elmo on
  ten benchmarking datasets}.
\newblock In \emph{BioNLP@ACL}, pages 58--65. Association for Computational
  Linguistics.

\bibitem[{Pikuliak et~al.(2021)Pikuliak, Grivalsky, Konopka, Blst{\'{a}}k,
  Tamajka, Bachrat{\'{y}}, Simko, Bal{\'{a}}zik, Trnka, and
  Uhl{\'{a}}rik}]{gerulata/slovakbert}
Mat{\'{u}}s Pikuliak, Stefan Grivalsky, Martin Konopka, Miroslav Blst{\'{a}}k,
  Martin Tamajka, Viktor Bachrat{\'{y}}, Mari{\'{a}}n Simko, Pavol
  Bal{\'{a}}zik, Michal Trnka, and Filip Uhl{\'{a}}rik. 2021.
\newblock \href {http://arxiv.org/abs/2109.15254} {Slovakbert: Slovak masked
  language model}.
\newblock \emph{CoRR}, abs/2109.15254.

\bibitem[{Raji et~al.(2021)Raji, Denton, Bender, Hanna, and Paullada}]{WWWB}
Inioluwa~Deborah Raji, Emily Denton, Emily~M. Bender, Alex Hanna, and
  Amandalynne Paullada. 2021.
\newblock \href
  {https://datasets-benchmarks-proceedings.neurips.cc/paper/2021/file/084b6fbb10729ed4da8c3d3f5a3ae7c9-Paper-round2.pdf}
  {{AI} and the everything in the whole wide world benchmark}.
\newblock In \emph{NeurIPS Datasets and Benchmarks}.

\bibitem[{Rao et~al.(2019)Rao, Bhattacharya, Thomas, Duan, Chen, Canny, Abbeel,
  and Song}]{TAPE}
Roshan Rao, Nicholas Bhattacharya, Neil Thomas, Yan Duan, Xi~Chen, John~F.
  Canny, Pieter Abbeel, and Yun~S. Song. 2019.
\newblock \href
  {https://proceedings.neurips.cc/paper/2019/hash/37f65c068b7723cd7809ee2d31d7861c-Abstract.html}
  {Evaluating protein transfer learning with {TAPE}}.
\newblock In \emph{NeurIPS}, pages 9686--9698.

\bibitem[{Rasiah et~al.(2023)Rasiah, Stern, Matoshi, Stürmer, Chalkidis, Ho,
  and Niklaus}]{rasiah2023scale}
Vishvaksenan Rasiah, Ronja Stern, Veton Matoshi, Matthias Stürmer, Ilias
  Chalkidis, Daniel~E. Ho, and Joel Niklaus. 2023.
\newblock \href {http://arxiv.org/abs/2306.09237} {Scale: Scaling up the
  complexity for advanced language model evaluation}.

\bibitem[{Sanh et~al.(2019)Sanh, Debut, Chaumond, and
  Wolf}]{Sanh2019DistilBERTAD}
Victor Sanh, Lysandre Debut, Julien Chaumond, and Thomas Wolf. 2019.
\newblock \href {https://arxiv.org/pdf/1910.01108.pdf} {Distilbert, a distilled
  version of bert: smaller, faster, cheaper and lighter}.
\newblock \emph{ArXiv}, abs/1910.01108.

\bibitem[{Semo et~al.(2022)Semo, Bernsohn, Hagag, Hayat, and
  Niklaus}]{semo-etal-2022-classactionprediction}
Gil Semo, Dor Bernsohn, Ben Hagag, Gila Hayat, and Joel Niklaus. 2022.
\newblock \href {https://doi.org/10.18653/v1/2022.nllp-1.3}
  {{C}lass{A}ction{P}rediction: A challenging benchmark for legal judgment
  prediction of class action cases in the {US}}.
\newblock In \emph{Proceedings of the Natural Legal Language Processing
  Workshop 2022}, pages 31--46, Abu Dhabi, United Arab Emirates (Hybrid).
  Association for Computational Linguistics.

\bibitem[{Shen et~al.(2022)Shen, Lo, Yu, Dahlberg, Schlanger, and
  Downey}]{shen2022multilexsum}
Zejiang Shen, Kyle Lo, Lauren Yu, Nathan Dahlberg, Margo Schlanger, and Doug
  Downey. 2022.
\newblock \href {https://openreview.net/forum?id=z1d8fUiS8Cr} {Multi-lexsum:
  Real-world summaries of civil rights lawsuits at multiple granularities}.
\newblock In \emph{Thirty-sixth Conference on Neural Information Processing
  Systems Datasets and Benchmarks Track}.

\bibitem[{Sido et~al.(2021)Sido, Pra{\v{z}}{\'a}k, P{\v{r}}ib{\'a}{\v{n}},
  Pa{\v{s}}ek, Sej{\'a}k, and Konop{\'\i}k}]{UWB-AIR/Czert-B-base-cased}
Jakub Sido, Ond{\v{r}}ej Pra{\v{z}}{\'a}k, Pavel P{\v{r}}ib{\'a}{\v{n}}, Jan
  Pa{\v{s}}ek, Michal Sej{\'a}k, and Miloslav Konop{\'\i}k. 2021.
\newblock \href {https://aclanthology.org/2021.ranlp-1.149} {Czert {--} {C}zech
  {BERT}-like model for language representation}.
\newblock In \emph{Proceedings of the International Conference on Recent
  Advances in Natural Language Processing (RANLP 2021)}, pages 1326--1338, Held
  Online. INCOMA Ltd.

\bibitem[{Souza et~al.(2020)Souza, Nogueira, and
  Lotufo}]{neuralmind/bert-base-portuguese-cased}
F\'{a}bio Souza, Rodrigo Nogueira, and Roberto Lotufo. 2020.
\newblock \href {https://doi.org/10.1007/978-3-030-61377-8_28} {Bertimbau:
  Pretrained bert models for brazilian portuguese}.
\newblock In \emph{Intelligent Systems: 9th Brazilian Conference, BRACIS 2020,
  Rio Grande, Brazil, October 20–23, 2020, Proceedings, Part I}, page
  403–417, Berlin, Heidelberg. Springer-Verlag.

\bibitem[{Tagarelli and Simeri(2022)}]{lamberta}
Andrea Tagarelli and Andrea Simeri. 2022.
\newblock \href
  {http://ircdl2022.dei.unipd.it/downloads/papers/IRCDL_2022_paper_13.pdf}
  {Lamberta: Law article mining based on bert architecture for the italian
  civil code}.
\newblock In \emph{{ICRDL}}.

\bibitem[{Tanvir et~al.(2021)Tanvir, Kittask, Eiche, and
  Sirts}]{tartuNLP/EstBERT}
Hasan Tanvir, Claudia Kittask, Sandra Eiche, and Kairit Sirts. 2021.
\newblock \href {https://aclanthology.org/2021.nodalida-main.2} {{E}st{BERT}: A
  pretrained language-specific {BERT} for {E}stonian}.
\newblock In \emph{Proceedings of the 23rd Nordic Conference on Computational
  Linguistics (NoDaLiDa)}, pages 11--19, Reykjavik, Iceland (Online).
  Link{\"o}ping University Electronic Press, Sweden.

\bibitem[{Tatiana and Valentin(2021)}]{shavrina2021not}
Shavrina Tatiana and Malykh Valentin. 2021.
\newblock \href {http://arxiv.org/abs/2112.01342} {How not to lie with a
  benchmark: Rearranging {NLP} leaderboards}.
\newblock In \emph{I (Still) Can’t Believe It’s Not Better Workshop at
  NeurIPS 2021}, volume abs/2112.01342.

\bibitem[{Torres~Cacoullos(2020)}]{10.3389/fpsyg.2020.02130}
Rena Torres~Cacoullos. 2020.
\newblock \href {https://doi.org/10.3389/fpsyg.2020.02130} {Code-switching
  strategies: Prosody and syntax}.
\newblock \emph{Frontiers in Psychology}, 11.

\bibitem[{Touvron et~al.(2023)Touvron, Lavril, Izacard, Martinet, Lachaux,
  Lacroix, Rozière, Goyal, Hambro, Azhar, Rodriguez, Joulin, Grave, and
  Lample}]{touvron_llama_2023}
Hugo Touvron, Thibaut Lavril, Gautier Izacard, Xavier Martinet, Marie-Anne
  Lachaux, Timothée Lacroix, Baptiste Rozière, Naman Goyal, Eric Hambro,
  Faisal Azhar, Aurelien Rodriguez, Armand Joulin, Edouard Grave, and Guillaume
  Lample. 2023.
\newblock \href {https://doi.org/10.48550/arXiv.2302.13971} {{LLaMA}: {Open}
  and {Efficient} {Foundation} {Language} {Models}}.

\bibitem[{Tsai et~al.(2022)Tsai, Chang, Huang, Huang, Lakhotia, Yang, Dong,
  Liu, Lai, Shi, Chang, Hall, Chen, Li, Watanabe, Mohamed, and Lee}]{SUPERB-SG}
Hsiang{-}Sheng Tsai, Heng{-}Jui Chang, Wen{-}Chin Huang, Zili Huang, Kushal
  Lakhotia, Shu{-}Wen Yang, Shuyan Dong, Andy~T. Liu, Cheng{-}I Lai, Jiatong
  Shi, Xuankai Chang, Phil Hall, Hsuan{-}Jui Chen, Shang{-}Wen Li, Shinji
  Watanabe, Abdelrahman Mohamed, and Hung{-}yi Lee. 2022.
\newblock \href {https://doi.org/10.18653/v1/2022.acl-long.580} {{SUPERB-SG:}
  enhanced speech processing universal performance benchmark for semantic and
  generative capabilities}.
\newblock In \emph{{ACL} {(1)}}, pages 8479--8492. Association for
  Computational Linguistics.

\bibitem[{Tsarapatsanis and Aletras(2021)}]{tsarapatsanis-aletras-2021-ethical}
Dimitrios Tsarapatsanis and Nikolaos Aletras. 2021.
\newblock \href {https://doi.org/10.18653/v1/2021.findings-acl.314} {On the
  ethical limits of natural language processing on legal text}.
\newblock In \emph{Findings of the Association for Computational Linguistics:
  ACL-IJCNLP 2021}, pages 3590--3599, Online. Association for Computational
  Linguistics.

\bibitem[{Tziafas et~al.(2021)Tziafas, de~Saint-Phalle, de~Vries, Egger, and
  Caselli}]{tzia2021}
Georgios Tziafas, Eugenie de~Saint-Phalle, Wietse de~Vries, Clara Egger, and
  Tommaso Caselli. 2021.
\newblock A multilingual approach to identify and classify exceptional measures
  against covid-19.
\newblock In \emph{Proceedings of the Natural Legal Language Processing
  Workshop 2021}, pages 46--62.
\newblock Dataset URL: https://tinyurl.com/ycysvtbm.

\bibitem[{Urchs. et~al.(2021)Urchs., Mitrović., and
  Granitzer.}]{Urchs_et_all2021}
Stefanie Urchs., Jelena Mitrović., and Michael Granitzer. 2021.
\newblock \href {https://doi.org/10.5220/0010187305150521} {Design and
  implementation of german legal decision corpora}.
\newblock In \emph{Proceedings of the 13th International Conference on Agents
  and Artificial Intelligence - Volume 2: ICAART,}, pages 515--521. SciTePress.

\bibitem[{Villata et~al.(2022)Villata, Araszkiewicz, Ashley, Bench-Capon,
  Branting, Conrad, and Wyner}]{Villata2022ThirtyYO}
Serena Villata, Michał Araszkiewicz, Kevin~D. Ashley, Trevor J.~M.
  Bench-Capon, L.~Karl Branting, Jack~G. Conrad, and Adam~Zachary Wyner. 2022.
\newblock Thirty years of artificial intelligence and law: the third decade.
\newblock \emph{Artificial Intelligence and Law}, 30:561--591.

\bibitem[{Virtanen et~al.(2019)Virtanen, Kanerva, Ilo, Luoma, Luotolahti,
  Salakoski, Ginter, and Pyysalo}]{TurkuNLP/bert-base-finnish-cased-v1}
Antti Virtanen, Jenna Kanerva, Rami Ilo, Jouni Luoma, Juhani Luotolahti, Tapio
  Salakoski, Filip Ginter, and Sampo Pyysalo. 2019.
\newblock Multilingual is not enough: Bert for finnish.
\newblock \emph{ArXiv}, abs/1912.07076.

\bibitem[{Wang et~al.(2019{\natexlab{a}})Wang, Pruksachatkun, Nangia, Singh,
  Michael, Hill, Levy, and Bowman}]{SUPERGLUE}
Alex Wang, Yada Pruksachatkun, Nikita Nangia, Amanpreet Singh, Julian Michael,
  Felix Hill, Omer Levy, and Samuel~R. Bowman. 2019{\natexlab{a}}.
\newblock \href
  {https://proceedings.neurips.cc/paper/2019/hash/4496bf24afe7fab6f046bf4923da8de6-Abstract.html}
  {Superglue: {A} stickier benchmark for general-purpose language understanding
  systems}.
\newblock In \emph{NeurIPS}, pages 3261--3275.

\bibitem[{Wang et~al.(2019{\natexlab{b}})Wang, Singh, Michael, Hill, Levy, and
  Bowman}]{GLUE}
Alex Wang, Amanpreet Singh, Julian Michael, Felix Hill, Omer Levy, and
  Samuel~R. Bowman. 2019{\natexlab{b}}.
\newblock \href {https://openreview.net/forum?id=rJ4km2R5t7} {{GLUE:} {A}
  multi-task benchmark and analysis platform for natural language
  understanding}.
\newblock In \emph{{ICLR} (Poster)}. OpenReview.net.

\bibitem[{Wang et~al.(2020)Wang, Wei, Dong, Bao, Yang, and Zhou}]{Wang2020}
Wenhui Wang, Furu Wei, Li~Dong, Hangbo Bao, Nan Yang, and Ming Zhou. 2020.
\newblock Minilm: Deep self-attention distillation for task-agnostic
  compression of pre-trained transformers.
\newblock In \emph{Proceedings of the 34th International Conference on Neural
  Information Processing Systems}, NIPS'20, Red Hook, NY, USA. Curran
  Associates Inc.

\bibitem[{Xiao et~al.(2021)Xiao, Hu, Liu, Tu, and Sun}]{lawformer}
Chaojun Xiao, Xueyu Hu, Zhiyuan Liu, Cunchao Tu, and Maosong Sun. 2021.
\newblock \href {https://doi.org/10.1016/j.aiopen.2021.06.003} {Lawformer: {A}
  pre-trained language model for chinese legal long documents}.
\newblock \emph{{AI} Open}, 2:79--84.

\bibitem[{Xu et~al.(2020)Xu, Hu, Zhang, Li, Cao, Li, Xu, Sun, Yu, Yu, Tian,
  Dong, Liu, Shi, Cui, Li, Zeng, Wang, Xie, Li, Patterson, Tian, Zhang, Zhou,
  Liu, Zhao, Zhao, Yue, Zhang, Yang, Richardson, and Lan}]{CLUE}
Liang Xu, Hai Hu, Xuanwei Zhang, Lu~Li, Chenjie Cao, Yudong Li, Yechen Xu, Kai
  Sun, Dian Yu, Cong Yu, Yin Tian, Qianqian Dong, Weitang Liu, Bo~Shi, Yiming
  Cui, Junyi Li, Jun Zeng, Rongzhao Wang, Weijian Xie, Yanting Li, Yina
  Patterson, Zuoyu Tian, Yiwen Zhang, He~Zhou, Shaoweihua Liu, Zhe Zhao, Qipeng
  Zhao, Cong Yue, Xinrui Zhang, Zhengliang Yang, Kyle Richardson, and Zhenzhong
  Lan. 2020.
\newblock \href {https://doi.org/10.18653/v1/2020.coling-main.419} {{CLUE}: A
  {C}hinese language understanding evaluation benchmark}.
\newblock In \emph{COLING}, pages 4762--4772, Barcelona, Spain (Online).
  International Committee on Computational Linguistics.

\bibitem[{Xue et~al.(2021)Xue, Constant, Roberts, Kale, Al-Rfou, Siddhant,
  Barua, and Raffel}]{xue_mt5_2021}
Linting Xue, Noah Constant, Adam Roberts, Mihir Kale, Rami Al-Rfou, Aditya
  Siddhant, Aditya Barua, and Colin Raffel. 2021.
\newblock \href {http://arxiv.org/abs/2010.11934} {{mT5}: {A} massively
  multilingual pre-trained text-to-text transformer}.
\newblock \emph{arXiv:2010.11934 [cs]}.
\newblock ArXiv: 2010.11934.

\bibitem[{Yang et~al.(2021)Yang, Chi, Chuang, Lai, Lakhotia, Lin, Liu, Shi,
  Chang, Lin, Huang, Tseng, Lee, Liu, Huang, Dong, Li, Watanabe, Mohamed, and
  Lee}]{SUPERB}
Shu{-}Wen Yang, Po{-}Han Chi, Yung{-}Sung Chuang, Cheng{-}I~Jeff Lai, Kushal
  Lakhotia, Yist~Y. Lin, Andy~T. Liu, Jiatong Shi, Xuankai Chang, Guan{-}Ting
  Lin, Tzu{-}Hsien Huang, Wei{-}Cheng Tseng, Ko{-}tik Lee, Da{-}Rong Liu, Zili
  Huang, Shuyan Dong, Shang{-}Wen Li, Shinji Watanabe, Abdelrahman Mohamed, and
  Hung{-}yi Lee. 2021.
\newblock \href {https://doi.org/10.21437/Interspeech.2021-1775} {{SUPERB:}
  speech processing universal performance benchmark}.
\newblock In \emph{Interspeech}, pages 1194--1198. {ISCA}.

\bibitem[{Yin and Habernal(2022)}]{Ying.Habernal.2022.NLLP}
Ying Yin and Ivan Habernal. 2022.
\newblock \href {https://doi.org/10.18653/v1/2022.nllp-1.14}
  {Privacy-preserving models for legal natural language processing}.
\newblock In \emph{Proceedings of the Natural Legal Language Processing
  Workshop 2022}, pages 172--183, Abu Dhabi, United Arab Emirates (Hybrid).
  Association for Computational Linguistics.

\bibitem[{Zhang et~al.(2022)Zhang, Chen, Bi, Liang, Li, Shang, Yin, Tan, Xu,
  Huang, Si, Ni, Xie, Sui, Chang, Zong, Yuan, Li, Yan, Zan, Zhang, Tang, and
  Chen}]{cblue}
Ningyu Zhang, Mosha Chen, Zhen Bi, Xiaozhuan Liang, Lei Li, Xin Shang, Kangping
  Yin, Chuanqi Tan, Jian Xu, Fei Huang, Luo Si, Yuan Ni, Guotong Xie, Zhifang
  Sui, Baobao Chang, Hui Zong, Zheng Yuan, Linfeng Li, Jun Yan, Hongying Zan,
  Kunli Zhang, Buzhou Tang, and Qingcai Chen. 2022.
\newblock \href {https://doi.org/10.18653/v1/2022.acl-long.544} {{CBLUE}: A
  {C}hinese biomedical language understanding evaluation benchmark}.
\newblock In \emph{ACL}, pages 7888--7915, Dublin, Ireland. Association for
  Computational Linguistics.

\bibitem[{Zheng et~al.(2021)Zheng, Guha, Anderson, Henderson, and
  Ho}]{zlucia/custom-legalbert}
Lucia Zheng, Neel Guha, Brandon~R. Anderson, Peter Henderson, and Daniel~E. Ho.
  2021.
\newblock \href {https://doi.org/10.1145/3462757.3466088} {When does
  pretraining help? assessing self-supervised learning for law and the casehold
  dataset of 53,000+ legal holdings}.
\newblock In \emph{Proceedings of the Eighteenth International Conference on
  Artificial Intelligence and Law}, ICAIL '21, page 159–168, New York, NY,
  USA. Association for Computing Machinery.

\bibitem[{Zhong et~al.(2020)Zhong, Xiao, Tu, Zhang, Liu, and
  Sun}]{zhong-etal-2020-nlp}
Haoxi Zhong, Chaojun Xiao, Cunchao Tu, Tianyang Zhang, Zhiyuan Liu, and Maosong
  Sun. 2020.
\newblock \href {https://doi.org/10.18653/v1/2020.acl-main.466} {How does {NLP}
  benefit legal system: A summary of legal artificial intelligence}.
\newblock In \emph{Proceedings of the 58th Annual Meeting of the Association
  for Computational Linguistics}, pages 5218--5230, Online. Association for
  Computational Linguistics.

\end{thebibliography}
\bibliographystyle{acl_natbib}

\appendix
\section{Experiment Details}
\label{sec:experiment_details}

\subsection{Maximum Sequence Lengths}
\label{sec:max_seq_length}
Brazilian Court Decisions: 1024 (128 x 8) \\
CoVID19: 256 \\
German Argument Mining: 256 \\
Greek Legal Code: 4096 (if speed is important: 2048) (128 x 32 / 16) \\
Greek Legal NER: 512 (max for non-hierarchical) \\
LegalNERo: 512 (max for non-hierarchical) \\
LeNER: 512 (max for non-hierarchical) \\
MAPA: 512 (max for non-hierarchical) \\
MultiEURLEX: 4096 (or for maximum performance 8192) (128 x 32 / 64)
Online Terms of Service: 256 \\
Swiss Judgment Prediction: 2048 (or for maximum performance on fr: 4096) (128 x 16 / 32) \\

\subsection{Total compute}
\label{sec:totel_compute}
We used a total of 689 GPU days.

\subsection{Hyperparameters}
\label{sec:hyperparameters}
We used learning rate 1e-5 for all models and datasets without tuning. We ran all experiments with 3 random seeds (1-3). We always used batch size 64. In case the GPU memory was insufficient, we additionally used gradient accumulation. We trained using early stopping on the validation loss with an early-stopping patience of 5 epochs. Because MultiEURLEX is very large and the experiment very long, we just train for 1 epoch and evaluated after every 1000\textsuperscript{th} step when finetuning multilingual models on the entire dataset. For finetuning the monolingual models on language-specific subsets of MultiEURLEX, we evaluated on the basis of epochs. We used AMP mixed precision training and evaluation to reduce costs. Mixed precision was not used in combination with microsoft/mdeberta-v3-base because it led to errors. For the experiments we used the following NVIDIA GPUs: 24GB RTX3090, 32GB V100 and 80GB A100. 

\section{Model Descriptions}
\label{sec:model_descriptions}

\paragraph{MiniLM.}
MiniLM \cite{Wang2020} is the result of a novel task-agnostic compression technique, also called distillation, in which a compact model — the so-called student — is trained to reproduce the behaviour of a larger pre-trained model — the so-called teacher. This is achieved by deep self-attention distillation, i.e. only the self-attention module of the last Transformer layer of the teacher, which stores a lot of contextual information \cite{jawahar-etal-2019-bert}, is distilled. The student is trained by closely imitating the teacher's final Transformer layer's self-attention behavior. To aid the learner in developing a better imitation, \cite{Wang2020} also introduce the self-attention value-relation transfer in addition to the self-attention distributions. The addition of a teacher assistant results in further improvements. For the training of multilingual MiniLM, XLM-R\textsubscript{BASE} was used.

\paragraph{DistilBERT}
DistilBERT \cite{Sanh2019DistilBERTAD} is a more compressed version of BERT \cite{devlin-etal-2019-bert} using teacher-student learning, similar to MiniLM. DistilBERT is distilled from BERT, thus both share a similar overall architecture. The pooler and token-type embeddings are eliminated, and the number of layers is decreased by a factor of 2 in DistilBERT. DistilBERT is distilled in very large batches while utilizing gradient accumulation and dynamic masking, but without the next sentence prediction objective. DistilBERT was trained on the same corpus as the original BERT.  

\paragraph{mDEBERTa}
\citet{He2021a} suggest a new model architecture called DeBERTa (Decoding-enhanced BERT with disentangled attention), which employs two novel methods to improve the BERT and RoBERTa models. The first is the disentangled attention mechanism, in which each word is represented by two vectors that encode its content and position, respectively, and the attention weights between words are calculated using disentangled matrices on their respective contents and relative positions. To predict the masked tokens during pre-training, an enhanced mask decoder is utilized, which incorporates absolute positions in the decoding layer. Additionally, the generalization of models is enhanced through fine-tuning using a new virtual adversarial training technique. \citet{He2021b} introduce mDEBERTa-v3 by further improving the efficiency of pre-training by replacing Masked-Language Modeling (MLM) in DeBERTa with the task of replaced token detection (RTD) where the model is trained to predict whether a token in the corrupted input is either original or replaced by agenerator. Further improvements are achieved via \textit{gradient-disentangled embedding sharing} (GDES).

\paragraph{XLM-RoBERTa}
XLM-R \cite{conneau-etal-2020-unsupervised} is a multilingual language model which has the same pretraining objectives as RoBERTa \cite{roberta-base}, such as  dynamic masking, but not next sentence prediction. It is pre-trained on a large corpus comprising 100 languages. The authors report a significant performance gain over multilingual BERT (mBERT) in a variety of tasks with results competitive with state-of-the-art monolingual models \cite{conneau-etal-2020-unsupervised}.

%\paragraph{mT5.}
%mT5 \cite{xue-etal-2021-mt5} is a multilingual encoder-decoder model that has been trained on a Common Crawl-based dataset covering 101 languages. The architecture and training procedure are based on that of T5 \cite{Raffel2020}. T5 treats every text processing problem as a “text-to-text” problem, i.e. taking a text as input and producing a new text as output. It uses a basic encoder-decoder Transformer architecture as proposed by \cite{vaswani_attention_2017} and is trained to predict missing or otherwise corrupted tokens in the input.

\section{Monolingual Models Overview}
\label{sec:monolingual_models_overview}

\begin{table*}[t]
\centering
\resizebox{\textwidth}{!}{
\begin{tabular}{lllllll}
\toprule
                                \textbf{Model} & \textbf{Language} &                                       \textbf{Source} & \textbf{Params} & \textbf{Vocab} &         \textbf{Specs} \\
\midrule
general\\
\midrule
              iarfmoose/roberta-base-bulgarian &                bg &                                                   -   &            126M &            52K &     200K steps /  BS 8 \\
                    UWB-AIR/Czert-B-base-cased &                cs &                     \cite{UWB-AIR/Czert-B-base-cased} &            109M &            31K &              50K steps \\
                     Maltehb/danish-bert-botxo &                da &                      \cite{Maltehb/danish-bert-botxo} &            111M &            32K &   1M steps  /  BS 1280 \\
                  dbmdz/bert-base-german-cased &                de &                                                  -    &            110M &            31K &             1.5M steps \\
                            deepset/gbert-base &                de &                             \cite{deepset/gbert-base} &            110M &            31K &  30K steps  /  BS 1024 \\
            nlpaueb/bert-base-greek-uncased-v1 &                el &             \cite{nlpaueb/bert-base-greek-uncased-v1} &            113M &            35K &    1M steps  /  BS 256 \\
\bottomrule
\end{tabular}
}
\caption{Monolingual models. BS is short for batch size. For a detailed overview of the pretraining corpora, we refer to the publications. For some models we were not able to find publications/specs.}.
\label{tab:overview_monolingual_models}
\end{table*}

\newgeometry{margin=2.5cm} 
\section{Dataset Splits}
\label{sec:dataset_splits}

\begin{sidewaystable}

\centering
\begin{adjustbox}{width=0.9\textwidth}
%%\medium
\begin{tabular}{llllllllll}
\toprule
\textbf{Language} &        \textbf{SJP} & \textbf{OTS-UL} &   \textbf{OTS-CT} &     \textbf{C19} &      \textbf{MEU-1} &      \textbf{MEU-2} &      \textbf{MEU-3} &   \textbf{MAP-C} &   \textbf{MAP-F} \\
\midrule
               bg &                     &                 &                   &                  & 15986 / 5000 / 5000 & 15986 / 5000 / 5000 & 15986 / 5000 / 5000 & 1411 / 166 / 560 & 1411 / 166 / 560 \\
               cs &                     &                 &                   &                  & 23187 / 5000 / 5000 & 23187 / 5000 / 5000 & 23187 / 5000 / 5000 & 1464 / 176 / 563 & 1464 / 176 / 563 \\
               da &                     &                 &                   &                  & 55000 / 5000 / 5000 & 55000 / 5000 / 5000 & 55000 / 5000 / 5000 & 1455 / 164 / 550 & 1455 / 164 / 550 \\
               de & 35458 / 4705 / 9725 &  491 / 42 / 103 & 4480 / 404 / 1027 &                  & 55000 / 5000 / 5000 & 55000 / 5000 / 5000 & 55000 / 5000 / 5000 & 1457 / 166 / 558 & 1457 / 166 / 558 \\
               el &                     &                 &                   &                  & 55000 / 5000 / 5000 & 55000 / 5000 / 5000 & 55000 / 5000 / 5000 & 1529 / 174 / 584 & 1529 / 174 / 584 \\
               en &                     &  526 / 49 / 103 & 5378 / 415 / 1038 &    648 / 81 / 81 & 55000 / 5000 / 5000 & 55000 / 5000 / 5000 & 55000 / 5000 / 5000 &   893 / 98 / 408 &   893 / 98 / 408 \\
               es &                     &                 &                   &                  & 52785 / 5000 / 5000 & 52785 / 5000 / 5000 & 52785 / 5000 / 5000 &  806 / 248 / 155 &  806 / 248 / 155 \\
\bottomrule
\end{tabular}
\end{adjustbox}
\caption{Overview of the number of examples for each language-specific subset of multilingual tasks. The order of the values is train / validation / test.}
\label{subset_for_language_specific_overview_num_ex}
\end{sidewaystable}

\begin{sidewaystable}
\centering
% \begin{adjustbox}{width=1\textwidth}
%%\medium
\begin{tabular}{llllllllll}
\toprule
\textbf{Language} & \textbf{SJP} & \textbf{OTS-UL} & \textbf{OTS-CT} & \textbf{C19} & \textbf{MEU-1} &  \textbf{MEU-2} &  \textbf{MEU-3} & \textbf{MAP-C} & \textbf{MAP-F} \\
\midrule
               bg &              &                 &                 &              &   21 / 21 / 21 & 127 / 126 / 127 & 481 / 454 / 465 &    11 / 11 / 8 &   24 / 16 / 13 \\
               cs &              &                 &                 &              &   21 / 21 / 21 & 127 / 126 / 127 & 486 / 454 / 465 &    11 / 11 / 9 &   30 / 17 / 16 \\
               da &              &                 &                 &              &   21 / 21 / 21 & 127 / 126 / 127 & 500 / 454 / 465 &   11 / 10 / 11 &   26 / 14 / 14 \\
               de &    2 / 2 / 2 &       3 / 3 / 3 &       9 / 7 / 9 &              &   21 / 21 / 21 & 127 / 126 / 127 & 500 / 454 / 465 &    11 / 9 / 10 &   28 / 14 / 14 \\
               el &              &                 &                 &              &   21 / 21 / 21 & 127 / 126 / 127 & 500 / 454 / 465 &   11 / 11 / 11 &   31 / 17 / 20 \\
               en &              &       3 / 3 / 3 &       9 / 8 / 9 &    6 / 6 / 5 &   21 / 21 / 21 & 127 / 126 / 127 & 500 / 454 / 465 &     11 / 9 / 9 &   28 / 17 / 18 \\
               es &              &                 &                 &              &   21 / 21 / 21 & 127 / 126 / 127 & 497 / 454 / 465 &    11 / 8 / 11 &   26 / 13 / 18 \\
\bottomrule
\end{tabular}
% \end{adjustbox}
\caption{Overview of the number of labels for each language-specific subset of multilingual tasks. The order of the values is train / validation / test.}
\label{subset_for_language_specific_overview_num_labels}
\end{sidewaystable}

\newpage
\section{Detailed Multilingual Results}
\label{sec:detailed_multilingual_results}

\begin{sidewaystable}
\begin{adjustbox}{width=1\textwidth}
\begin{tabular}{llllllllllllllllllll}
\toprule
{} & \textbf{\textbf{Mean}} & \textbf{\textbf{BCD-J}} & \textbf{\textbf{BCD-U}} & \textbf{\textbf{GAM}} & \textbf{\textbf{GLC-V}} & \textbf{\textbf{GLC-C}} & \textbf{\textbf{GLC-S}} & \textbf{\textbf{SJP}} & \textbf{\textbf{OTS-UL}} & \textbf{\textbf{OTS-CT}} & \textbf{\textbf{C19}} & \textbf{\textbf{MEU-1}} & \textbf{\textbf{MEU-2}} & \textbf{\textbf{MEU-3}} & \textbf{\textbf{GLN}} & \textbf{\textbf{LNR}} & \textbf{\textbf{LNB}} & \textbf{\textbf{MAP-C}} & \textbf{\textbf{MAP-F}} \\
Model                      &                        &                         &                         &                       &                         &                         &                         &                       &                          &                          &                       &                         &                         &                         &                       &                       &                       &                         &                         \\
\midrule
MiniLM                     &                   51.0 &            52.8 (±6.7)  &            55.1 (±6.6)  &          72.1 (±0.9)  &            82.0 (±1.0)  &            39.4 (±1.0)  &         \bf 5.1 (±1.6)  &          68.9 (±0.7)  &             71.0 (±5.0)  &             15.3 (±3.4)  &           5.8 (±1.5)  &            64.8 (±0.3)  &            23.5 (±0.6)  &         6.5 (±0.3)  &      \bf 66.6 (±1.4)  &          62.9 (±6.4)  &      \bf 98.2 (±0.1)  &        \bf 79.0 (±0.3)  &            49.3 (±1.4)  \\
\bottomrule
\end{tabular}
\end{adjustbox}
\caption{Arithmetic mean of macro-F1 and the standard deviation over all seeds for multilingual models from the validation set. The best scores are in bold.}
\label{tab:arithmetic_mean_multilingual_models_eval_set}\vspace{1cm}\begin{adjustbox}{width=1\textwidth}
\begin{tabular}{llllllllllllllllllll}
\toprule
{} & \textbf{\textbf{Mean}} & \textbf{\textbf{BCD-J}} & \textbf{\textbf{BCD-U}} & \textbf{\textbf{GAM}} & \textbf{\textbf{GLC-V}} & \textbf{\textbf{GLC-C}} & \textbf{\textbf{GLC-S}} & \textbf{\textbf{SJP}} & \textbf{\textbf{OTS-UL}} & \textbf{\textbf{OTS-CT}} & \textbf{\textbf{C19}} & \textbf{\textbf{MEU-1}} & \textbf{\textbf{MEU-2}} & \textbf{\textbf{MEU-3}} & \textbf{\textbf{GLN}} & \textbf{\textbf{LNR}} & \textbf{\textbf{LNB}} & \textbf{\textbf{MAP-C}} & \textbf{\textbf{MAP-F}} \\
Model                      &                        &                         &                         &                       &                         &                         &                         &                       &                          &                          &                       &                         &                         &                         &                       &                       &                       &                         &                         \\
\midrule
MiniLM                     &                   46.4 &            49.4 (±7.4)  &            56.7 (±7.9)  &      \bf 73.3 (±0.9)  &            81.7 (±0.5)  &            39.4 (±1.4)  &         \bf 5.2 (±1.6)  &          67.6 (±1.2)  &             74.6 (±1.1)  &             14.1 (±3.1)  &       \bf 6.0 (±1.9)  &            62.0 (±0.4)  &            21.7 (±0.5)  &         5.6 (±0.3)  &          43.6 (±2.4)  &          46.5 (±1.5)  &          86.1 (±0.3)  &            62.7 (±1.7)  &            39.9 (±2.3)  \\
\bottomrule
\end{tabular}
\end{adjustbox}
\caption{Arithmetic mean of macro-F1 and the standard deviation over all seeds for multilingual models from the test set. The best scores are in bold.}
\label{tab:arithmetic_mean_multilingual_models_test_set}\end{sidewaystable}

\newpage
\section{Detailed Monolingual Results}
\label{sec:detailed_monolingual_results}

\begin{sidewaystable}
\begin{adjustbox}{width=1\textwidth}
\begin{tabular}{llllllllllllllllllll}
\toprule
{} & \textbf{\textbf{Mean}} & \textbf{\textbf{BCD-J}} & \textbf{\textbf{BCD-U}} & \textbf{\textbf{GAM}} & \textbf{\textbf{GLC-V}} & \textbf{\textbf{GLC-C}} & \textbf{\textbf{GLC-S}} & \textbf{\textbf{SJP}} & \textbf{\textbf{OTS-UL}} & \textbf{\textbf{OTS-CT}} & \textbf{\textbf{C19}} & \textbf{\textbf{MEU-1}} & \textbf{\textbf{MEU-2}} & \textbf{\textbf{MEU-3}} & \textbf{\textbf{GLN}} & \textbf{\textbf{LNR}} & \textbf{\textbf{LNB}} & \textbf{\textbf{MAP-C}} & \textbf{\textbf{MAP-F}} \\
Model                                          &                        &                         &                         &                       &                         &                         &                         &                       &                          &                          &                       &                         &                         &                         &                       &                       &                       &                         &                         \\
\midrule
iarfmoose/roberta-base-bulgarian               &                   52.3 &                       - &                       - &                     - &                       - &                       - &                       - &                     - &                        - &                        - &                     - &            67.4 (±0.7)  &            35.7 (±2.1)  &             3.7 (±5.2)  &                     - &                     - &                     - &            82.8 (±2.1)  &            72.0 (±2.8)  \\
\bottomrule
\end{tabular}
\end{adjustbox}
\caption{Arithmetic mean of macro-F1 and the standard deviation over all seeds for monolingual models from the validation set.}
\label{tab:arithmetic_mean_monolingual_models_eval_set}
\end{sidewaystable}

\begin{sidewaystable}
\begin{adjustbox}{width=1\textwidth}
\begin{tabular}{llllllllllllllllllll}
\toprule
{} & \textbf{\textbf{Mean}} & \textbf{\textbf{BCD-J}} & \textbf{\textbf{BCD-U}} & \textbf{\textbf{GAM}} & \textbf{\textbf{GLC-V}} & \textbf{\textbf{GLC-C}} & \textbf{\textbf{GLC-S}} & \textbf{\textbf{SJP}} & \textbf{\textbf{OTS-UL}} & \textbf{\textbf{OTS-CT}} & \textbf{\textbf{C19}} & \textbf{\textbf{MEU-1}} & \textbf{\textbf{MEU-2}} & \textbf{\textbf{MEU-3}} & \textbf{\textbf{GLN}} & \textbf{\textbf{LNR}} & \textbf{\textbf{LNB}} & \textbf{\textbf{MAP-C}} & \textbf{\textbf{MAP-F}} \\
Model                                          &                        &                         &                         &                       &                         &                         &                         &                       &                          &                          &                       &                         &                         &                         &                       &                       &                       &                         &                         \\
\midrule
iarfmoose/roberta-base-bulgarian               &                   46.7 &                       - &                       - &                     - &                       - &                       - &                       - &                     - &                        - &                        - &                     - &            63.1 (±0.1)  &            32.6 (±1.8)  &             3.2 (±4.5)  &                     - &                     - &                     - &            62.5 (±3.6)  &            72.1 (±8.2)  \\
\bottomrule
\end{tabular}
\end{adjustbox}
\caption{Arithmetic mean of macro-F1 and the standard deviation over all seeds for monolingual models from the test set.}
\label{tab:arithmetic_mean_monolingual_models_test_set}
\end{sidewaystable}

\newpage
\afterpage{\clearpage}

\section{Original Paper Results}
\label{sec:paper_results}

In this section, we present an overview of scores for each configuration of the LEXTREME dataset as provided in the original papers. When certain configurations were not available, no scores were obtained. It should be noted that different papers provide varying scores, making direct comparisons with our results challenging. Additionally, the variability in the training and evaluation procedure used across different papers may impact the resulting scores, which is an important factor to consider. To gain a better understanding of the training and evaluation procedure please refer to the cited references. 
The LEXTREME scores are calculated by taking the arithmetic mean of each seed (three in total). 

\begin{table*}[ht]
\centering
\resizebox{\textwidth}{!}{
\begin{tabular}{lllllllllllll}
\toprule
           \textbf{Source} &                               \textbf{Method} & \textbf{TrainLang} & \textbf{TestLang} & \textbf{macro-precision} & \textbf{macro-recall} & \textbf{macro-f1} & \textbf{micro-precision} & \textbf{micro-recall} & \textbf{micro-f1} & \textbf{precision} & \textbf{recall} & \textbf{accuracy} \\
\midrule
\citet{lage2022predicting} &                                   BERT-Imbau  &                 pt &                pt &                        - &                     - &          \bf 73.0 &                        - &                     - &                 - &               66.0 &            63.0 &              73.0 \\
\citet{lage2022predicting} & Bidirectional Long Short-Term Memory (BiLSTM) &                 pt &                pt &                        - &                     - &              55.0 &                        - &                     - &                 - &               43.0 &            41.0 &              55.0 \\
\citet{lage2022predicting} &           Convolutional Neural Networks (CNN) &                 pt &                pt &                        - &                     - &              61.0 &                        - &                     - &                 - &               65.0 &            59.0 &              70.0 \\
\citet{lage2022predicting} &                                 decision tree &                 pt &                pt &                        - &                     - &              63.0 &                        - &                     - &                 - &               67.0 &            61.0 &              72.0 \\
\citet{lage2022predicting} &           eXtreme Gradient Boosting (XGBoost) &                 pt &                pt &                        - &                     - &              70.0 &                        - &                     - &                 - &           \bf 73.0 &        \bf 68.0 &          \bf 77.0 \\
\bottomrule
\end{tabular}
}
\caption{BCD-J. The best scores are in bold.}
\label{paper_results_BCD-J}
\end{table*}
\begin{table*}[h]
\centering
\resizebox{\textwidth}{!}{
\begin{tabular}{lllllllllllll}
\toprule
           \textbf{Source} &                               \textbf{Method} & \textbf{TrainLang} & \textbf{TestLang} & \textbf{macro-precision} & \textbf{macro-recall} & \textbf{macro-f1} & \textbf{micro-precision} & \textbf{micro-recall} & \textbf{micro-f1} & \textbf{precision} & \textbf{recall} & \textbf{accuracy} \\
\midrule
\citet{lage2022predicting} &                                   BERT-Imbau  &                 pt &                pt &                        - &                     - &              98.0 &                        - &                     - &                 - &               59.0 &            53.0 &              98.0 \\
\citet{lage2022predicting} & Bidirectional Long Short-Term Memory (BiLSTM) &                 pt &                pt &                        - &                     - &          \bf 99.0 &                        - &                     - &                 - &               80.0 &            65.0 &          \bf 99.0 \\
\citet{lage2022predicting} &           Convolutional Neural Networks (CNN) &                 pt &                pt &                        - &                     - &          \bf 99.0 &                        - &                     - &                 - &               88.0 &            69.0 &          \bf 99.0 \\
\citet{lage2022predicting} &                                 decision tree &                 pt &                pt &                        - &                     - &              81.0 &                        - &                     - &                 - &               88.0 &        \bf 77.0 &          \bf 99.0 \\
\citet{lage2022predicting} &           eXtreme Gradient Boosting (XGBoost) &                 pt &                pt &                        - &                     - &              81.0 &                        - &                     - &                 - &           \bf 92.0 &            74.0 &          \bf 99.0 \\
\bottomrule
\end{tabular}
}
\caption{BCD-U. The best scores are in bold.}
\label{paper_results_BCD-U}
\end{table*}
\begin{table*}[h]
\centering
\resizebox{\textwidth}{!}{
\begin{tabular}{lllllllllllll}
\toprule
 \textbf{Source} &                                        \textbf{Method} & \textbf{TrainLang} & \textbf{TestLang} & \textbf{macro-precision} & \textbf{macro-recall} & \textbf{macro-f1} & \textbf{micro-precision} & \textbf{micro-recall} & \textbf{micro-f1} & \textbf{precision} & \textbf{recall} & \textbf{accuracy} \\
\midrule
\citet{tzia2021} &                                            XLM-RoBERTa &                all &               all &                        - &                     - &              59.2 &                        - &                     - &                 - &               62.6 &            60.0 &              54.6 \\
\citet{tzia2021} &                          XLM-RoBERTa pretrained on C19 &                all &               all &                        - &                     - &              59.8 &                        - &                     - &                 - &               55.9 &            62.8 &              57.7 \\
\citet{tzia2021} &                          XLM-RoBERTa pretrained on C19 &                all &             fr-be &                        - &                     - &              72.0 &                        - &                     - &                 - &           \bf 84.9 &            64.5 &                 - \\
\citet{tzia2021} &                          XLM-RoBERTa pretrained on C19 &                all &                pl &                        - &                     - &              58.3 &                        - &                     - &                 - &               53.3 &            66.7 &                 - \\
\citet{tzia2021} &                          XLM-RoBERTa pretrained on C19 &                all &                fr &                        - &                     - &          \bf 81.8 &                        - &                     - &                 - &               82.9 &        \bf 84.7 &                 - \\
\bottomrule
\end{tabular}
}
\caption{C19. The best scores are in bold.}
\label{paper_results_C19}
\end{table*}
\begin{table*}[h]
\centering
\resizebox{\textwidth}{!}{
\begin{tabular}{llllllllllllll}
\toprule
         \textbf{Source} &                        \textbf{Method} & \textbf{TrainLang} & \textbf{TestLang} & \textbf{macro-precision} & \textbf{macro-recall} & \textbf{macro-f1} & \textbf{micro-precision} & \textbf{micro-recall} & \textbf{micro-f1} & \textbf{precision} & \textbf{recall} & \textbf{f1} & \textbf{accuracy} \\
\midrule
\citet{Urchs_et_all2021} &                  tf-idf/decision stump &                 de &                de &                        - &                     - &                 - &                        - &                     - &                 - &               13.0 &            25.0 &        17.0 &              53.0 \\
\citet{Urchs_et_all2021} &             tf-idf/logistic regression &                 de &                de &                        - &                     - &                 - &                        - &                     - &                 - &           \bf 79.0 &            63.0 &        68.0 &              77.0 \\
\citet{Urchs_et_all2021} &          tf-idf/support vector machine &                 de &                de &                        - &                     - &                 - &                        - &                     - &                 - &               74.0 &        \bf 67.0 &    \bf 70.0 &              77.0 \\
\citet{Urchs_et_all2021} &                 Unigram/decision stump &                 de &                de &                        - &                     - &                 - &                        - &                     - &                 - &               13.0 &            25.0 &        17.0 &              53.0 \\
\citet{Urchs_et_all2021} &            Unigram/logistic regression &                 de &                de &                        - &                     - &                 - &                        - &                     - &                 - &               74.0 &        \bf 67.0 &    \bf 70.0 &              77.0 \\
\bottomrule
\end{tabular}
}
\caption{GAM. The best scores are in bold.}
\label{paper_results_GAM}
\end{table*}
\begin{table*}[h]
\centering
\resizebox{\textwidth}{!}{
\begin{tabular}{lllllllllll}
\toprule
  \textbf{Source} &                                  \textbf{Method} & \textbf{TrainLang} & \textbf{TestLang} & \textbf{macro-precision} & \textbf{macro-recall} & \textbf{macro-f1} & \textbf{micro-precision} & \textbf{micro-recall} & \textbf{micro-f1} & \textbf{accuracy} \\
\midrule
\citet{Papa2021m} &                                        BIGRU-ATT &                 el &                el &                        - &                     - &                 - &                     81.0 &                  81.0 &              81.0 &                 - \\
\citet{Papa2021m} &                                       BIGRU-LWAN &                 el &                el &                        - &                     - &                 - &                     77.0 &                  77.0 &              77.0 &                 - \\
\citet{Papa2021m} &                                        BIGRU-MAX &                 el &                el &                        - &                     - &                 - &                     78.0 &                  78.0 &              78.0 &                 - \\
\citet{Papa2021m} &                                      GREEK-BERT  &                 el &                el &                        - &                     - &                 - &                     82.0 &                  82.0 &              82.0 &                 - \\
\citet{Papa2021m} &                                GREEK-LEGAL-BERT  &                 el &                el &                        - &                     - &                 - &                     84.0 &                  84.0 &              84.0 &                 - \\
\citet{Papa2021m} &                                           MBERT  &                 el &                el &                        - &                     - &                 - &                     80.0 &                  80.0 &              80.0 &                 - \\
\bottomrule
\end{tabular}
}
\caption{GLC-C. The best scores are in bold.}
\label{paper_results_GLC-C}
\end{table*}

\begin{table*}[h]
\centering
\resizebox{\textwidth}{!}{
\begin{tabular}{lllllllllll}
\toprule
  \textbf{Source} &                                  \textbf{Method} & \textbf{TrainLang} & \textbf{TestLang} & \textbf{macro-precision} & \textbf{macro-recall} & \textbf{macro-f1} & \textbf{micro-precision} & \textbf{micro-recall} & \textbf{micro-f1} & \textbf{accuracy} \\
\midrule
\citet{Papa2021m} &                                        BIGRU-ATT &                 el &                el &                        - &                     - &                 - &                     75.0 &                  75.0 &              75.0 &                 - \\
\citet{Papa2021m} &                                       BIGRU-LWAN &                 el &                el &                        - &                     - &                 - &                     65.0 &                  65.0 &              65.0 &                 - \\
\citet{Papa2021m} &                                        BIGRU-MAX &                 el &                el &                        - &                     - &                 - &                     63.0 &                  63.0 &              63.0 &                 - \\
\citet{Papa2021m} &                                      GREEK-BERT  &                 el &                el &                        - &                     - &                 - &                     79.0 &                  79.0 &              79.0 &                 - \\
\citet{Papa2021m} &                                GREEK-LEGAL-BERT  &                 el &                el &                        - &                     - &                 - &                 \bf 81.0 &              \bf 81.0 &          \bf 81.0 &                 - \\
\citet{Papa2021m} &                                           MBERT  &                 el &                el &                        - &                     - &                 - &                     77.0 &                  77.0 &              77.0 &                 - \\
\bottomrule
\end{tabular}
}
\caption{GLC-S. The best scores are in bold.}
\label{paper_results_GLC-S}
\end{table*}
\begin{table*}[h]
\centering
\resizebox{\textwidth}{!}{
\begin{tabular}{lllllllllll}
\toprule
  \textbf{Source} &                                  \textbf{Method} & \textbf{TrainLang} & \textbf{TestLang} & \textbf{macro-precision} & \textbf{macro-recall} & \textbf{macro-f1} & \textbf{micro-precision} & \textbf{micro-recall} & \textbf{micro-f1} & \textbf{accuracy} \\
\midrule
\citet{Papa2021m} &                                        BIGRU-ATT &                 el &                el &                        - &                     - &                 - &                     86.0 &                  86.0 &              86.0 &                 - \\
\citet{Papa2021m} &                                       BIGRU-LWAN &                 el &                el &                        - &                     - &                 - &                     84.0 &                  84.0 &              84.0 &                 - \\
\citet{Papa2021m} &                                        BIGRU-MAX &                 el &                el &                        - &                     - &                 - &                     84.0 &                  84.0 &              84.0 &                 - \\
\citet{Papa2021m} &                                      GREEK-BERT  &                 el &                el &                        - &                     - &                 - &                     88.0 &                  88.0 &              88.0 &                 - \\
\citet{Papa2021m} &                                GREEK-LEGAL-BERT  &                 el &                el &                        - &                     - &                 - &                     89.0 &                  89.0 &              89.0 &                 - \\
\citet{Papa2021m} &                                           MBERT  &                 el &                el &                        - &                     - &                 - &                     86.0 &                  86.0 &              86.0 &                 - \\
\bottomrule
\end{tabular}
}
\caption{GLC-V. The best scores are in bold.}
\label{paper_results_GLC-V}
\end{table*}
\begin{table*}[h]
\centering
\resizebox{\textwidth}{!}{
\begin{tabular}{lllllllllll}
\toprule
             \textbf{Source} &                        \textbf{Method} & \textbf{TrainLang} & \textbf{TestLang} & \textbf{macro-precision} & \textbf{macro-recall} & \textbf{macro-f1} & \textbf{micro-precision} & \textbf{micro-recall} & \textbf{micro-f1} & \textbf{accuracy} \\
\midrule
\citet{Angelidis2018NamedER} &                       BILSTM-BILSTM-LR &                 el &                el &                 \bf 91.0 &              \bf 85.0 &          \bf 88.0 &                        - &                     - &                 - &                 - \\
\citet{Angelidis2018NamedER} &                             BILSTM-CRF &                 el &                el &                     87.0 &                  80.0 &              83.0 &                        - &                     - &                 - &                 - \\
\citet{Angelidis2018NamedER} &                              BILSTM-LR &                 el &                el &                     89.0 &                  79.0 &              84.0 &                        - &                     - &                 - &                 - \\
                    lextreme &     distilbert-base-multilingual-cased &                 el &                el &                     67.5 &                  76.0 &              71.0 &                     97.3 &                  96.8 &              97.0 &              96.6 \\
                    lextreme &             microsoft/mdeberta-v3-base &                 el &                el &                     73.5 &                  74.5 &              73.3 &                     97.4 &                  97.1 &              97.3 &              96.9 \\
                    lextreme & microsoft/Multilingual-MiniLM-L12-H384 &                 el &                el &                     75.3 &                  73.2 &              74.0 &                     97.5 &                  97.2 &              97.3 &              96.9 \\
\bottomrule
\end{tabular}
}
\caption{GLN. The best scores are in bold.}
\label{paper_results_GLN}
\end{table*}
\begin{table*}[h]
\centering
\resizebox{\textwidth}{!}{
\begin{tabular}{lllllllllll}
\toprule
                 \textbf{Source} &                                                                    \textbf{Method} & \textbf{TrainLang} & \textbf{TestLang} & \textbf{macro-precision} & \textbf{macro-recall} & \textbf{macro-f1} & \textbf{micro-precision} & \textbf{micro-recall} & \textbf{micro-f1} & \textbf{accuracy} \\
\midrule
\citet{pais_vasile_2021_4922385} &                        CoRoLa word embeddings + MARCELL word embeddings+BiLSTM-CRF &                 ro &                ro &                        - &                     - &              84.7 &                        - &                     - &                 - &                 - \\
\citet{pais_vasile_2021_4922385} &           CoRoLa word embeddings + MARCELL word embeddings+BiLSTM-CRF + gazetteers &                 ro &                ro &                        - &                     - &              84.8 &                        - &                     - &                 - &                 - \\
\citet{pais_vasile_2021_4922385} & CoRoLa word embeddings + MARCELL word embeddings+BiLSTM-CRF + gazetteers + affixes &                 ro &                ro &                        - &                     - &              83.4 &                        - &                     - &                 - &                 - \\
\citet{pais_vasile_2021_4922385} &                                                  CoRoLa word embeddings+BiLSTM-CRF &                 ro &                ro &                        - &                     - &              83.9 &                        - &                     - &                 - &                 - \\
\citet{pais_vasile_2021_4922385} &                                     CoRoLa word embeddings+BiLSTM-CRF + gazetteers &                 ro &                ro &                        - &                     - &              85.0 &                        - &                     - &                 - &                 - \\
\bottomrule
\end{tabular}
}
\caption{LNR. The best scores are in bold.}
\label{paper_results_LNR}
\end{table*}
\begin{table*}[h]
\centering
\resizebox{\textwidth}{!}{
\begin{tabular}{llllllllllllll}
\toprule
 \textbf{Source} &                                              \textbf{Method} & \textbf{TrainLang} & \textbf{TestLang} & \textbf{macro-precision} & \textbf{macro-recall} & \textbf{macro-f1} & \textbf{micro-precision} & \textbf{micro-recall} & \textbf{micro-f1} & \textbf{precision} & \textbf{recall} & \textbf{f1} & \textbf{accuracy} \\
\midrule
\citet{luz2018a} & LSTM-CRF (long short-term memory + conditional random field) &                 pt &                pt &                        - &                     - &                 - &                        - &                     - &                 - &           \bf 93.2 &        \bf 91.9 &    \bf 92.5 &                 - \\
        lextreme &                           distilbert-base-multilingual-cased &                 pt &                pt &                     89.9 &                  89.6 &              89.6 &                     98.5 &                  98.6 &              98.6 &                  - &               - &           - &              98.0 \\
        lextreme &                                   microsoft/mdeberta-v3-base &                 pt &                pt &                     94.7 &                  94.9 &              94.8 &                     99.0 &              \bf 99.4 &          \bf 99.2 &                  - &               - &           - &              98.8 \\
        lextreme &                       microsoft/Multilingual-MiniLM-L12-H384 &                 pt &                pt &                     93.9 &                  93.4 &              93.6 &                     98.9 &                  99.2 &              99.1 &                  - &               - &           - &              98.7 \\
\bottomrule
\end{tabular}
}
\caption{LNB. The best scores are in bold.}
\label{paper_results_LNB}
\end{table*}

\begin{table*}[h]
\centering
\resizebox{\textwidth}{!}{
\begin{tabular}{llllllllllll}
\toprule
      \textbf{Source} &                                \textbf{Method} & \textbf{TrainLang} & \textbf{TestLang} & \textbf{macro-precision} & \textbf{macro-recall} & \textbf{macro-f1} & \textbf{micro-precision} & \textbf{micro-recall} & \textbf{micro-f1} & \textbf{accuracy} & \textbf{mean r-precision} \\
\midrule
\citet{chalkidis-etal-2021-multieurlex} &                              xlm-roberta-base  &                all &               all &                        - &                     - &                 - &                        - &                     - &                 - &                 - &                      65.7 \\
\citet{chalkidis-etal-2021-multieurlex} &                              xlm-roberta-base  &                all &                en &                        - &                     - &                 - &                        - &                     - &                 - &                 - &                      66.4 \\
\citet{chalkidis-etal-2021-multieurlex} &                              xlm-roberta-base  &                all &                da &                        - &                     - &                 - &                        - &                     - &                 - &                 - &                      66.2 \\
\citet{chalkidis-etal-2021-multieurlex} &                              xlm-roberta-base  &                all &                de &                        - &                     - &                 - &                        - &                     - &                 - &                 - &                      66.2 \\
\citet{chalkidis-etal-2021-multieurlex} &                              xlm-roberta-base  &                all &                nl &                        - &                     - &                 - &                        - &                     - &                 - &                 - &                      66.1 \\
\bottomrule
\end{tabular}
}
\caption{MEU-3. The best scores are in bold.}
\label{paper_results_MEU-3}
\end{table*}
\begin{table*}[h]
\centering
\resizebox{\textwidth}{!}{
\begin{tabular}{lllllllllll}
\toprule
  \textbf{Source} &                         \textbf{Method} & \textbf{TrainLang} & \textbf{TestLang} & \textbf{macro-precision} & \textbf{macro-recall} & \textbf{macro-f1} & \textbf{micro-precision} & \textbf{micro-recall} & \textbf{micro-f1} & \textbf{accuracy} \\
\midrule
\citet{Nikl2021s} &                            French BERT  &                 fr &                fr &                        - &                     - &              58.6 &                        - &                     - &              74.7 &                 - \\
\citet{Nikl2021s} &               French Hierarchical BERT  &                 fr &                fr &                        - &                     - &          \bf 70.2 &                        - &                     - &              80.2 &                 - \\
\citet{Nikl2021s} &                       French Long BERT  &                 fr &                fr &                        - &                     - &              68.0 &                        - &                     - &              77.2 &                 - \\
\citet{Nikl2021s} &               German Hierarchical BERT  &                 de &                de &                        - &                     - &              68.5 &                        - &                     - &              77.1 &                 - \\
\citet{Nikl2021s} &                       German Long BERT  &                 de &                de &                        - &                     - &              67.9 &                        - &                     - &              76.5 &                 - \\
\citet{Nikl2021s} &                            German-BERT  &                 de &                de &                        - &                     - &              63.7 &                        - &                     - &              74.0 &                 - \\
\bottomrule
\end{tabular}
}
\caption{SJP. The best scores are in bold.}
\label{paper_results_SJP}
\end{table*}

\newpage
\afterpage{\clearpage}

\section{Histograms}
\label{sec:histograms}

In the following, we provide the histograms for the distribution of the sequence length of the input (sentence or entire document) from each dataset. The length is measured by counting the tokens using the tokenizers of the multilingual models, i.e., DistilBERT, MiniLM, mDeBERTa v3, XLM-R base, XLM-R large. We only display the distribution within the 99th percentile; the rest is grouped together at the end.

\begin{figure}[h]
\centering
\includegraphics[width=\textwidth]{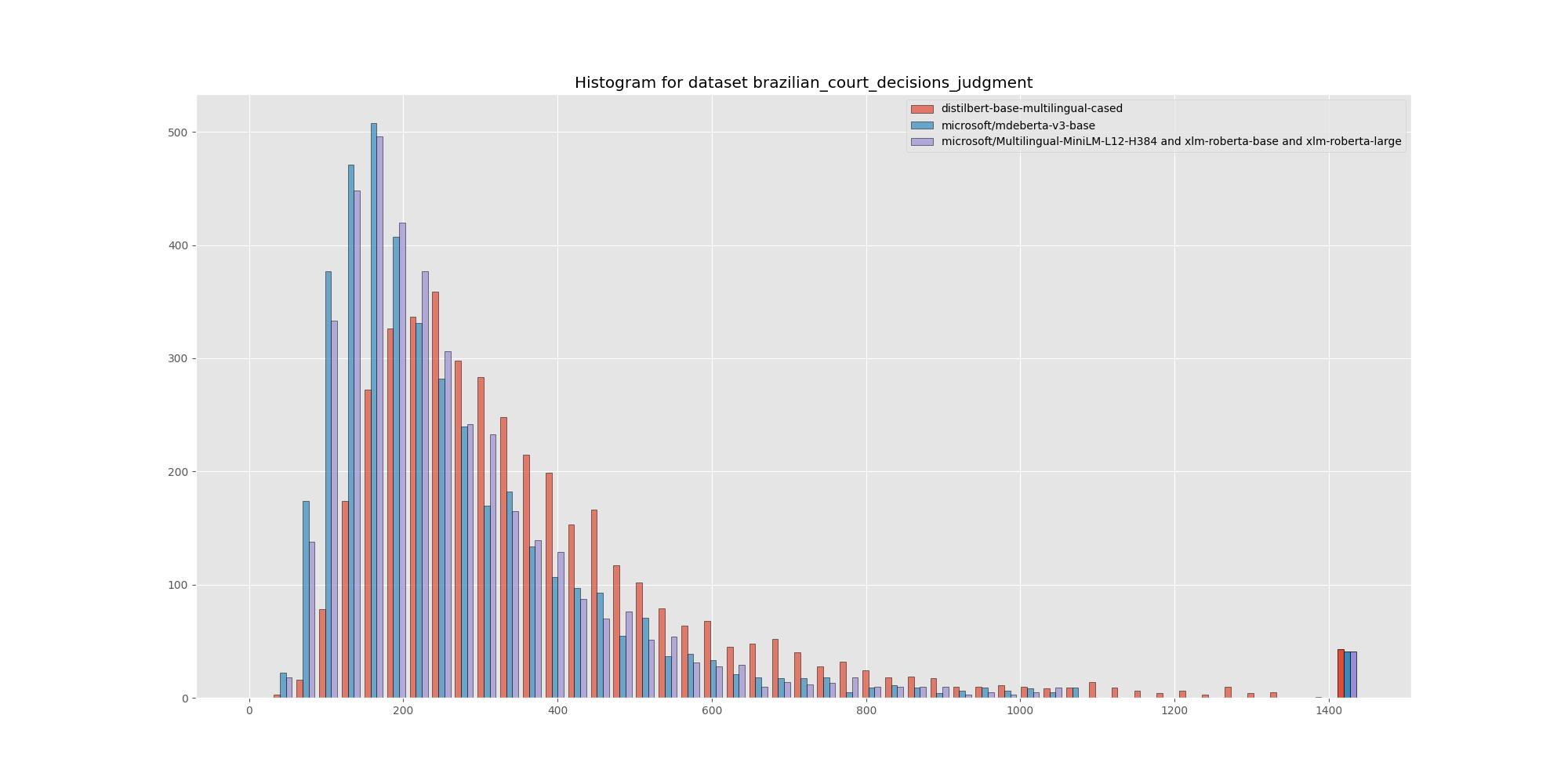}
\caption{Histogram for dataset BCD-J}
\label{histogram_BCD-J}
\end{figure}

\begin{figure}[h]
\centering
\includegraphics[width=\textwidth]{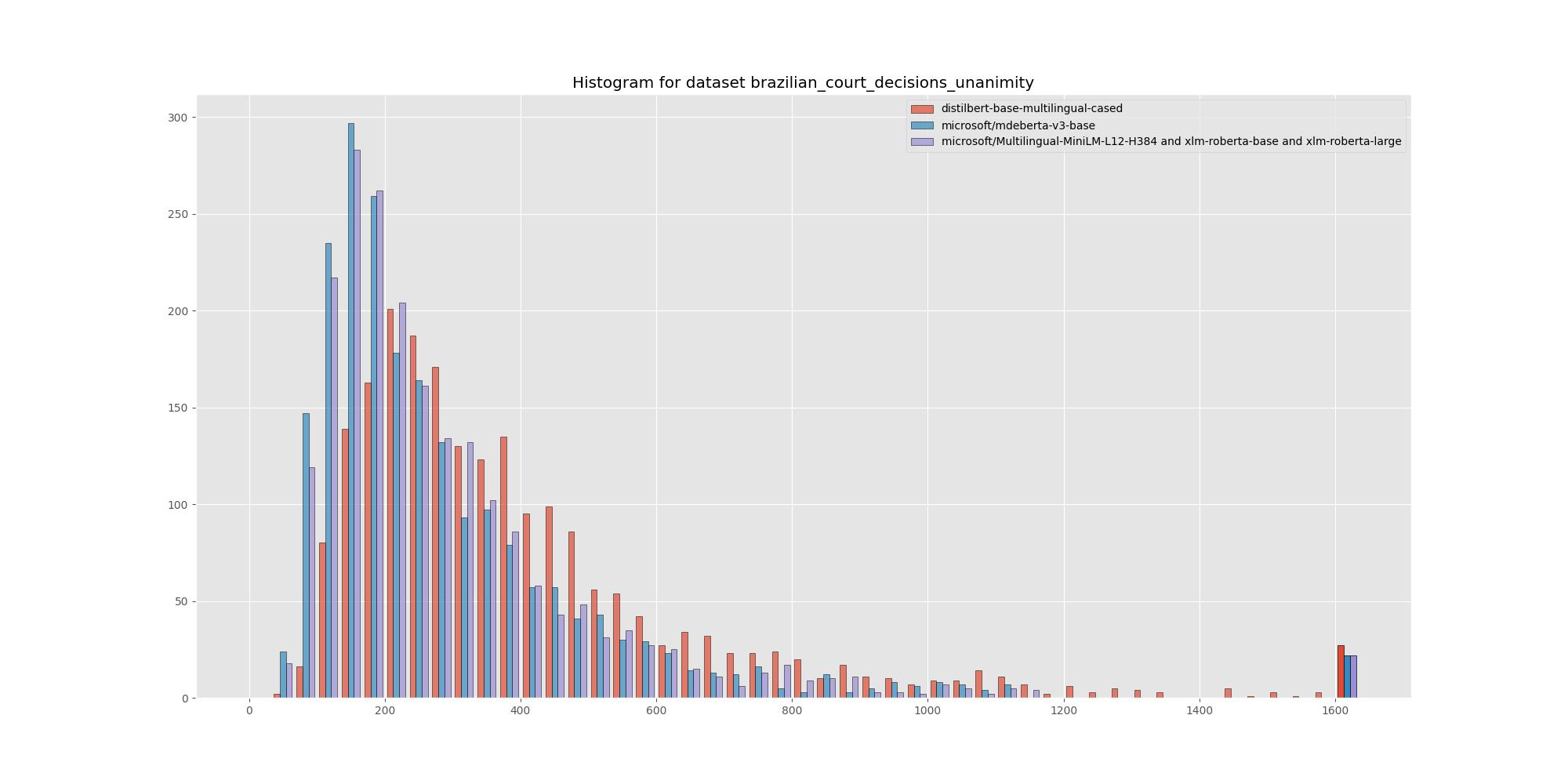}
\caption{Histogram for dataset BCD-U}
\label{histogram_BCD-U}
\end{figure}

\begin{figure}[h]
\centering
\includegraphics[width=\textwidth]{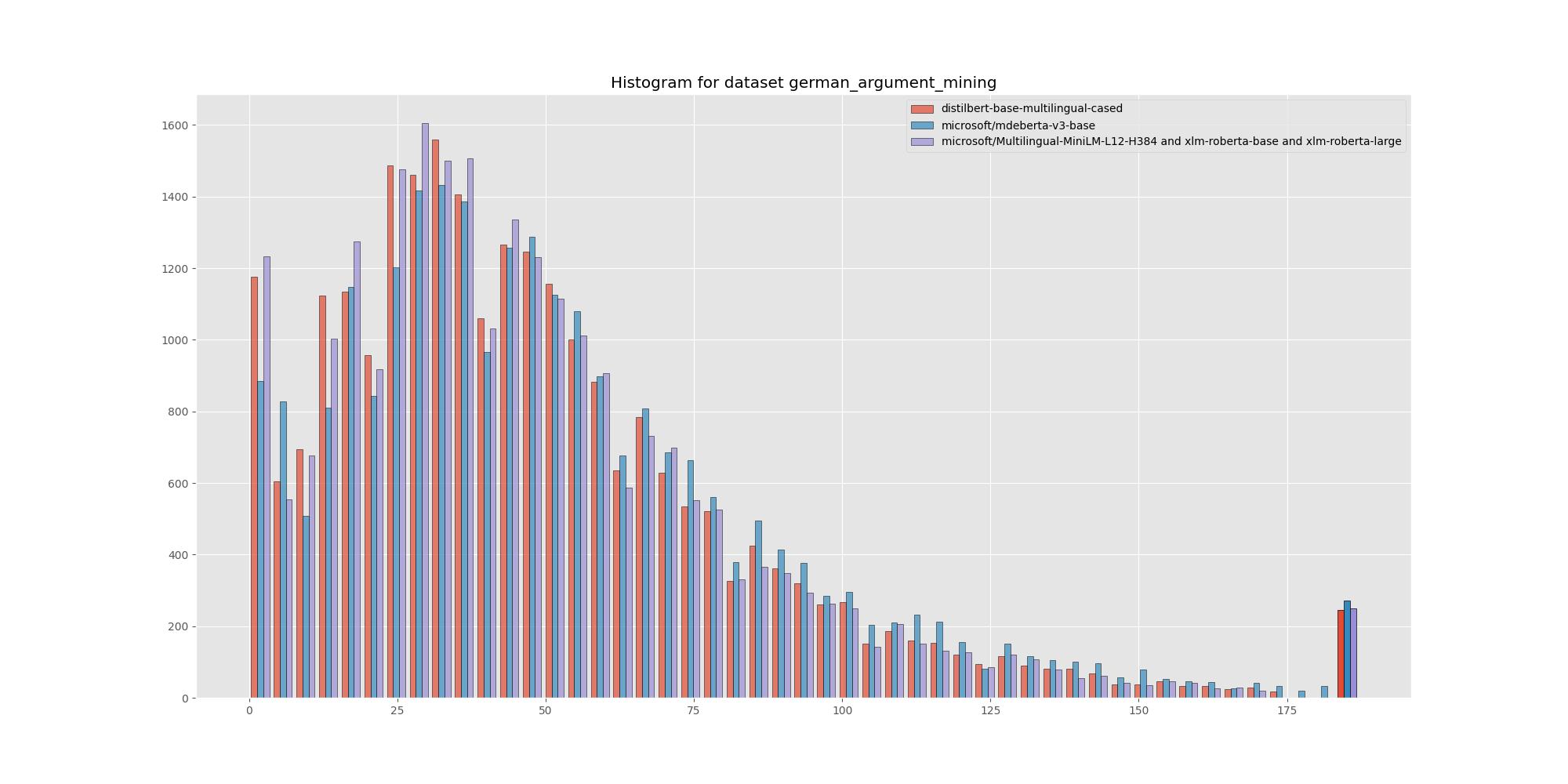}
\caption{Histogram for dataset GAM}
\label{histogram_GAM}
\end{figure}

\begin{figure}[h]
\centering
\includegraphics[width=\textwidth]{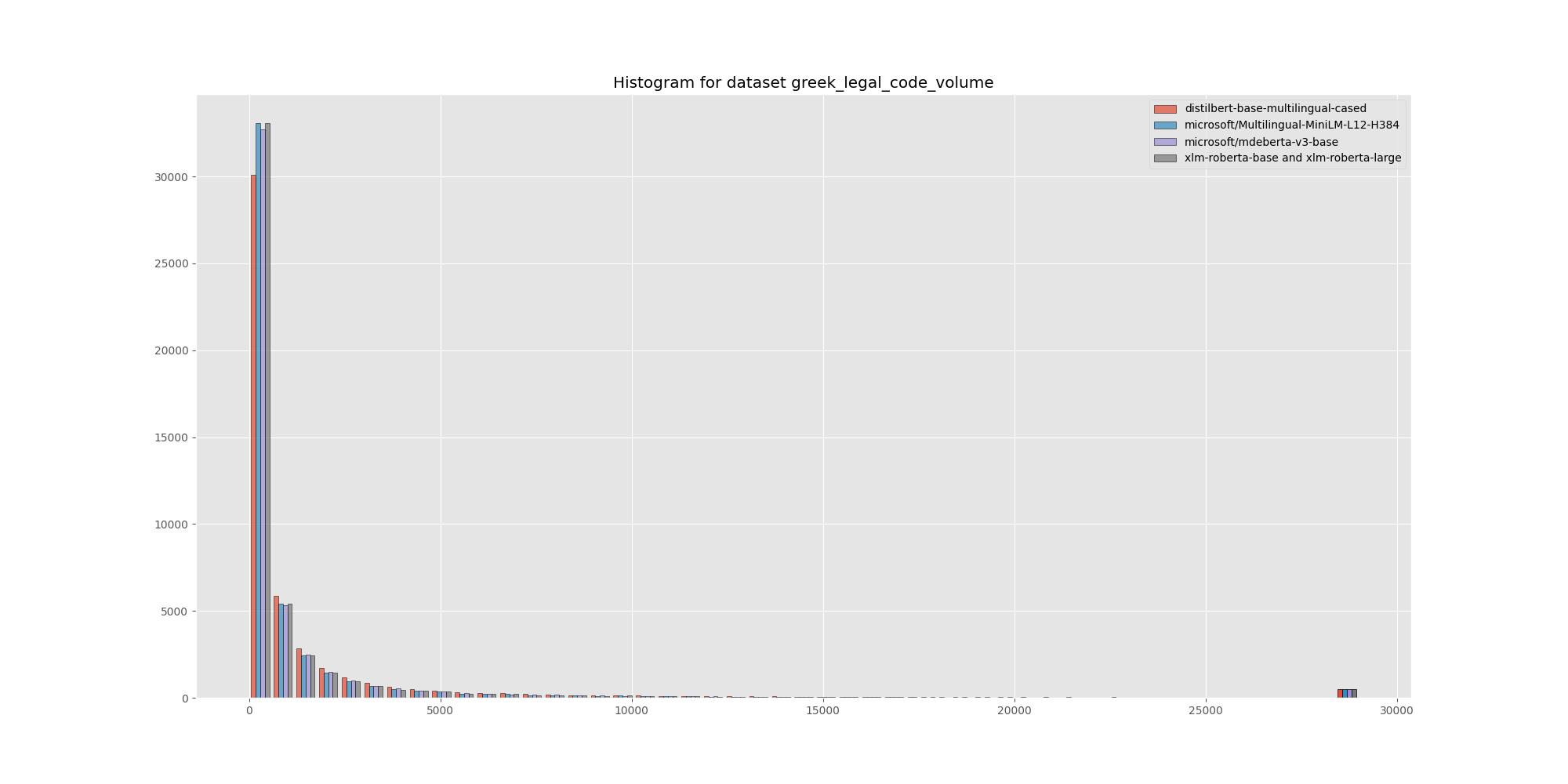}
\caption{Histogram for dataset GLC-V}
\label{histogram_GLC-V}
\end{figure}

\begin{figure}[h]
\centering
\includegraphics[width=\textwidth]{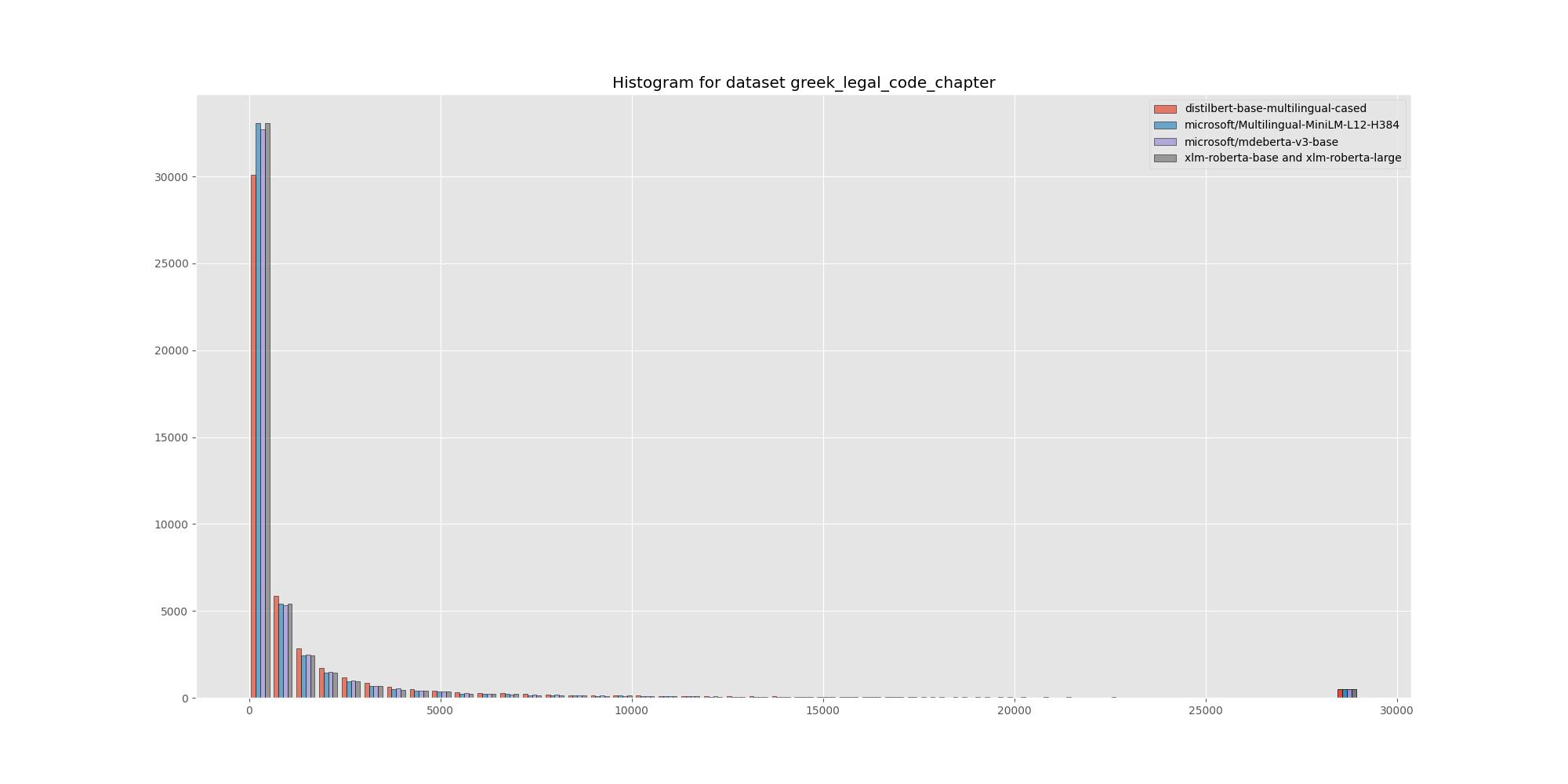}
\caption{Histogram for dataset GLC-C}
\label{histogram_GLC-C}
\end{figure}

\begin{figure}[h]
\centering
\includegraphics[width=\textwidth]{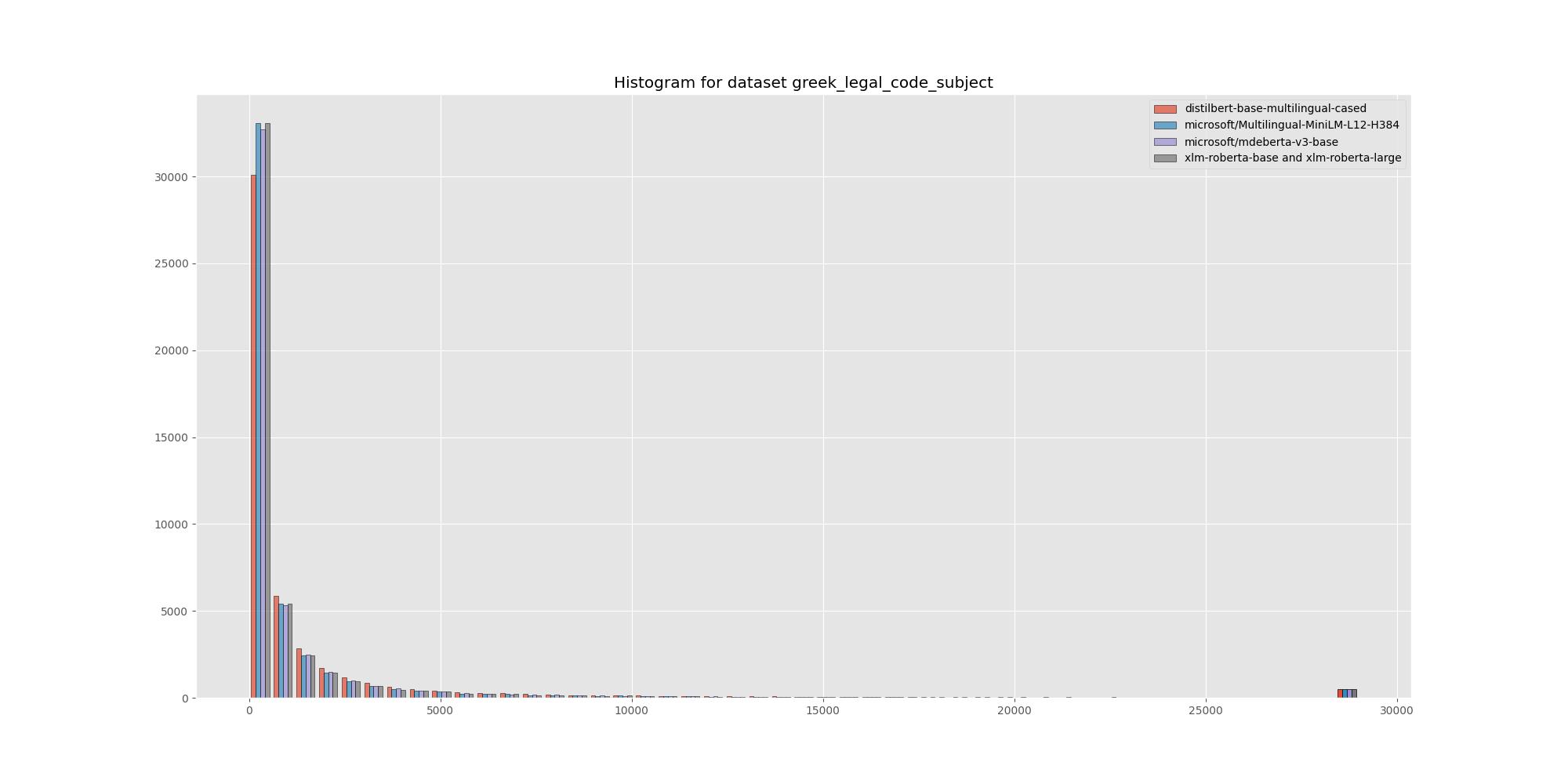}
\caption{Histogram for dataset GLC-S}
\label{histogram_GLC-S}
\end{figure}

\begin{figure}[h]
\centering
\includegraphics[width=\textwidth]{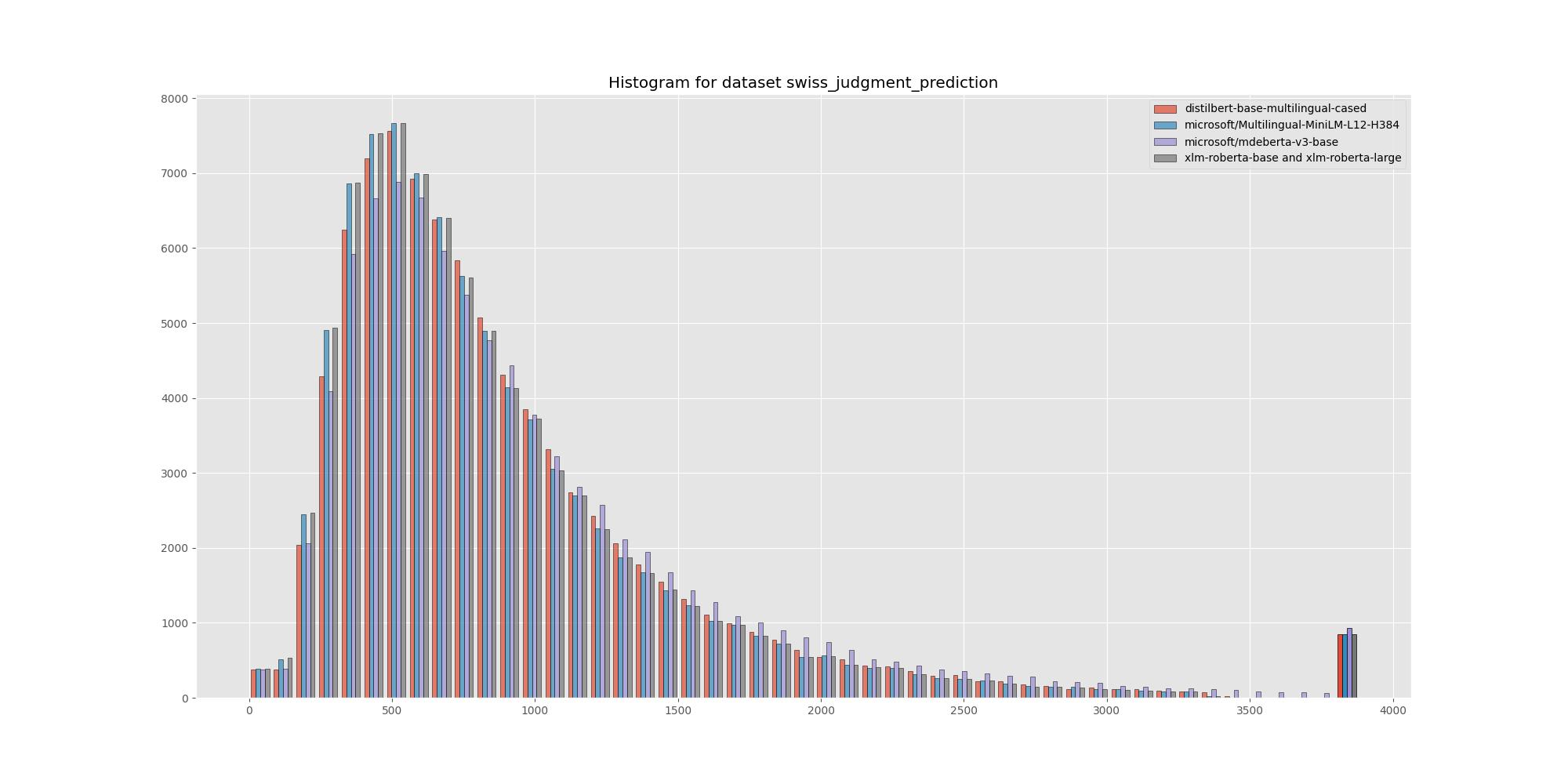}
\caption{Histogram for dataset SJP}
\label{histogram_SJP}
\end{figure}

\begin{figure}[h]
\centering
\includegraphics[width=\textwidth]{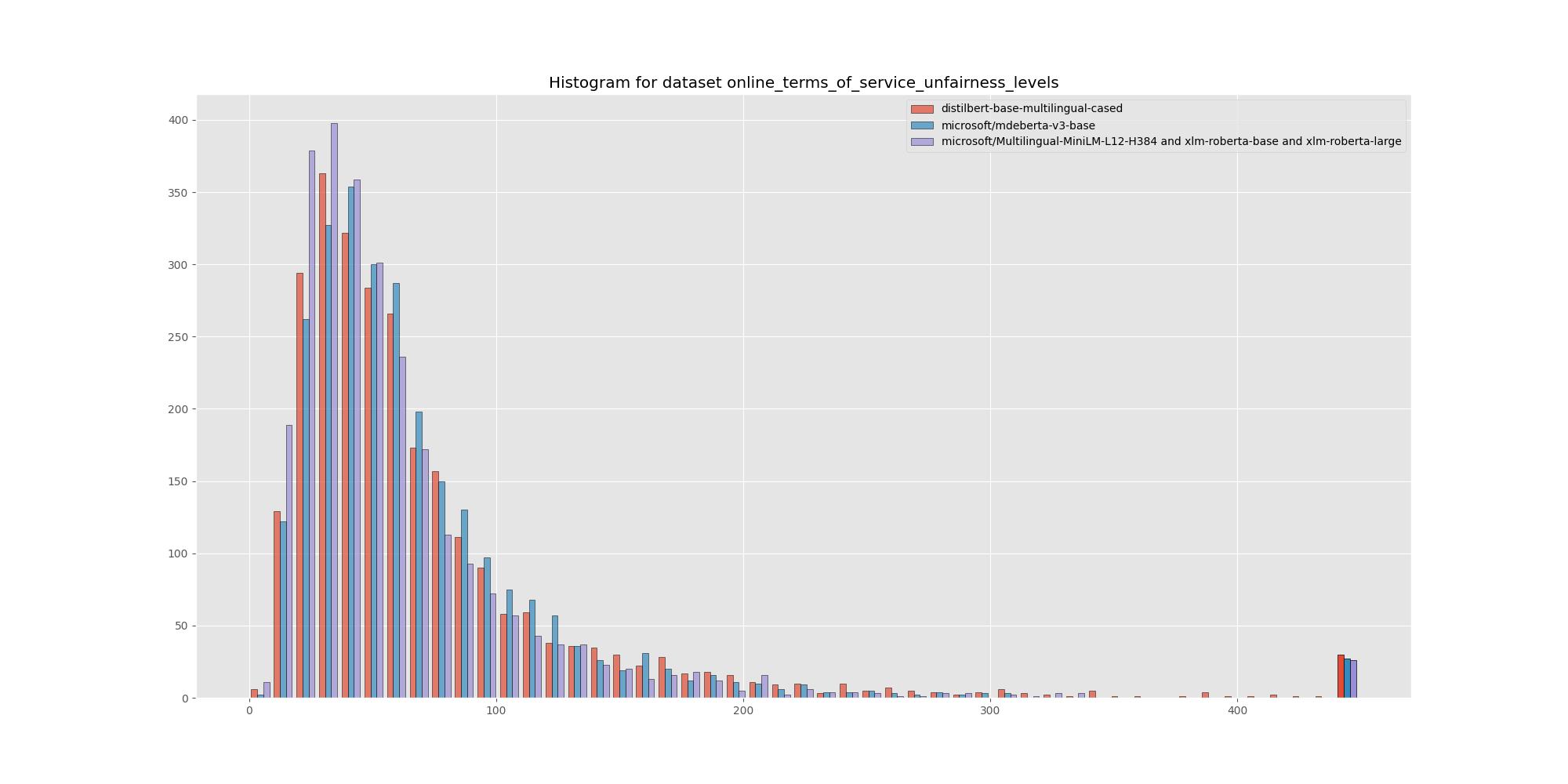}
\caption{Histogram for dataset OTS-UL}
\label{histogram_OTS-UL}
\end{figure}

\begin{figure}[h]
\centering
\includegraphics[width=\textwidth]{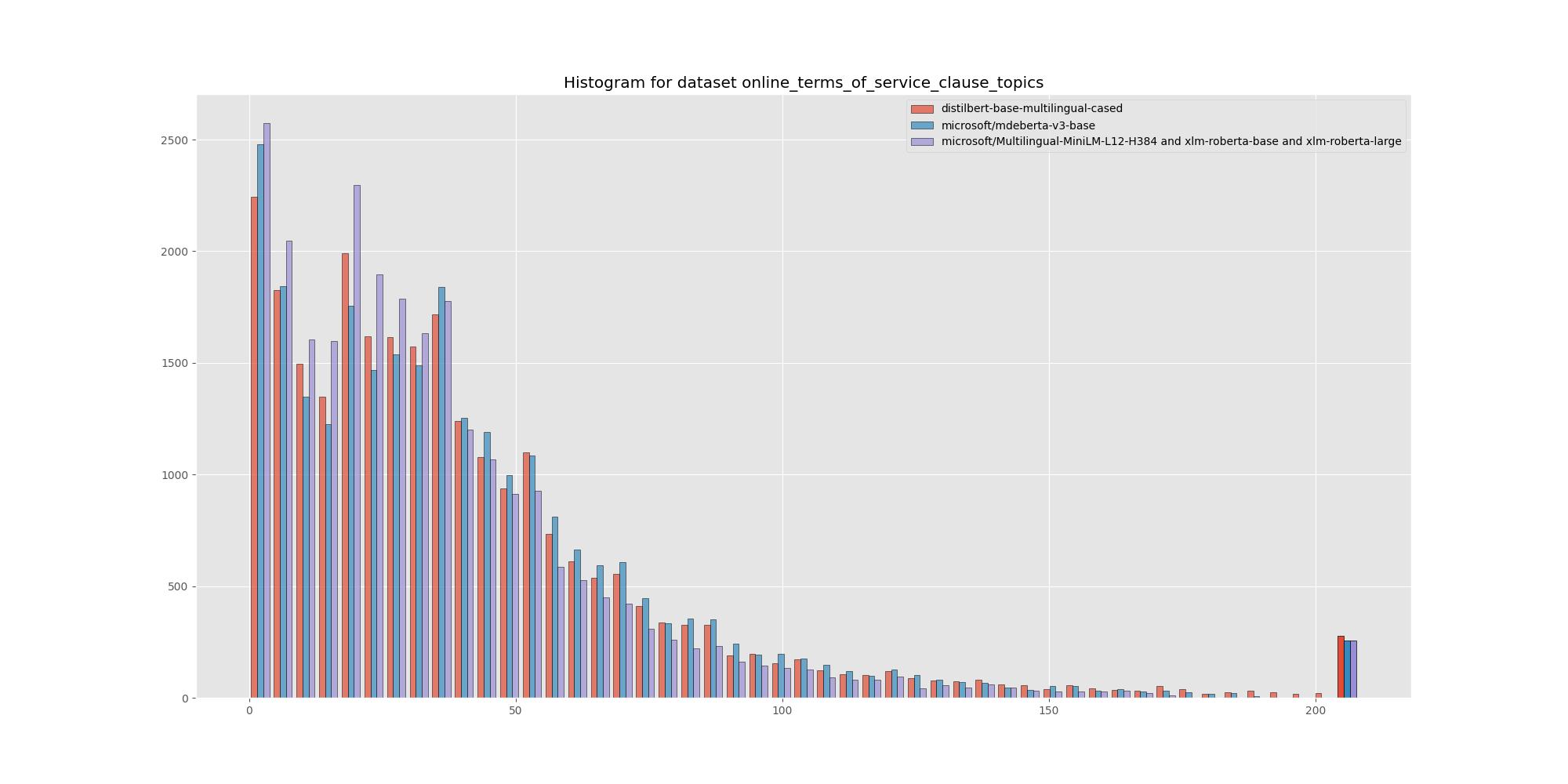}
\caption{Histogram for dataset OTS-CT}
\label{histogram_OTS-CT}
\end{figure}

\begin{figure}[h]
\centering
\includegraphics[width=\textwidth]{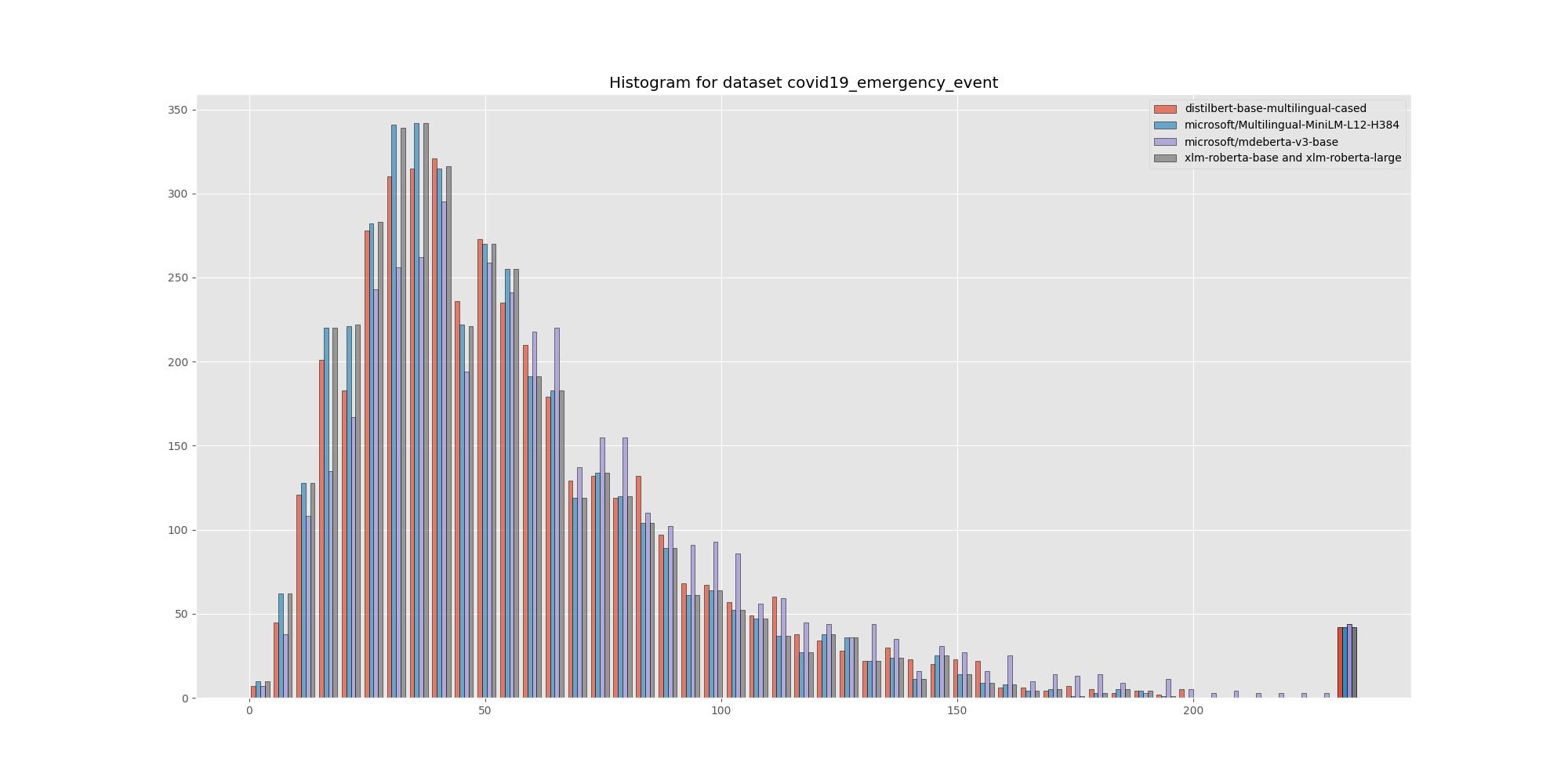}
\caption{Histogram for dataset C19}
\label{histogram_C19}
\end{figure}

\begin{figure}[h]
\centering
\includegraphics[width=\textwidth]{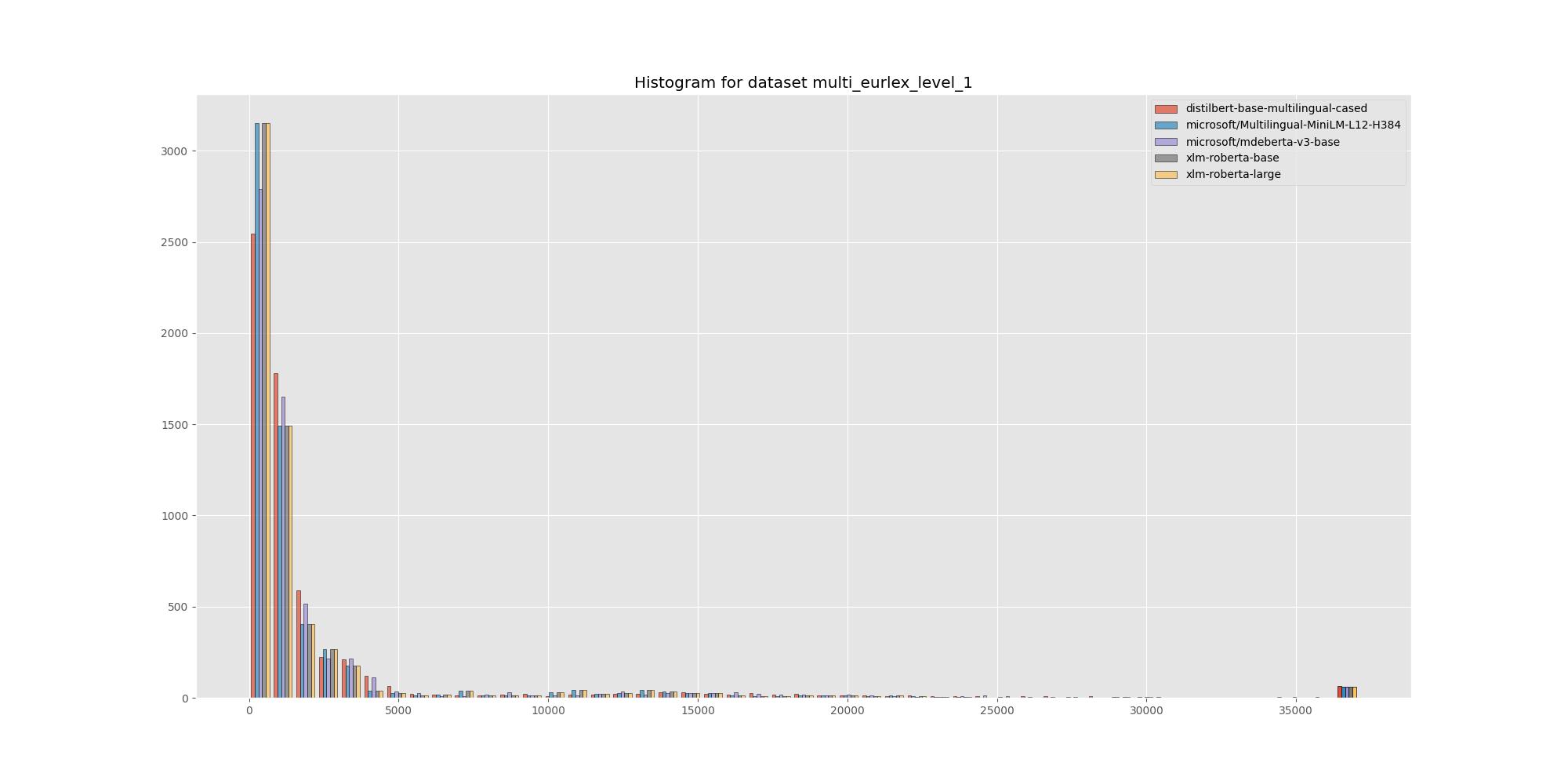}
\caption{Histogram for dataset MEU-1}
\label{histogram_MEU-1}
\end{figure}

\begin{figure}[h]
\centering
\includegraphics[width=\textwidth]{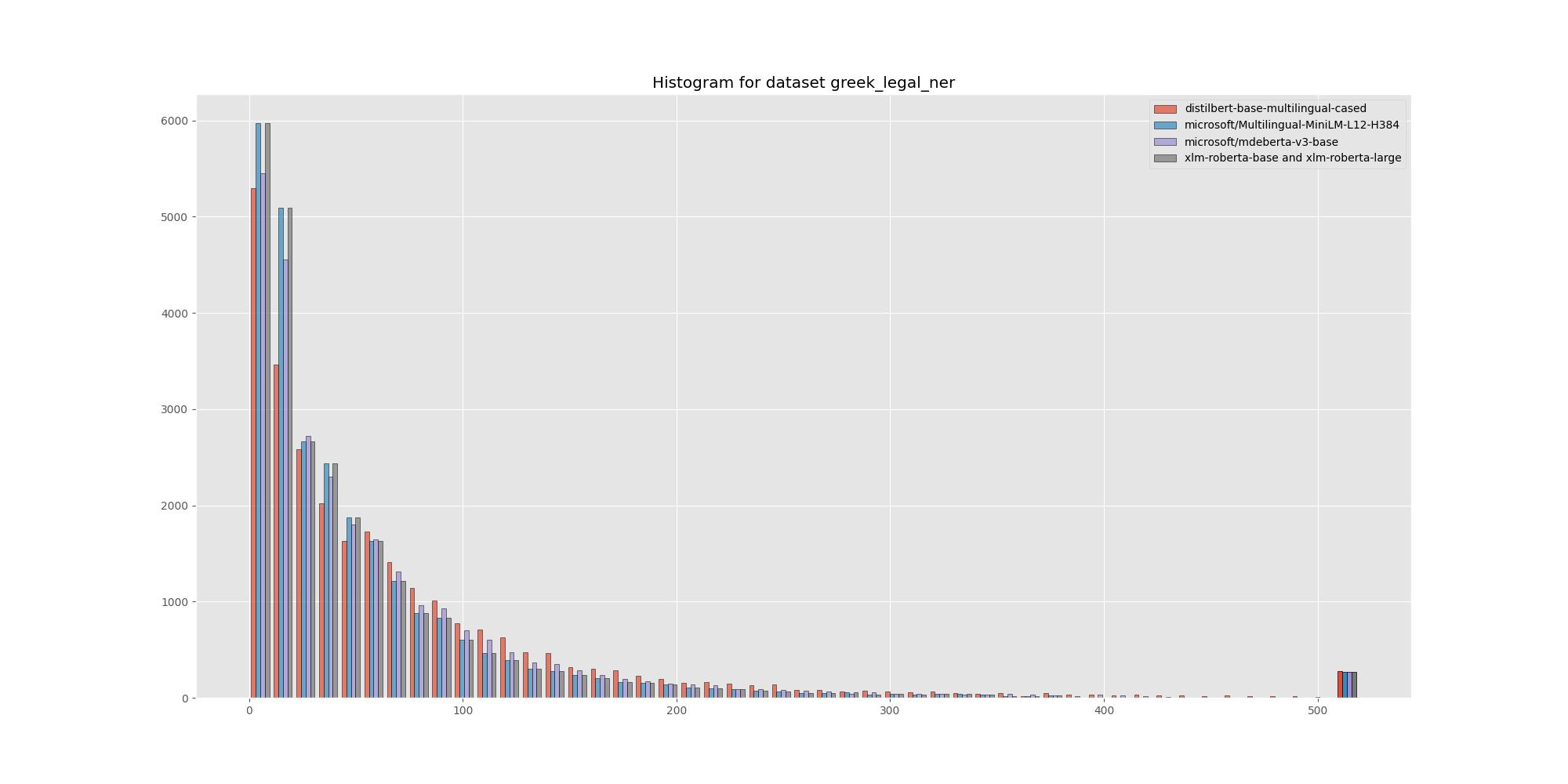}
\caption{Histogram for dataset GLN}
\label{histogram_GLN}
\end{figure}

\begin{figure}[h]
\centering
\includegraphics[width=\textwidth]{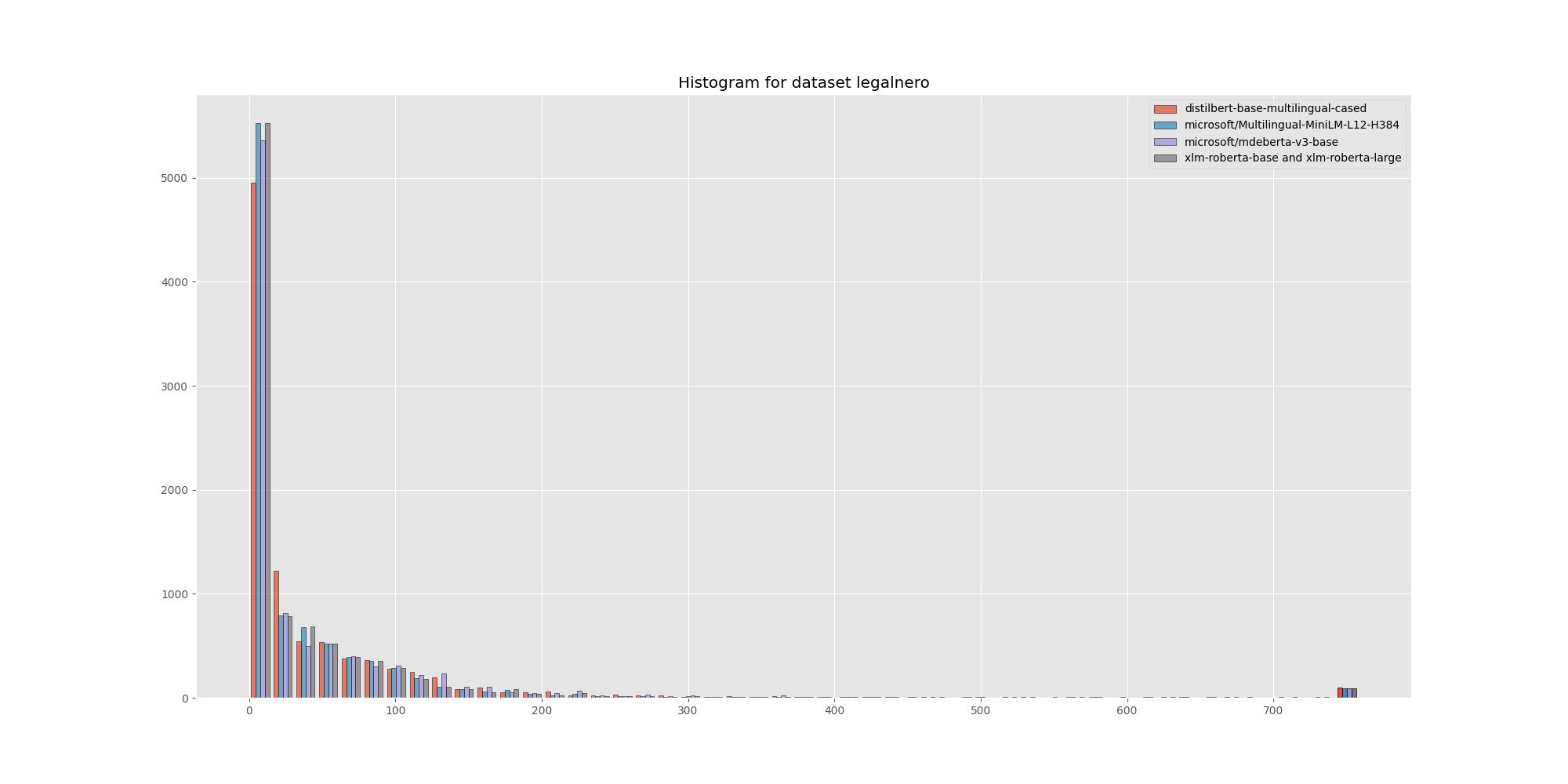}
\caption{Histogram for dataset LNR}
\label{histogram_LNR}
\end{figure}

\begin{figure}[h]
\centering
\includegraphics[width=\textwidth]{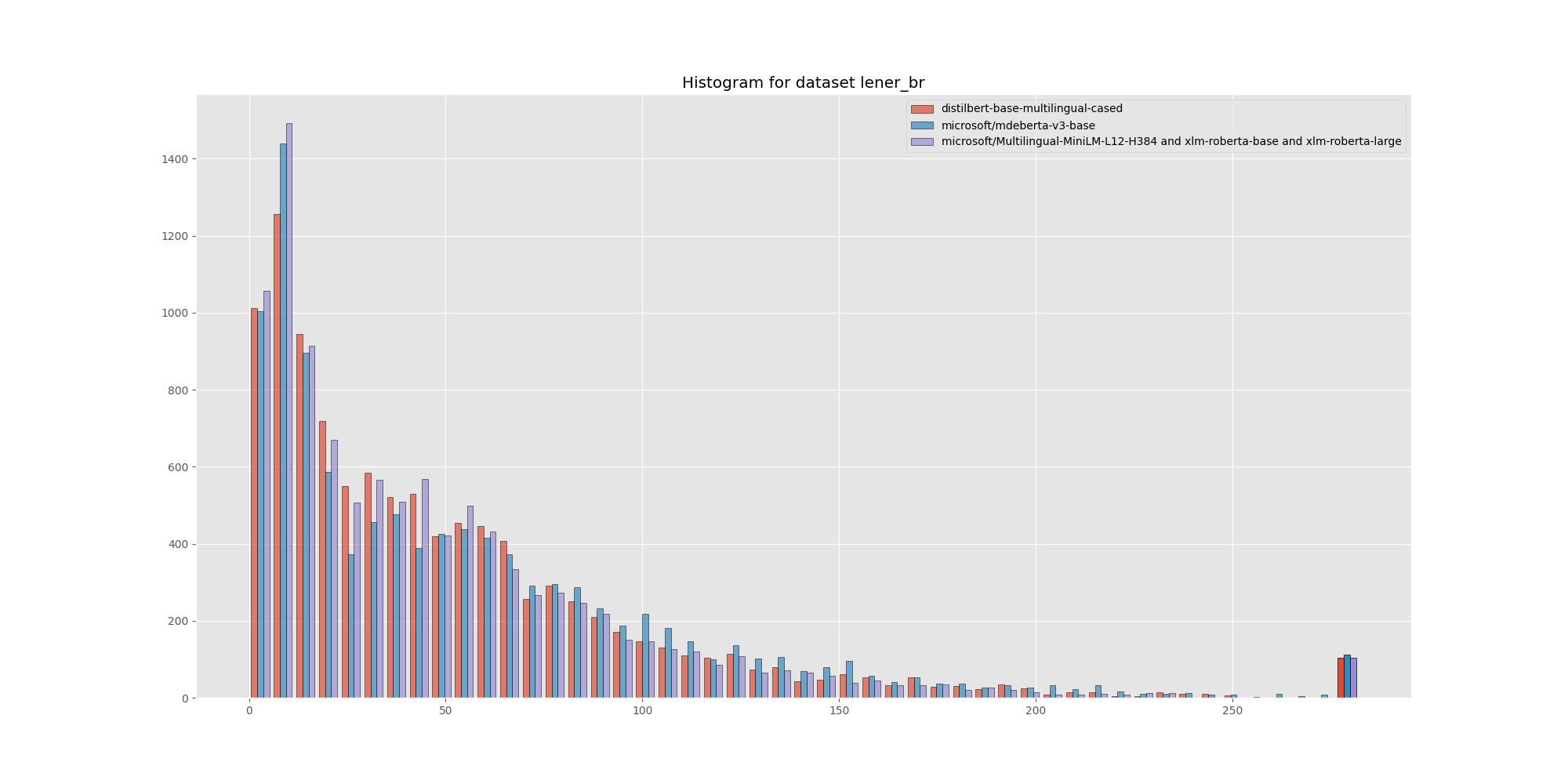}
\caption{Histogram for dataset LNB}
\label{histogram_LNB}
\end{figure}

\begin{figure}[h]
\centering
\includegraphics[width=\textwidth]{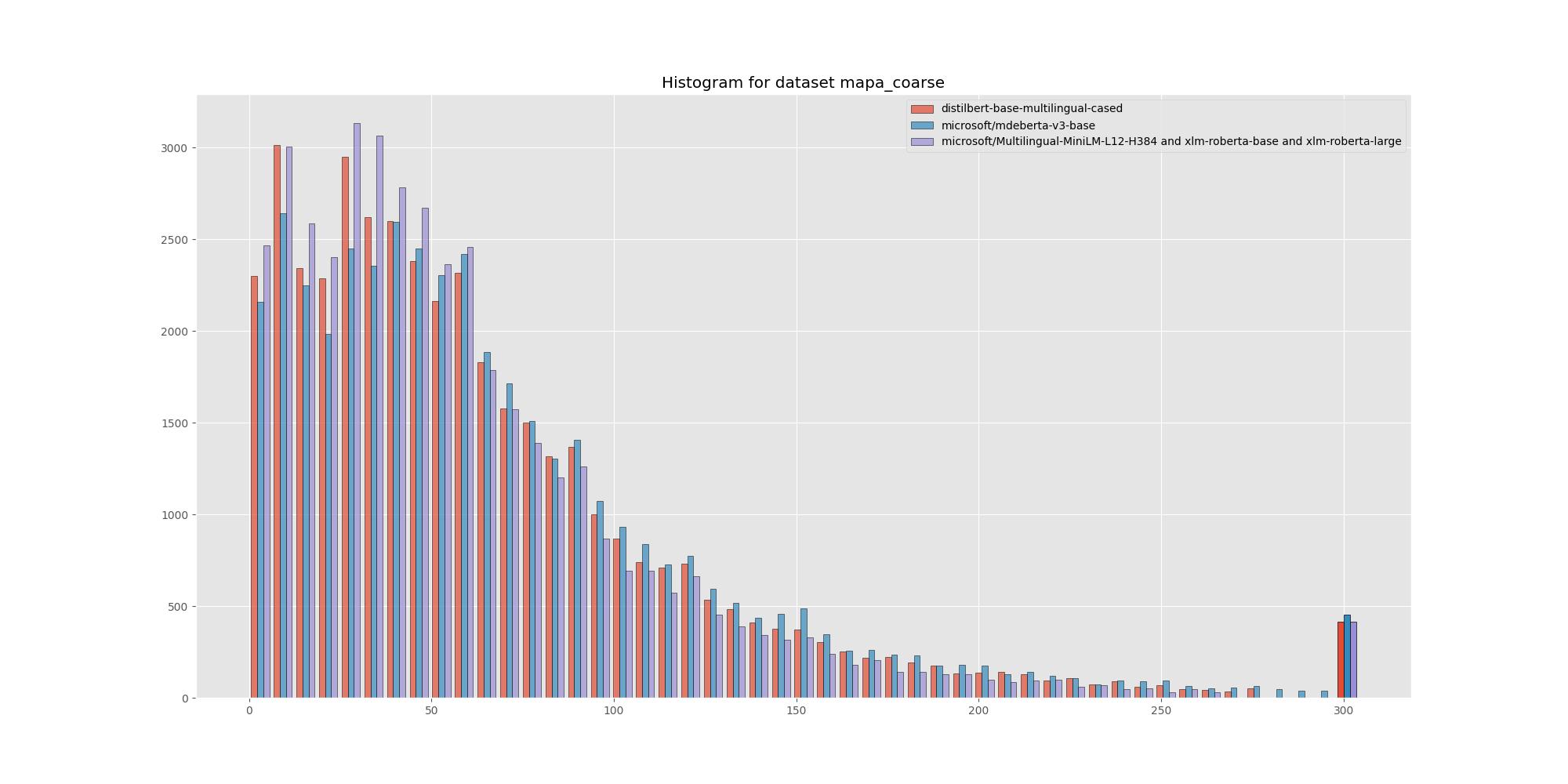}
\caption{Histogram for dataset MAP-C}
\label{histogram_MAP-C}
\end{figure}

\begin{figure}[h]
\centering
\includegraphics[width=\textwidth]{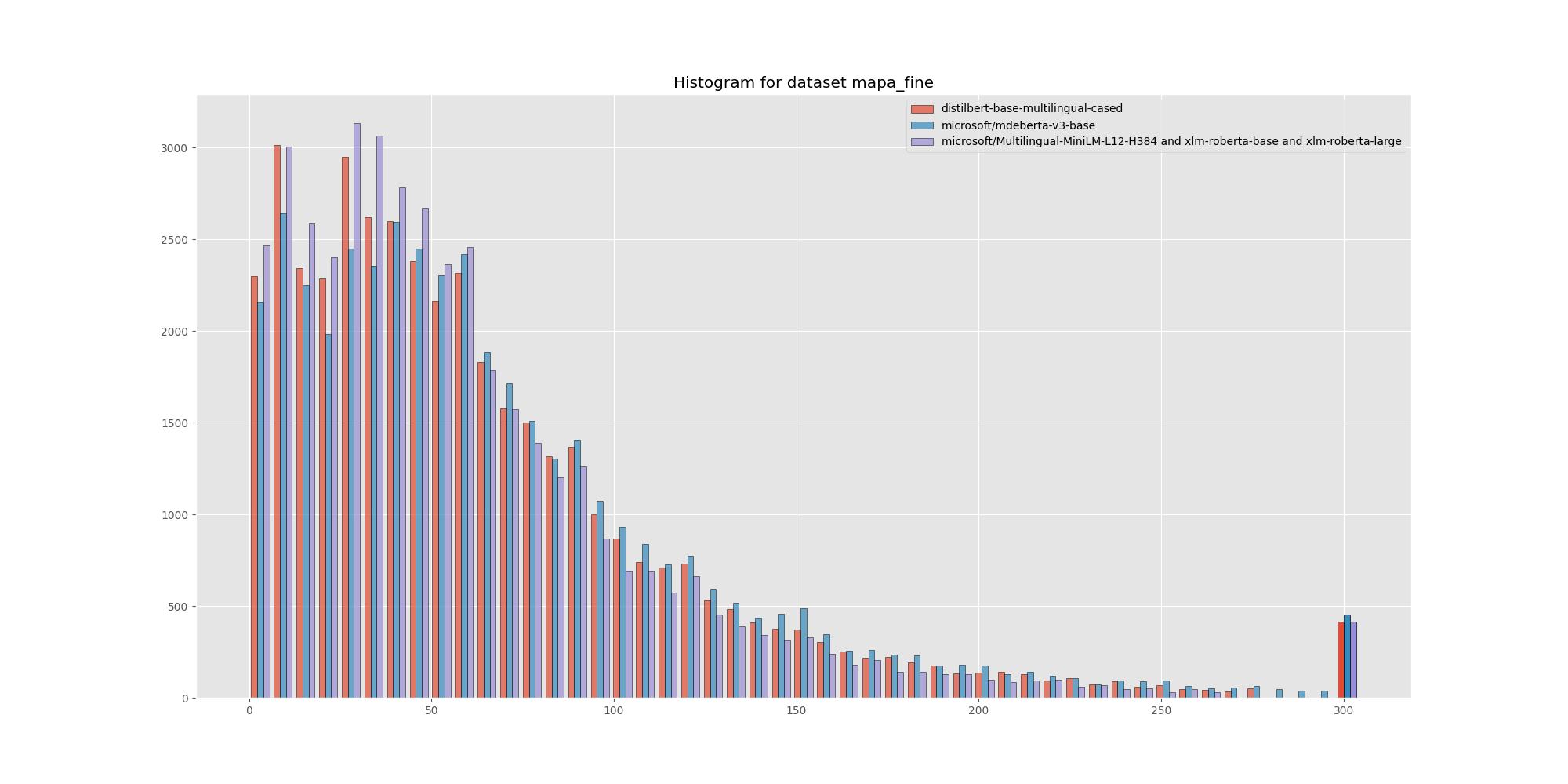}
\caption{Histogram for dataset MAP-F}
\label{histogram_MAP-F}
\end{figure}

\end{document}